\documentclass[lettersize,journal]{IEEEtran}
\usepackage{amsmath,amsfonts}
\usepackage{algorithmic}
\usepackage{algorithm}
\usepackage{array}
\usepackage[caption=false,font=normalsize,labelfont=sf,textfont=sf]{subfig}
\usepackage{textcomp}
\usepackage{stfloats}
\usepackage{url}
\usepackage{verbatim}
\usepackage{graphicx}
\usepackage[style=ieee]{biblatex}
\usepackage{booktabs} 
\usepackage{color}

\begin{document}

\title{Unpaired Overwater Image Defogging Using Prior Map Guided CycleGAN}

\author{Yaozong Mo, Chaofeng Li, ~\IEEEmembership{Senior Member,~IEEE}, Wenqi Ren, ~\IEEEmembership{Member,~IEEE}, Shaopeng Shang, Wenwu Wang, ~\IEEEmembership{Senior Member,~IEEE}, and Xiao-jun Wu
\thanks{This work was supported by the National Natural Science Foundation of China under Grant 62176150. (Corresponding author: Chaofeng Li and Shaopeng Shao.)

Yaozong Mo and Chaofeng Li are with the Institute of Logistics Science and Engineering, Shanghai Maritime University, Shanghai, China, 201306 (e-mail: moyaozong@outlook.com; wxlichaofeng@126.com).

Wenqi Ren is with the School of Cyber Science and Technology, Sun Yat-sen University, Shenzhen, China, 518000 (e-mail: renwenqi@iie.ac.cn).

Shaopeng Shang is with the Vocational College, Shanghai Jian Qiao University, Shanghai, China, 201306 (e-mail: shaopengshang@126.com).

Wenwu Wang is with the Center for Vision Speech and Signal Processing, Department of Electrical and Electronic Engineering, University of Surrey, Surrey GU2 7XH, U.K. (e-mail: w.wang@surrey.ac.uk).

Xiao-jun Wu is with the School of Artificial Intelligence and Computer Science, Jiangnan University, Wuxi, China, 214122 (e-mail: wu\_xiaojun@jiangnan.edu.cn).}
}



\maketitle

\begin{abstract}
Deep learning-based methods have achieved significant performance for image defogging. However, existing methods are mainly developed for land scenes and perform poorly when dealing with overwater foggy images, since overwater scenes typically contain large expanses of sky and water. In this work, we propose a Prior map Guided CycleGAN (PG-CycleGAN) for defogging of images with overwater scenes. To promote the recovery of the objects on water in the image, two loss functions are exploited for the network where a prior map is designed to invert the dark channel and the min-max normalization is used to suppress the sky and emphasize objects. However, due to the unpaired training set, the network may learn an under-constrained domain mapping from foggy to fog-free image, leading to artifacts and loss of details. Thus, we propose an intuitive Upscaling Inception Module (UIM) and a Long-range Residual Coarse-to-fine framework (LRC) to mitigate this issue. Extensive experiments on qualitative and quantitative comparisons demonstrate that the proposed method outperforms the state-of-the-art supervised, semi-supervised, and unsupervised defogging approaches. 
\end{abstract}


\section{Introduction}
\IEEEPARstart{D}{ue} to the presence of fog, images captured on lakes, rivers, and seas can be significantly degraded. This can adversely impact the performance of subsequent downstream computer vision tasks, such as object detection and tracking. Although many promising image defogging algorithms \cite{he2010single}\cite{li2017aod}\cite{qu2019enhanced} have been developed, they rarely focus on the scene of overwater and are limited in defogging overwater image. Therefore, it is necessary and valuable to conduct research on the defogging of overwater images.

According to the atmospheric scattering model \cite{mccartney1976optics}\cite{nayar1999vision}, a foggy image can be formulated as:
\begin{equation}
\label{deqn_ex1a}
I(x_i)=J(x_i)t(x_i)+A(1-t(x_i)),
\end{equation}
where $x_i$ refers to the position of a pixel, $I(x_i)$ is the captured foggy image, and $J(x_i)$ is the scene radiance. $A$ and $t(x_i)$ are the global atmospheric light and the transmission map, respectively. Existing defogging methods can be classified into two categories, prior-based and learning-based. The prior-based methods \cite{berman2016non}\cite{fattal2014dehazing}\cite{zhu2015fast} recover clean images by using physical priors through the model (1). However, these priors are not always reliable. For example, He \textit{et al.} \cite{he2010single} proposed dark channel prior to obtain the transmission map which assumes that at least one channel in the RGB space is close to zero in a clean natural image. As a result, this method may fail when dealing with the sky or other scene objects which are similar to atmospheric light.

\begin{figure}[!t]
\centering
\subfloat{\includegraphics[width=1in]{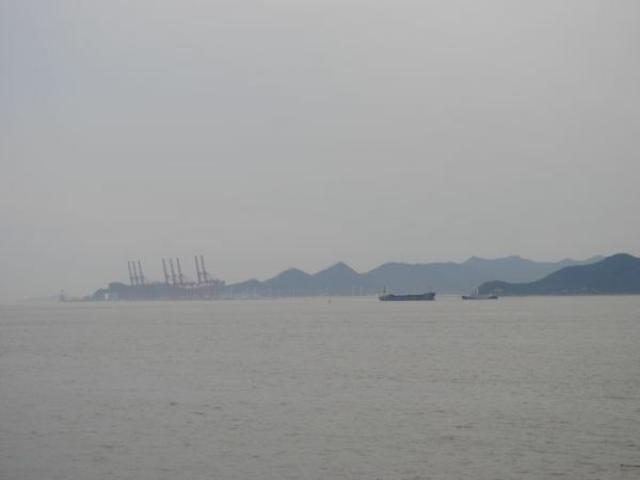}%
\label{}}
\hspace{0.3mm}
\subfloat{\includegraphics[width=1in]{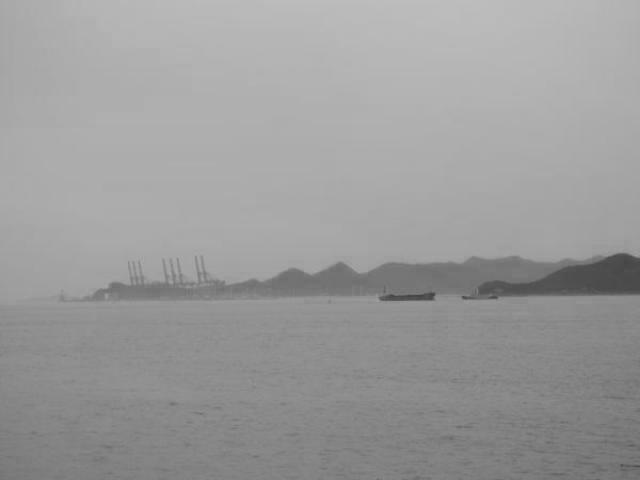}%
\label{}}
\hspace{0.3mm}
\subfloat{\includegraphics[width=1in]{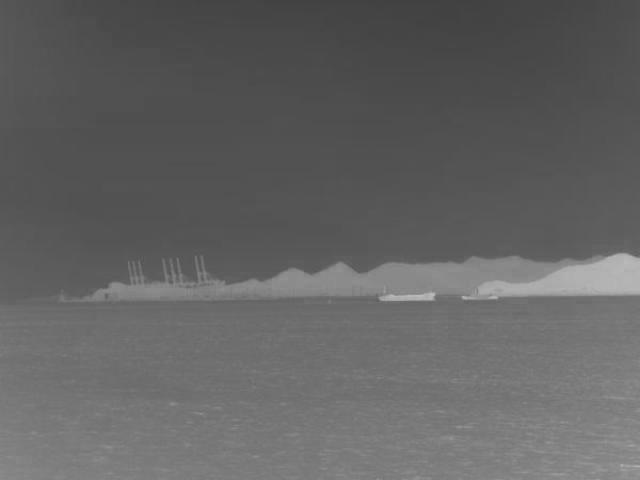}%
\label{}}\\
\makebox[1in][c]{\scriptsize (a) Foggy}\hspace{0.3mm}\makebox[1in][c]{\scriptsize (b) Dark channel}\hspace{0.3mm}\makebox[1in][c]{\scriptsize (c) Inverted DC}\\\vspace{-2mm}
\subfloat{\includegraphics[width=1in]{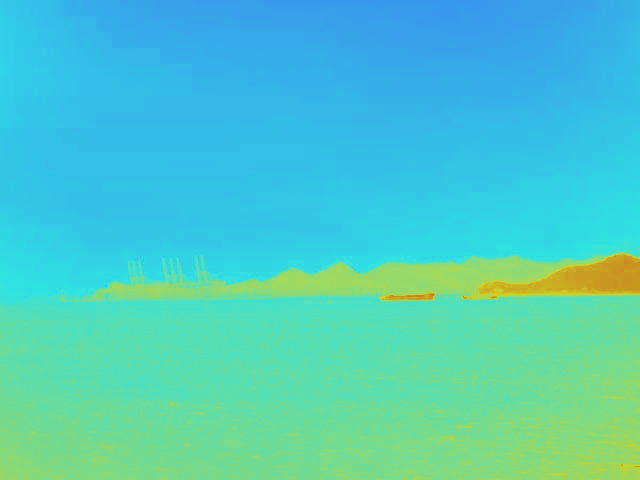}%
\label{}}
\hspace{0.3mm}
\subfloat{\includegraphics[width=1in]{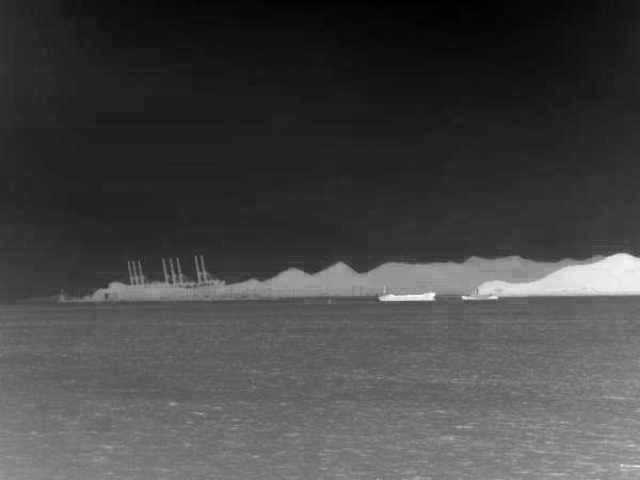}%
\label{}}
\hspace{0.3mm}
\subfloat{\includegraphics[width=1in]{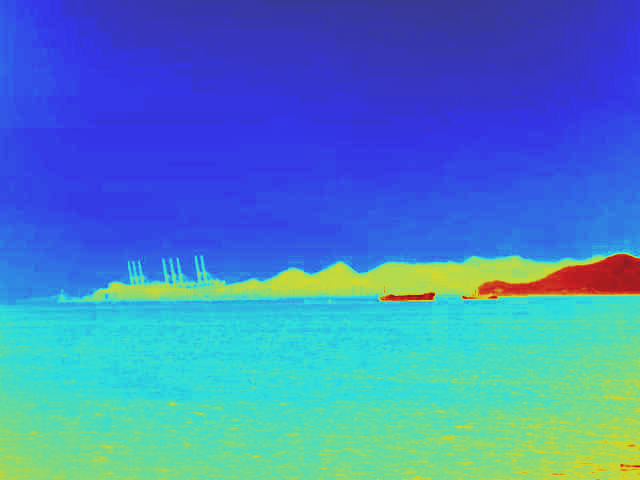}%
\label{}}\\
\makebox[1in][c]{\scriptsize (d) Heatmap of (c)}\hspace{0.3mm}\makebox[1in][c]{\scriptsize (e) Prior map}\hspace{0.3mm}\makebox[1in][c]{\scriptsize (f) Heatmap of (e)}
\caption{Comparison between the dark channel and the prior map.}
\label{}
\end{figure}

Unlike prior-based methods which perform well only for certain scenarios, the learning-based methods have better generalization ability. Conventional dehazing methods estimate clean image by using convolutional neural networks (CNNs) or generative adversarial networks (GANs) \cite{goodfellow2014generative} via supervised learning. Due to the inherent domain shift problem (i.e., mismatch between the synthetic and real domains), these methods often struggle when dealing with real-world foggy images. Thus, some algorithms have been developed to bridge the gap between the synthetic and real domains. For example, Yang \textit{et al.} \cite{yang2018towards} proposed an unsupervised Disentangled Dehazing Network which blends two recovered scene radiances from different aspects. However, due to the losses being calculated on these two intermediate outputs, the final results may be suboptimal. Li \textit{et al.} \cite{9745359} proposed a USID-Net that maps the unpaired images into a shared latent space and then reconstructs the corresponding images. Due to the under-constrained domain mapping from the foggy to fog-free image, the weights of the loss functions optimized for the USID-Net need to be carefully adjusted to guide the network training. Although many semi-supervised and unsupervised learning algorithms  \cite{golts2019unsupervised}\cite{li2019semi}\cite{shao2020domain} have achieved significant performance, little attention has been paid to the task of overwater image defogging in the literature. In this work, we focus on the problem of overwater image defogging, and propose a new method based on the characteristics of the overwater scene.

Different from the images captured on land, images captured on water usually contain large sky and water areas. Although objects such as boats on the water often occupy only a small area, their recovery is more important than the recovery of sky, as it directly affects subsequent downstream tasks such as object detection and tracking. Taking this into account, we propose a prior map which facilitates the network to focus on the recovery of objects. Fig. 1 compares the dark channel (DC) and the prior map. In dark channels, the sky and water regions tend to be brighter than the objects on the water because their pixel values are generally close to the atmospheric light. Based on this feature, the objects will be highlighted when the dark channel is inverted, as shown in Fig. 1 (c). To emphasize this feature, we utilize the min-max normalization to enhance the contrast, which stretches the pixel values in the middle region while keeping the sky regions darker and the objects brighter, as illustrated in Fig. 1 (e), and we call it the prior map. By taking advantage of the prior map, we develop a prior map guided cycle-consistency loss and a prior map guided GAN loss, which are weighted by the prior map to impose a larger penalty on the objects than the sky. Further experiments show that the proposed prior map is not only applicable to overwater images, but also to land images, by facilitating the network to concentrate on foreground parts, such as buildings.

The original CycleGAN \cite{zhu2017unpaired} introduced a cycle-consistency loss to promote the similarity between the reconstructed image and the original input image. However, due to the unpaired training set, the network learns a mapping from a source domain (foggy) to a target domain (fog-free) that is under-constrained and may lead to artifacts and loss of detailed information. To address this problem, we design a network architecture by using an Upscaling Inception Module (UIM) inspired by Inception Network \cite{szegedy2015going} and a Long-range Residual Coarse-to-fine framework (LRC).

The motivation for using the UIM is based on the finding that deconvolution blocks tend to introduce checkerboard artifacts due to the uneven overlap when the kernel size is not divisible by the stride. On the contrary, bilinear interpolation could generate smooth results, but it restricts the representation power of the network since it is a fixed upscaling procedure. To mitigate the overlap issue, a sub-pixel convolution could be used as in Pixelshuffle \cite{shi2016real}, where the kernel size is divided by the stride. However, it is prone to introducing artifacts and difficult to achieve favorable defogging results when used alone, as observed empirically in our experiments. In contrast, the UIM can improve the results by leveraging the strengths of these three upscaling methods.

To preserve the detailed information from the input image, we build the LRC framework by establishing a long-range residual connection between the input and output layers. In addition, we extend the defogging process into three stages: coarse, finer, and finest. The latter two stages adopt the downscaled defogged image from the previous stage as additional input to enhance the defogging performance.

In summary, this paper presents the following contributions:
\vspace{-0.5cm}

\begin{itemize}
\item{We propose an overwater scene-oriented unsupervised image defogging framework with a prior map that can highlight the objects on the water, based on the characteristic of the overwater images which often contain large sky and water areas.}
\item{We design a prior map guided GAN loss and a prior map guided cycle-consistency loss to facilitate the network in recovering the objects on the water. Experimental results show that the proposed prior map is not only applicable to overwater images, but also to land images.}
\item{We further propose an Upscaling Inception Module and a Long-range Residual Coarse-to-fine framework for mitigating the checkerboard artifacts and for recovering image details.}
\end{itemize}

The organization of the paper is as follows. Section \uppercase\expandafter{\romannumeral2} reviews the relevant literature. Section \uppercase\expandafter{\romannumeral3} describes the details of the prior map guided CycleGAN. Section \uppercase\expandafter{\romannumeral4} presents the experimental results and Section \uppercase\expandafter{\romannumeral5} concludes this paper.

\section{Related Work}
In this section, we briefly review the prior-based and learning-based single image defogging methods, which are related to our work.

\subsection{Prior-Based Methods}
The prior-based methods have been widely used in the past few years, via estimating the transmission maps and atmospheric light based on the statistics of natural images. Fattal \textit{et al.} \cite{fattal2008single} formulated a refined image degradation model \cite{mccartney1976optics} that accounts for a shading factor and surface reflectance coefficients. Tan \textit{et al.} \cite{tan2008visibility} proposed an automated method based on two observations: clean images have a higher contrast than foggy images and the variation of airlight changes smoothly across small regions. He \textit{et al.} \cite{he2010single} proposed a dark channel prior (DCP), which assumes that there is at least one channel in the RGB space that is close to zero in a clean natural image, to estimate transmission map and atmospheric light. Following He \textit{et al.} \cite{he2010single}, many works, such as \cite{tarel2009fast}\cite{meng2013efficient}, improved the dark channel prior for the image defogging problem. In addition, Zhu \textit{et al.} \cite{zhu2015fast} recover the depth information by building a linear model under the color attenuation prior, where the parameters of the linear model are learned with a supervised learning method. Fattal \textit{et al.} \cite{fattal2014dehazing} assert that pixels in local patches of images typically exhibit a one-dimensional distribution called color-lines. Similarly, Berman \textit{et al.}  \cite{berman2016non} assume that a few hundred distinct colors can well approximate the colors of a clean image, and then perform image defogging based on this non-local prior. Based on the assumption that there is a linear relationship between the foggy and clean images in the minimum channel, Wang \textit{et al.} \cite{7814312} developed a computationally efficient defogging method. However, the accuracy of the estimated transmission map greatly affects the defogging effect. To address this problem, Liu \textit{et al.} \cite{9537303} reformulate the image defogging problem as a task of enhancing local visibility and global contrast that does not rely on atmospheric physical models. Although these methods show effectiveness in image defogging, they may fail when dealing with certain scenarios. For instance, dark channel prior \cite{he2010single} may fail for some scene objects (such as sky and white building) which are similar to the atmospheric light. Using the haze-line prior \cite{berman2016non} for the dense fog may introduce color distortion. In comparison, we optimize the network by building the prior into the loss functions, and our data-driven model offers good generalization ability.

\begin{figure*}
\centering
\includegraphics[width=5in]{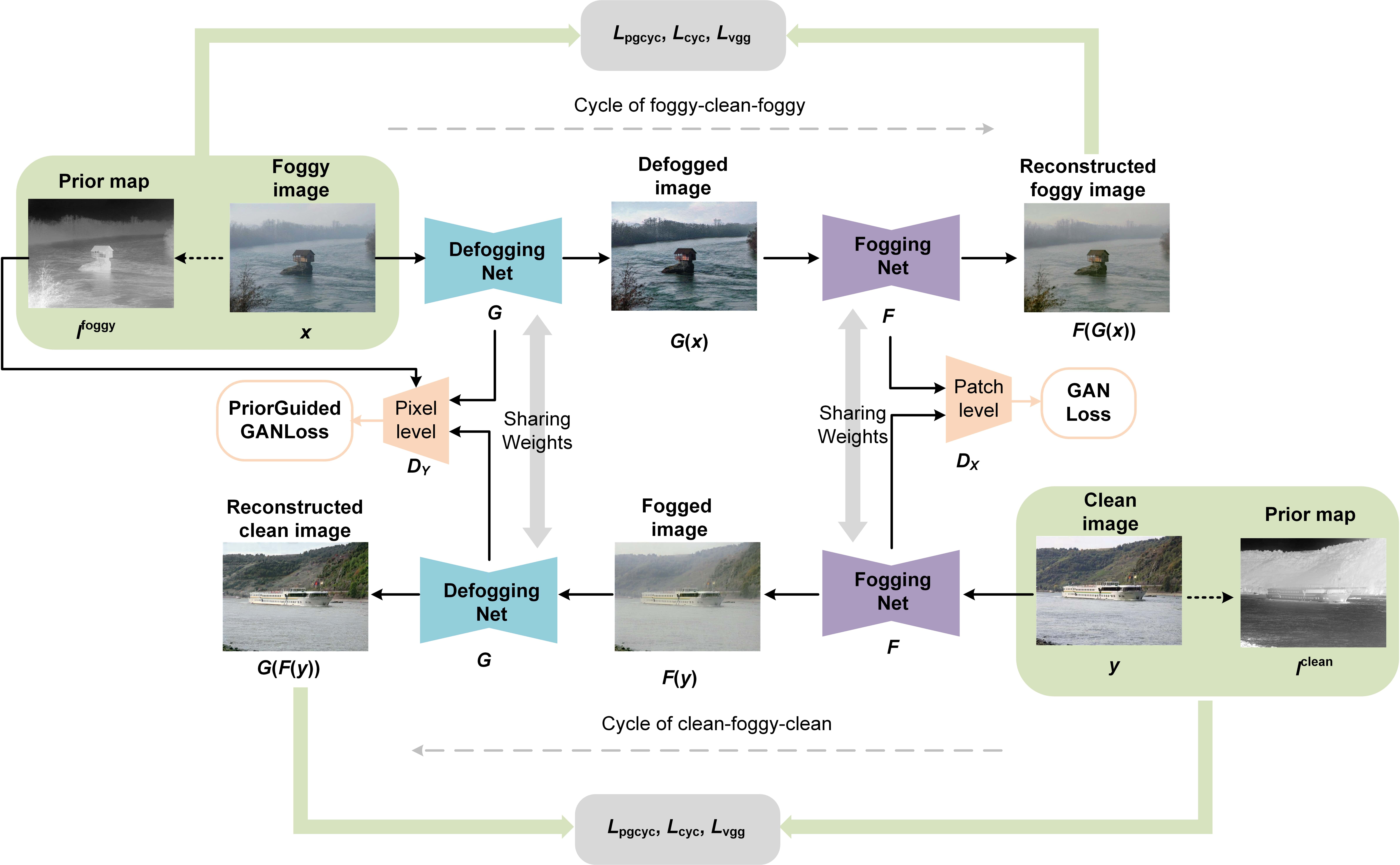}
\caption{The architecture of the proposed PG-CycleGAN. It consists of two generators, Defogging Net $G$ and Fogging Net $F$, as well as two discriminators, $D_X$ and $D_Y$. The generator $G$ converts foggy image to clean image, while $F$ converts clean image to foggy image. The two discriminators $D_X$ and $D_Y$ distinguish between the real image and the generated image. After forward propagation, several losses are calculated, where the prior map is used as weight map, to update the network.}
\label{fig_sim}
\end{figure*}

\subsection{Learning-Based Methods}
With the advance in CNNs and the support of large-scale datasets, data-driven methods for image defogging have drawn significant attention in recent years. Cai \textit{et al.} \cite{cai2016dehazenet} exploited an end-to-end trainable CNN to estimate the transmission map in an atmospheric scattering model for recovering the clean image. Ren \textit{et al.} \cite{ren2016single} proposed a multi-scale neural network named MSCNN for image defogging, which consists of a coarse net and a fine net to predict the transmission map. Unlike the above methods using CNN for estimating the transmission map, Li \textit{et al.} \cite{li2017aod}  reformulated the atmospheric scattering model to combine the two unknown variables, i.e., transmission map and atmospheric light, into one variable. Then, they leverage a lightweight CNN to estimate this variable. Instead of using physical models, Ren \textit{et al.} \cite{ren2018gated} adopted a fusion-based strategy to obtain a confidence map by integrating white balance, contrast enhancing, and gamma correction, derived from the original foggy image, and then recovered the clean image by weighting these three maps with a confidence map. Liu \textit{et al.} \cite{liu2019griddehazenet} remove fog by using a grid network constructed with three modules: pre-processing, backbone, and post-processing, which can effectively alleviate the bottleneck issue and help reduce the artifacts in the final output. Qu \textit{et al.} \cite{qu2019enhanced} proposed an Enhanced Pix2pix Dehazing Network (EPDN) which contains two enhancing blocks in the tail of decoder to refine the intermediate output. Different from other GAN-based methods, the discriminator supervises only the intermediate result of EPDN. Dong \textit{et al.} \cite{dong2020multi} proposed a multi-scale and dense connected network for image defogging, where the strengthen-operate-subtract boosting strategy is incorporated in the decoder, and the dense feature fusion module is used to extract the features from non-adjacent levels and to restore the spatial information. Qin \textit{et al.} \cite{qin2020ffa} also adopted a fusion strategy, where they applied channel attention and pixel attention after the high-level feature for giving more weight to important features. To efficiently extract the features from the input image, Yi \textit{et al.} \cite{9473023} proposed a Multi-scale Topological Network (MSTN) that uses a multi-scale feature fusion module and an adaptive feature selection module to achieve progressive image defogging. However, this multi-branch network architecture increases the complexity of the network and further increases the training time.

Although the supervised learning-based methods have achieved favorable performance, they are limited to synthetic images and perform poorly on real-world images because of the inherent domain shift introduced by the synthetic training set. Hence, some semi-supervised and unsupervised methods are integrated to enhance the real-world foggy images. Engin \textit{et al.} \cite{engin2018cycle} introduced an enhanced CycleGAN which adopt unpaired real-world images as training set. Li \textit{et al.} \cite{li2019semi} proposed a semi-supervised network by combining a supervised learning branch with an unsupervised learning branch. Shao \textit{et al.} \cite{shao2020domain} proposed a domain adaptation paradigm by incorporating two image defogging modules and an image translation module to bridge the gap between the synthetic and real domains. Golts \textit{et al.} \cite{golts2019unsupervised} exploited a dark channel prior energy function, which no longer relies on paired images to calculate the loss for the optimization of the network. Zhao \textit{et al.} \cite{zhao2021refinednet} proposed a semi-supervised two stage network, which utilized a CNN to refine the defogged image of DCP \cite{he2010single}. Chen \textit{et al.} \cite{chen2021psd} select three physical priors to constitute a loss committee, i.e., dark channel loss, bright channel loss, and contrast limited adaptive histogram equalization loss, that guides the unsupervised training.

The above mentioned semi-supervised and unsupervised methods offer promising results on real-world images, but little attention has been paid to the image defogging of overwater scenes. Recently, Zheng \textit{et al.} \cite{zheng2020overwater} proposed an enhanced CycleGAN to tackle the overwater image defogging problem. Nevertheless, this method simply replaces the training set of land foggy images as used in other methods with overwater images. Its defogging performance is limited as this method does not leverage the properties of the overwater scenes. To this end, our method focuses on the characteristics of overwater foggy images, and improves the defogging performance and visual quality of the defogged images.

\section{Proposed Method}
In this section, the proposed Prior map Guided CycleGAN (PG-CycleGAN) will be presented in detail. Fig. 2 illustrates the architecture of the PG-CycleGAN, it consists of two cycles: foggy-clean-foggy and clean-foggy-clean. In the foggy-clean-foggy cycle, the defogging network $G$ takes the foggy image $x$ as input and learns the residual from it to get the defogged image, where the network utilizes three upscaling methods: deconvolution, bilinear interpolation, and PixelShuffle \cite{shi2016real} when decoding the intermediate features. Furthermore, the defogging procedure is split into three stages: the coarse, finer, and finest to enhance the defogging performance. Then, the discriminator $D_Y$ is used to distinguish between the clean image $y$ and the defogged image $G(x)$ to get the reconstructed foggy image $F(G(x))$. Finally, the proposed prior map is used to weight the losses for updating the networks for the recovery of the objects on water. The clean-foggy-clean cycle is similar to the foggy-clean-foggy cycle.

\begin{figure}[!t]
\centering
\subfloat{\includegraphics[width=0.16\linewidth]{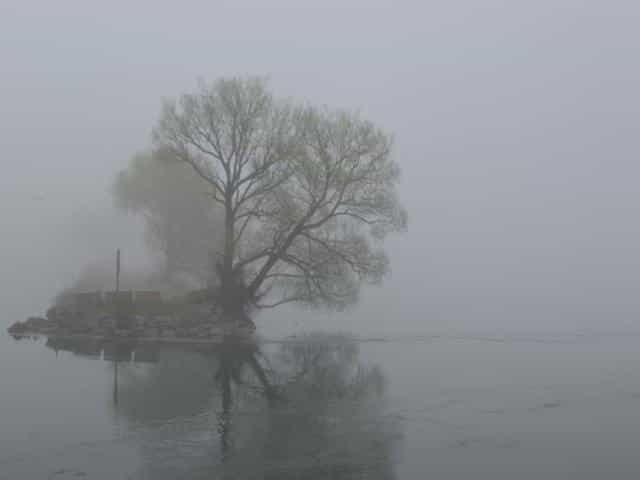}%
\label{}}
\hfill
\subfloat{\includegraphics[width=0.16\linewidth]{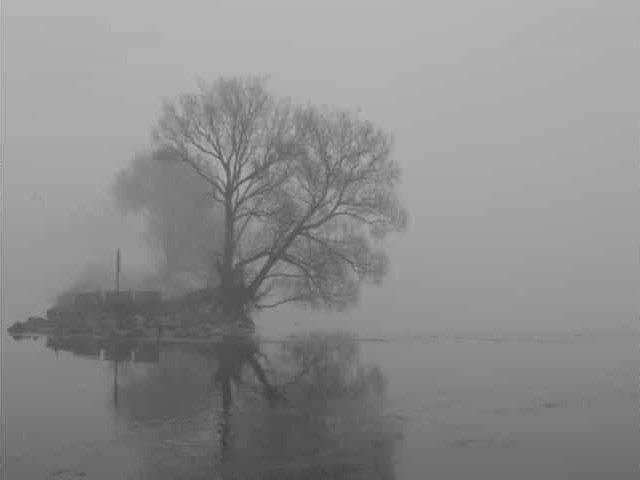}%
\label{}}
\hfill
\subfloat{\includegraphics[width=0.16\linewidth]{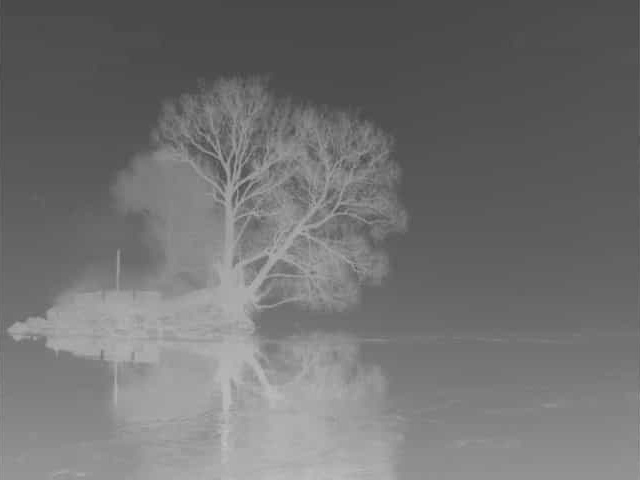}%
\label{}}
\hfill
\subfloat{\includegraphics[width=0.16\linewidth]{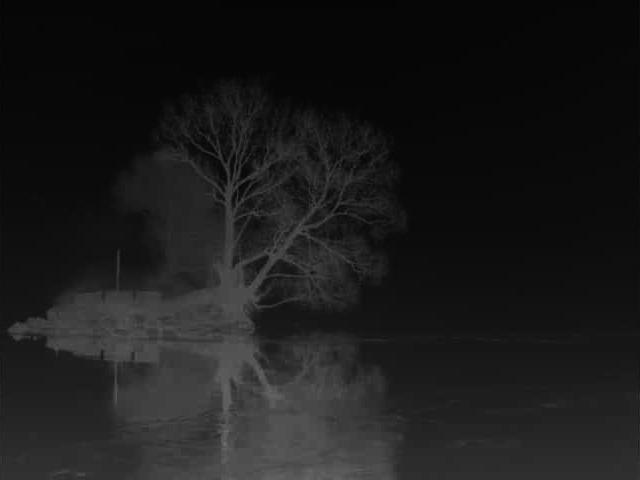}%
\label{}}
\hfill
\subfloat{\includegraphics[width=0.16\linewidth]{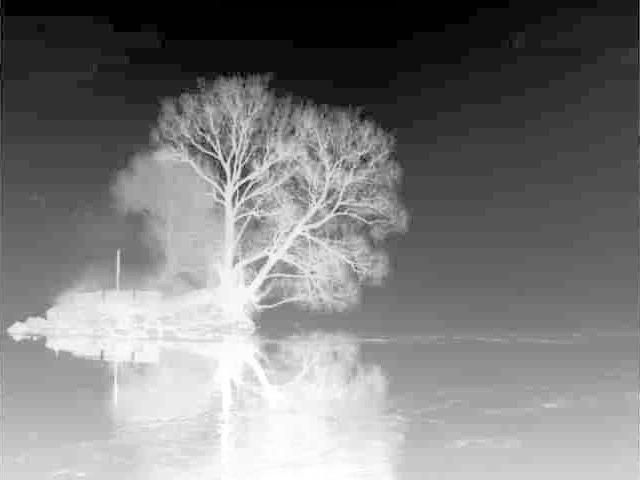}%
\label{}}
\hfill
\subfloat{\includegraphics[width=0.16\linewidth]{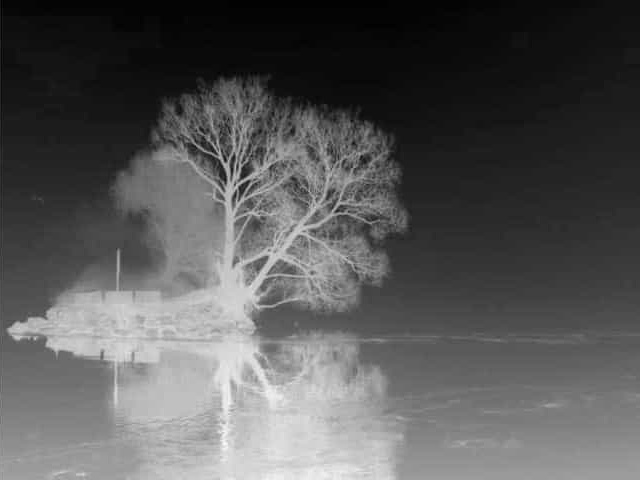}%
\label{}}\\\vspace{-3mm}
\subfloat{\includegraphics[width=0.16\linewidth]{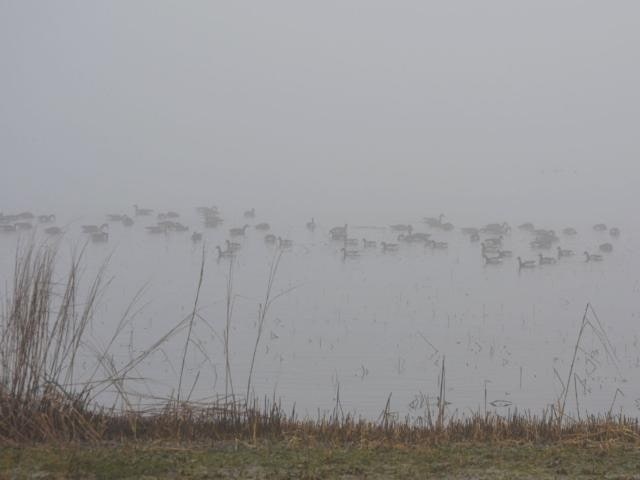}%
\label{}}
\hfill
\subfloat{\includegraphics[width=0.16\linewidth]{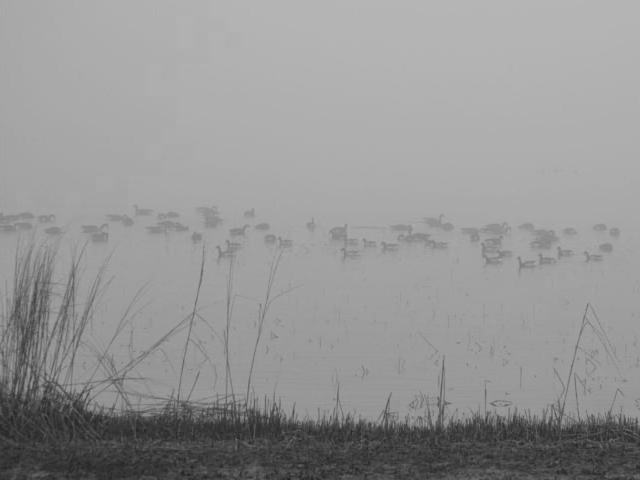}%
\label{}}
\hfill
\subfloat{\includegraphics[width=0.16\linewidth]{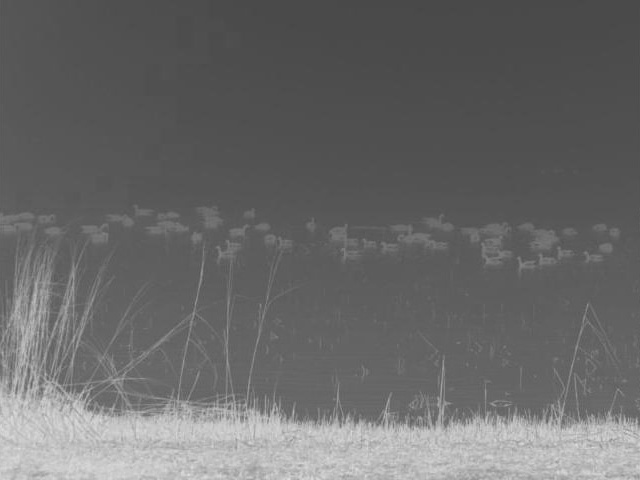}%
\label{}}
\hfill
\subfloat{\includegraphics[width=0.16\linewidth]{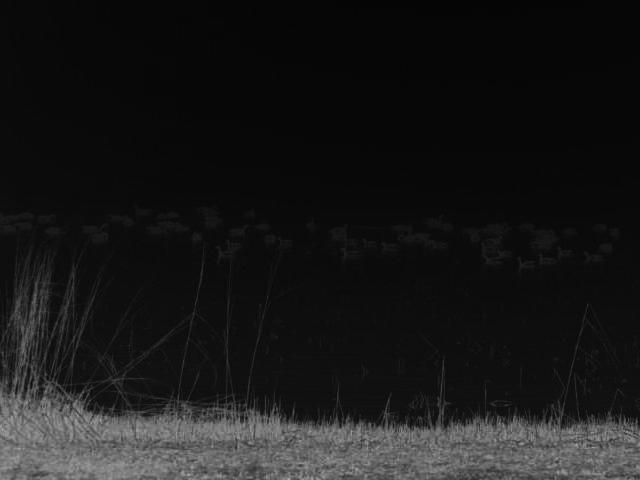}%
\label{}}
\hfill
\subfloat{\includegraphics[width=0.16\linewidth]{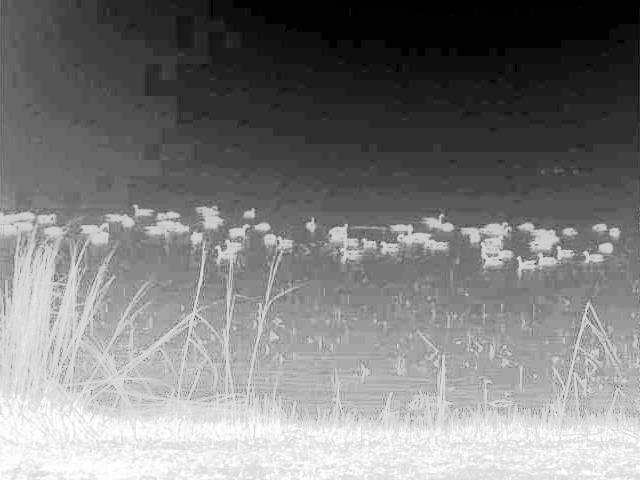}%
\label{}}
\hfill
\subfloat{\includegraphics[width=0.16\linewidth]{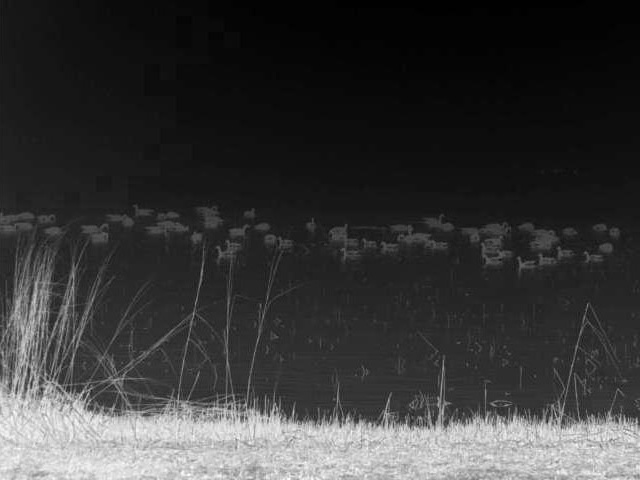}%
\label{}}\\\vspace{-3mm}
\subfloat{\includegraphics[width=0.16\linewidth]{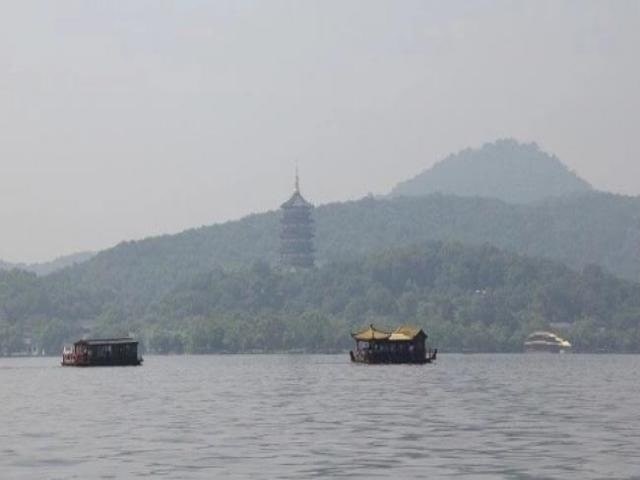}%
\label{}}
\hfill
\subfloat{\includegraphics[width=0.16\linewidth]{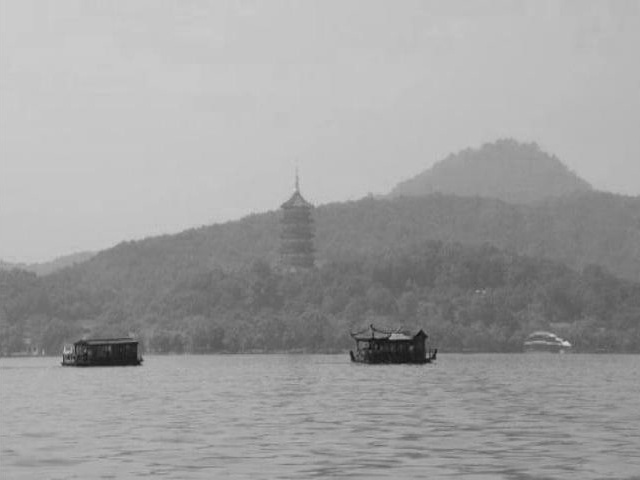}%
\label{}}
\hfill
\subfloat{\includegraphics[width=0.16\linewidth]{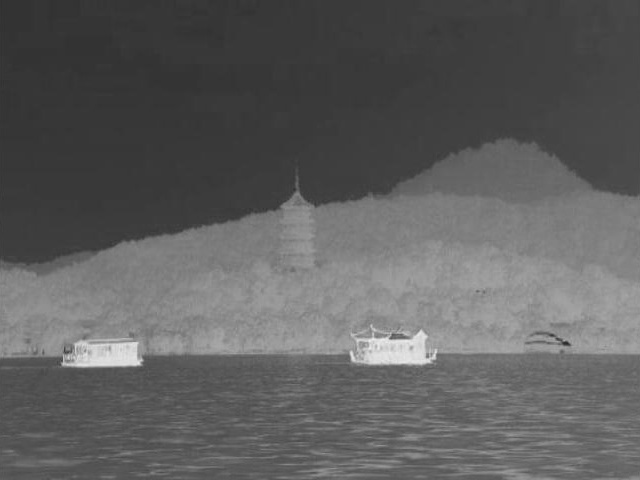}%
\label{}}
\hfill
\subfloat{\includegraphics[width=0.16\linewidth]{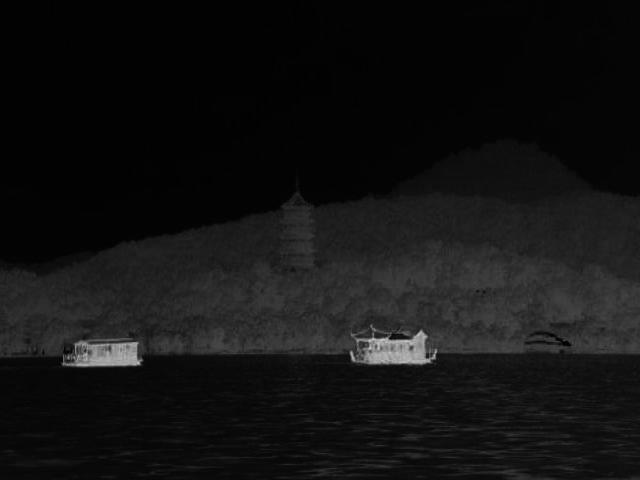}%
\label{}}
\hfill
\subfloat{\includegraphics[width=0.16\linewidth]{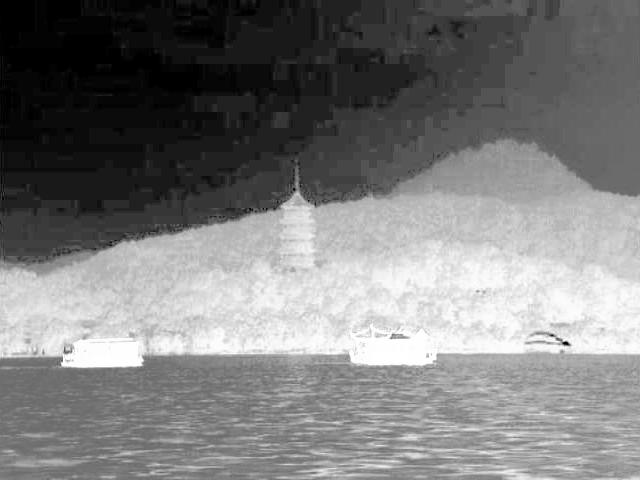}%
\label{}}
\hfill
\subfloat{\includegraphics[width=0.16\linewidth]{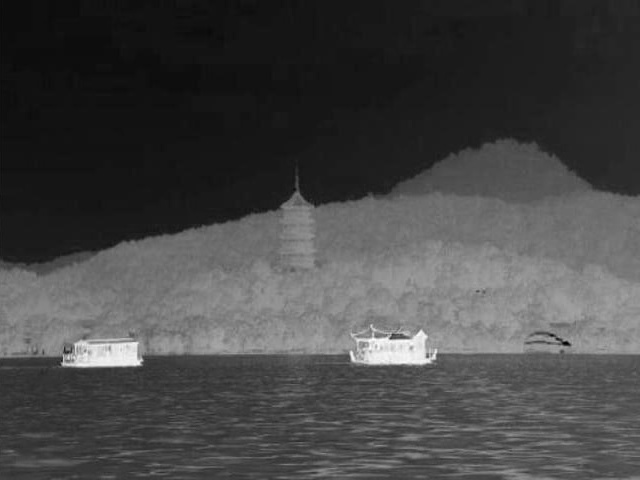}%
\label{}}\\\vspace{-3mm}
\subfloat{\includegraphics[width=0.16\linewidth]{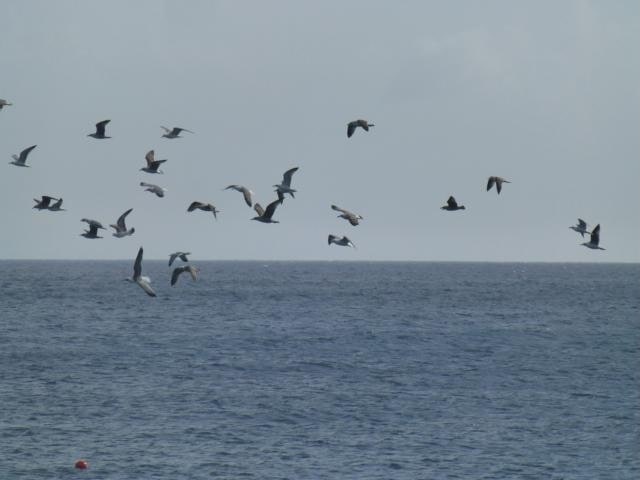}%
\label{}}
\hfill
\subfloat{\includegraphics[width=0.16\linewidth]{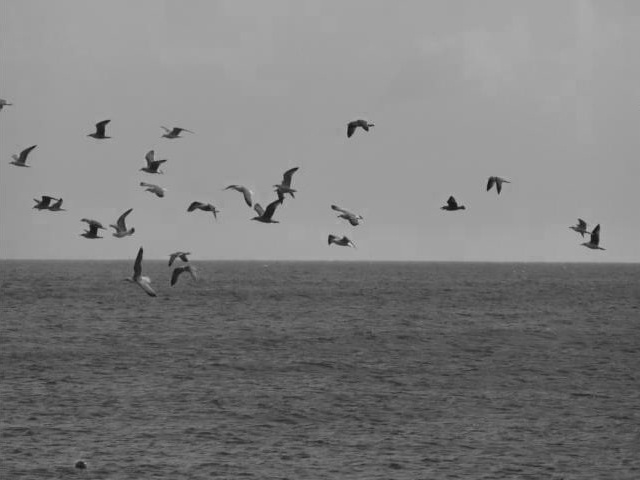}%
\label{}}
\hfill
\subfloat{\includegraphics[width=0.16\linewidth]{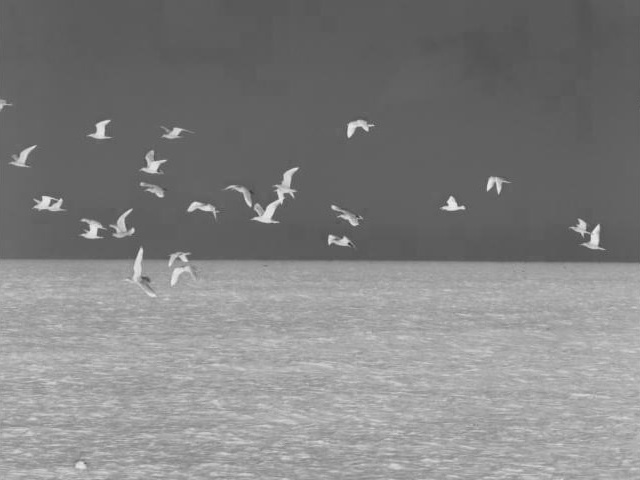}%
\label{}}
\hfill
\subfloat{\includegraphics[width=0.16\linewidth]{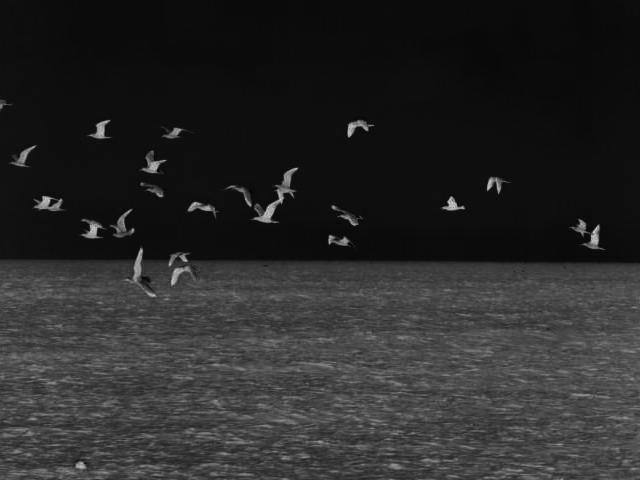}%
\label{}}
\hfill
\subfloat{\includegraphics[width=0.16\linewidth]{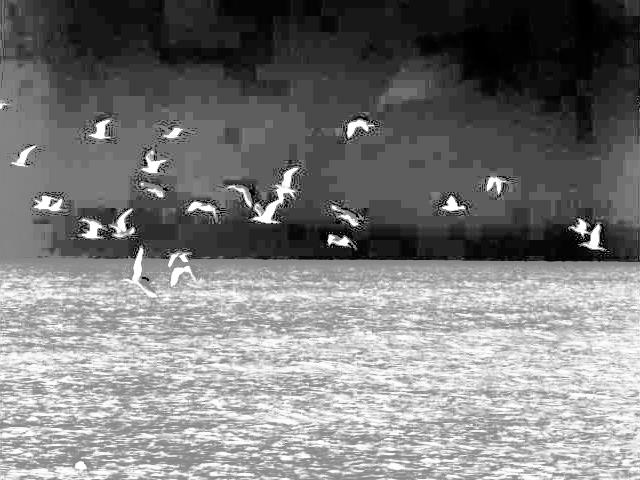}%
\label{}}
\hfill
\subfloat{\includegraphics[width=0.16\linewidth]{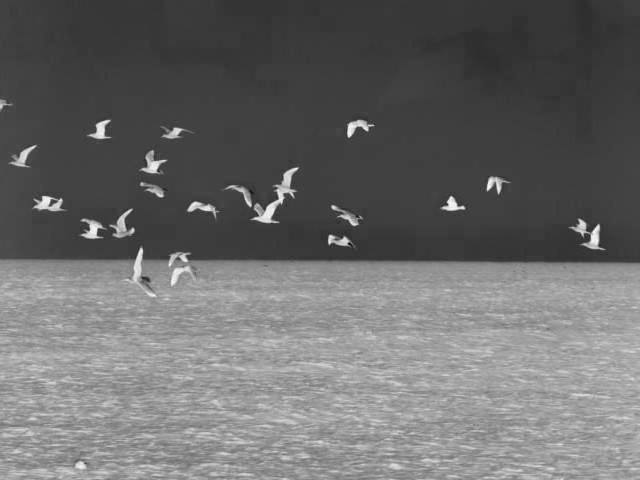}%
\label{}}\\\vspace{-3mm}
\subfloat{\includegraphics[width=0.16\linewidth]{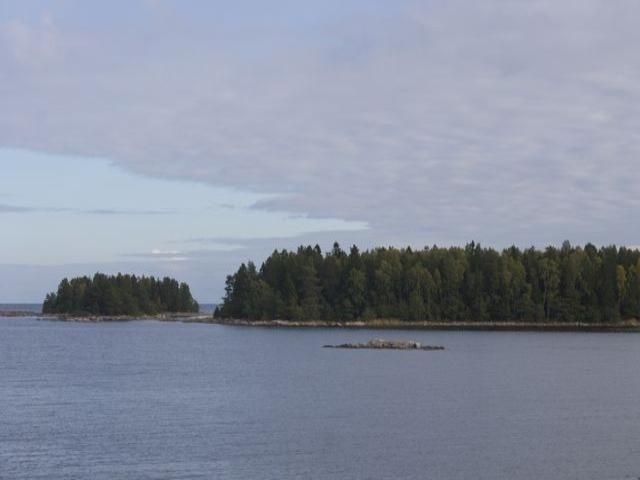}%
\label{}}
\hfill
\subfloat{\includegraphics[width=0.16\linewidth]{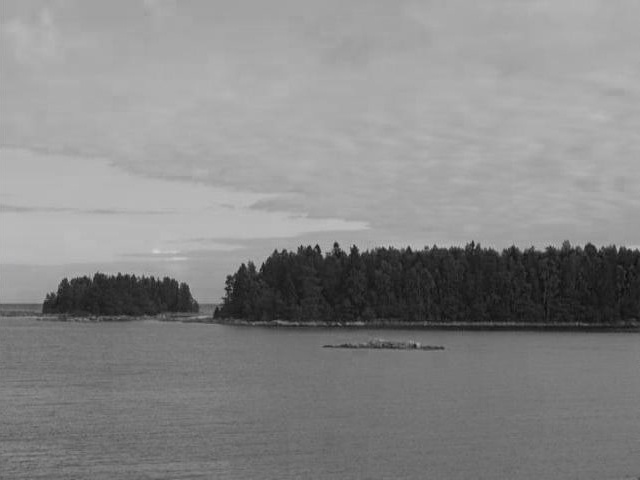}%
\label{}}
\hfill
\subfloat{\includegraphics[width=0.16\linewidth]{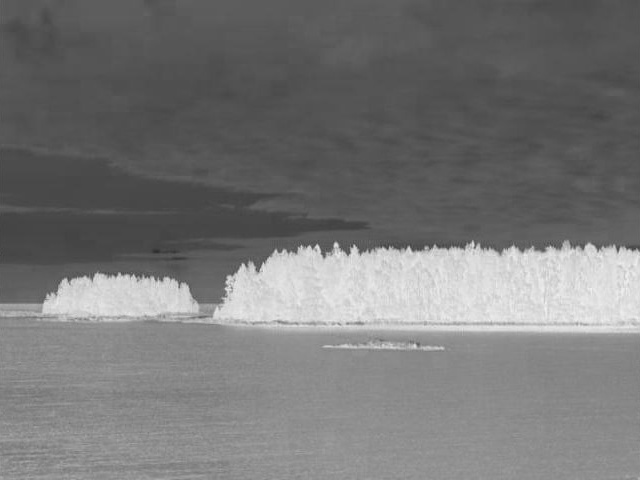}%
\label{}}
\hfill
\subfloat{\includegraphics[width=0.16\linewidth]{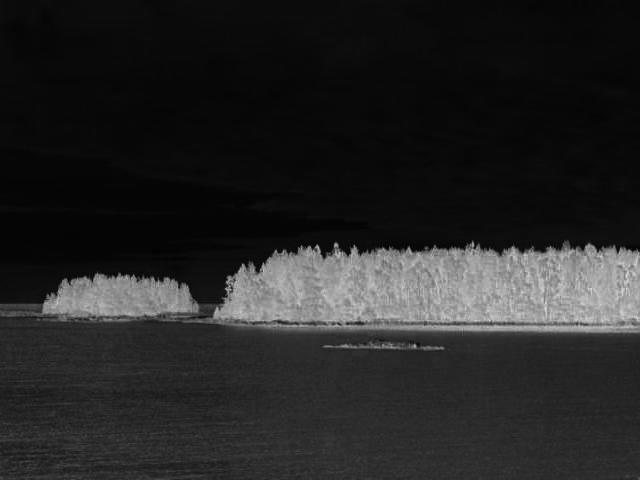}%
\label{}}
\hfill
\subfloat{\includegraphics[width=0.16\linewidth]{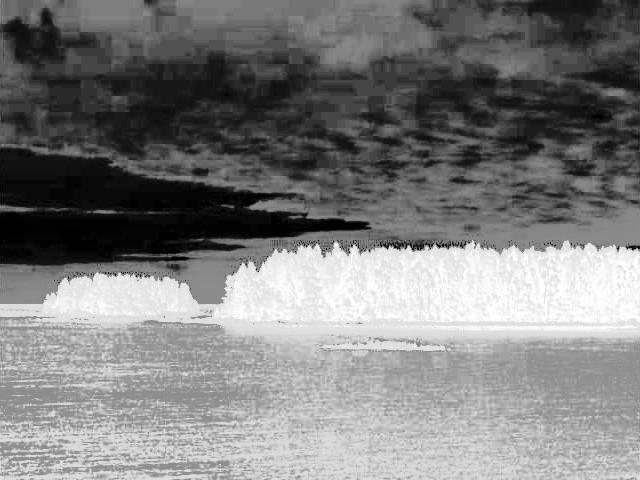}%
\label{}}
\hfill
\subfloat{\includegraphics[width=0.16\linewidth]{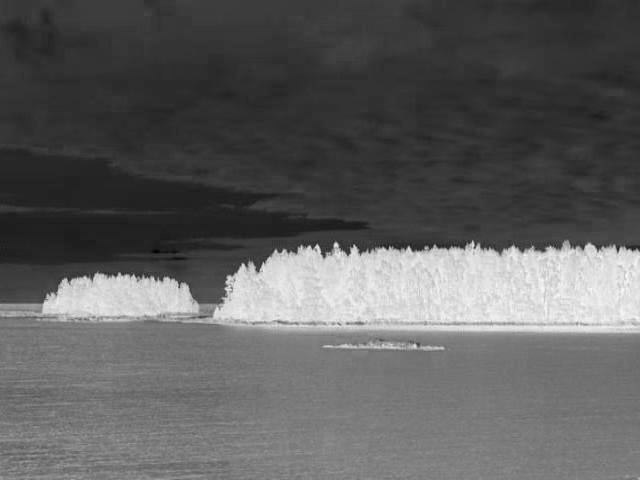}%
\label{}}\\\vspace{-3mm}
\subfloat{\includegraphics[width=0.16\linewidth]{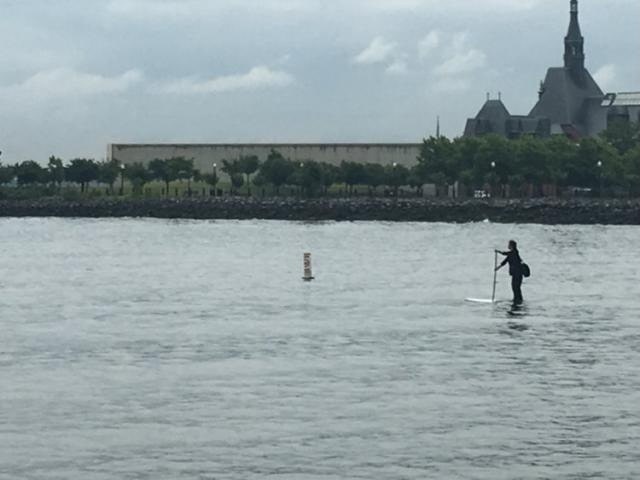}%
\label{}}
\hfill
\subfloat{\includegraphics[width=0.16\linewidth]{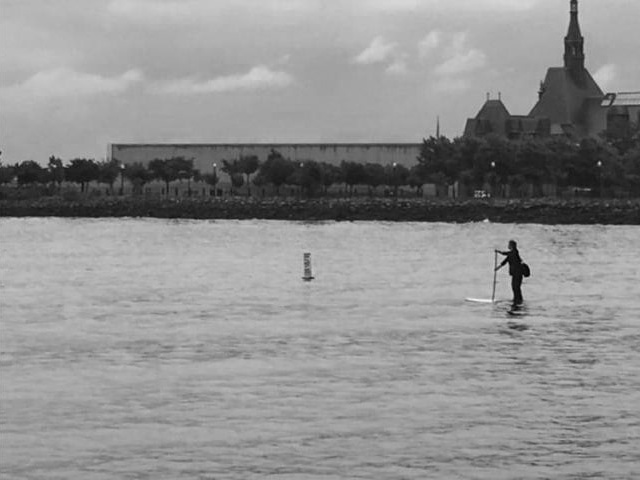}%
\label{}}
\hfill
\subfloat{\includegraphics[width=0.16\linewidth]{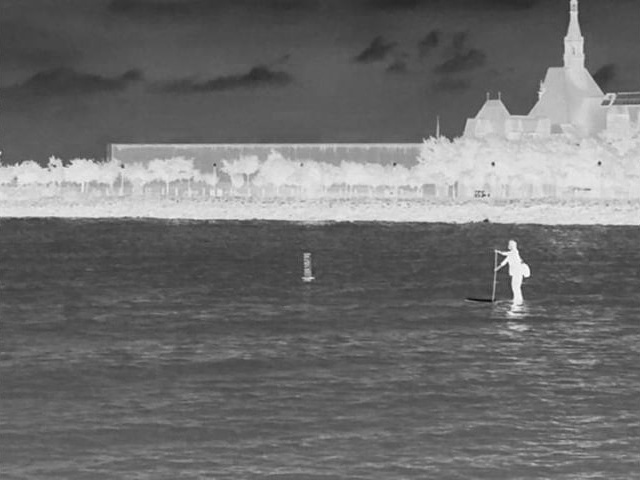}%
\label{}}
\hfill
\subfloat{\includegraphics[width=0.16\linewidth]{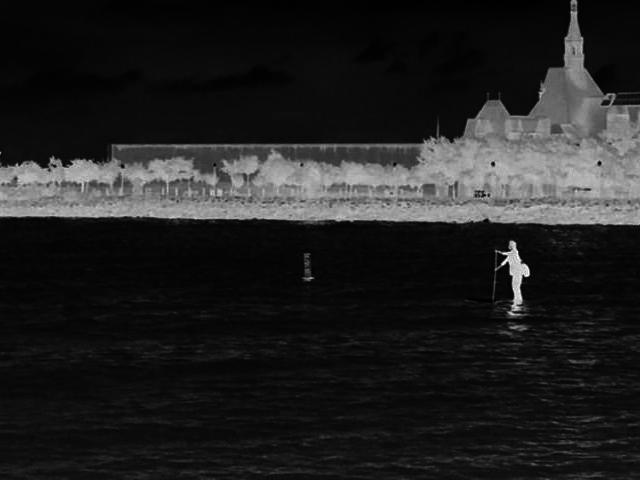}%
\label{}}
\hfill
\subfloat{\includegraphics[width=0.16\linewidth]{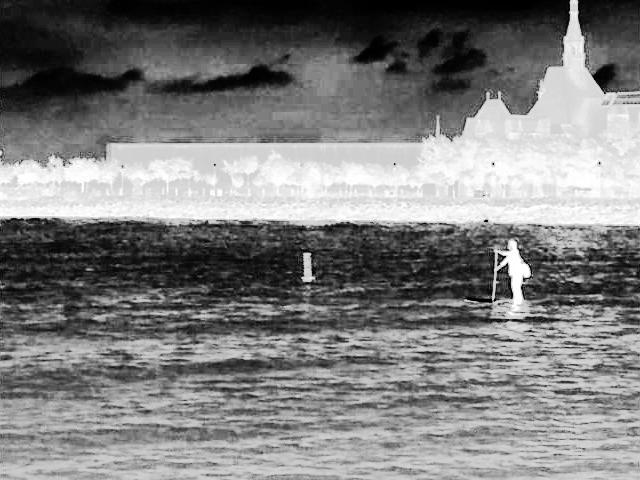}%
\label{}}
\hfill
\subfloat{\includegraphics[width=0.16\linewidth]{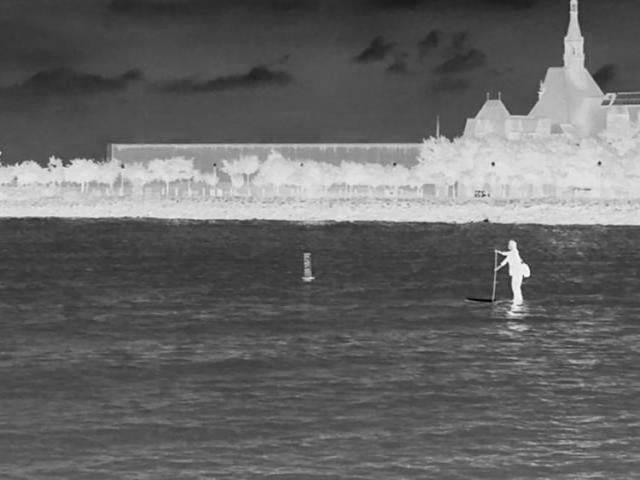}%
\label{}}\\
\makebox[0.16\linewidth][c]{\footnotesize (a) Input}\hfill\makebox[0.16\linewidth][c]{\footnotesize (b) DC}\hfill\makebox[0.16\linewidth][c]{\footnotesize (c) IDC}\hfill\makebox[0.16\linewidth][c]{\footnotesize (d) Gamma}\hfill\makebox[0.16\linewidth][c]{\footnotesize (e) HE}\hfill\makebox[0.16\linewidth][c]{\footnotesize (f) Prior map}
\caption{Difference on overwater images between the dark channel (DC), inverted dark channel (IDC),  IDC after gamma correction (gamma=3), IDC after histogram equalization (HE), and prior map. The top three rows present the comparison on foggy images, and the bottom three rows present the comparison on clean images. Here the dark channels are without performing minimum filtering.}
\label{}
\end{figure}

\begin{figure}
    \centering
    \subfloat{\includegraphics[width=0.50\linewidth]{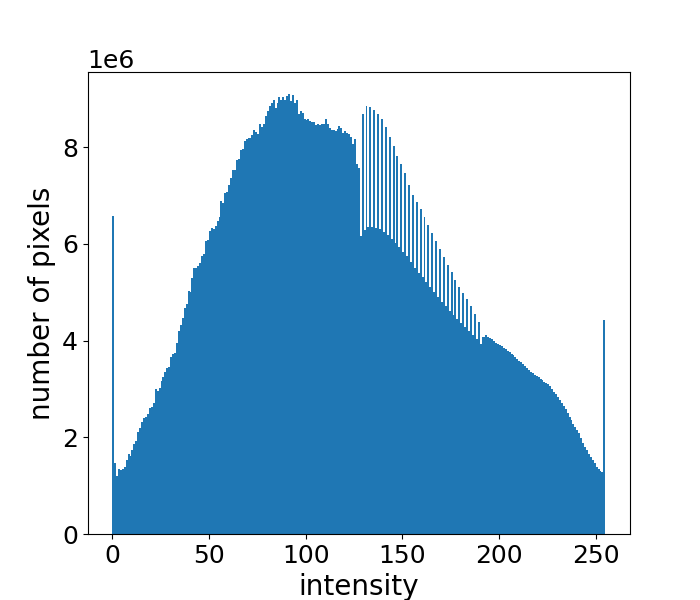}%
    \label{}}
    \subfloat{\includegraphics[width=0.50\linewidth]{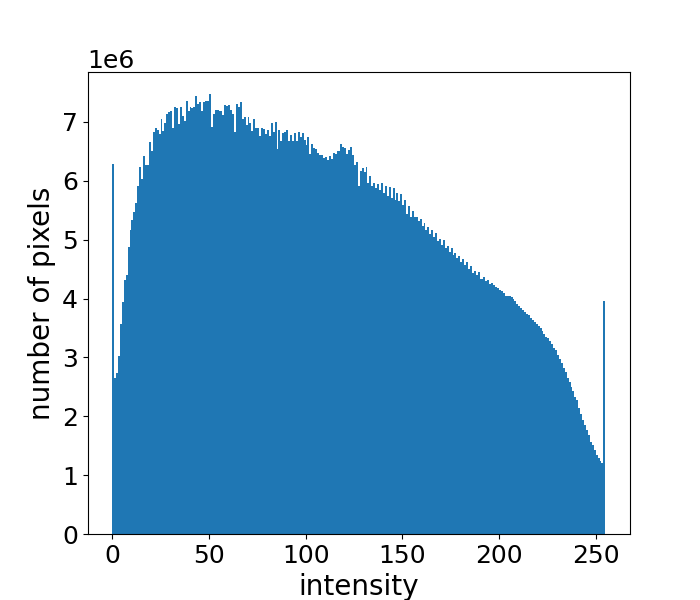}%
    \label{}}\\
    \makebox[0.49\linewidth][c]{\footnotesize(a) unnormalized}\makebox[0.49\linewidth][c]{\footnotesize (b) min-max normalized}\hfill
    \caption{Intensity histogram of 4,531 inverted dark channels. (a) histogram of unnormalized inverted dark channels. (b) histogram of min-max normalized inverted dark channels.}
    \label{fig:added_1}
\end{figure}

\subsection{Prior Map}
As He \textit{et al.} \cite{he2010single} claimed, the dark channel prior is invalid when the scene objects are similar to the atmospheric light. This is also reflected in Fig. 3 (b). It can be seen that the pixel value of the water and sky is not close to 0 in clean images.

A characteristic of the overwater scenes is that the sky and water usually occupy most of the area. In dark channels, they tend to be brighter than objects on the water because their pixel values are generally close to the atmospheric light. Based on this feature, the objects will be highlighted when the dark channel of the overwater image is inverted, as shown in Fig. 3 (c). To emphasize this feature, we try to enhance the contrast of the inverted dark channel. Fig. 3 shows the difference between the dark channel (DC), inverted dark channel (IDC), IDC after gamma correction, IDC after histogram equalization (HE), and IDC after min-max normalization (i.e. the prior map). It can be seen that though the contrast is enhanced, the gamma correction may lower the pixel value of the objects, and the histogram equalization may amplify background noise. In contrast, the min-max normalization maintains the highlight of the object while reducing noise. Fig. 4 is the intensity histogram over 4,531 inverted dark channels of an overwater image dataset \cite{zheng2020overwater} (details described in Sec. \uppercase\expandafter{\romannumeral4}-A; the images are rescaled to 256$\times$256 for computational efficiency). It can be seen that the min-max normalization stretches the pixel values in the middle region while keeping the sky regions darker and the objects brighter, this statistic provides strong support to our proposed prior map.

Specifically, given a foggy image $I$ whose pixel value is in the range of [0, 1], its corresponding dark channel can be expressed as:
\begin{equation}
\label{deqn_ex1a}
I_{dc}=\mathop{min}\limits_{y_i\in N(x_i)}(\mathop{min}\limits_{c\in \{r,g,b\}}I^c(y_i)),
\end{equation}
where $x_i$ and $y_i$ are pixel coordinates of $I$, $I^c$ denotes the $c$-th color channel of $I$, and $N(x_i)$ denotes the local neighborhood centered at $x_i$. To obtain the prior map, we first remove the minimum filtering $\mathop{min}\limits_{y_i\in N(x_i)}(\cdot)$ from the procedure for dark channel calculation, and invert the pixel value as follows:
\begin{equation}
\label{deqn_ex1a}
I_{idc}=1-\mathop{min}\limits_{c\in \{r,g,b\}}I^c,
\end{equation}
where $I_{idc}$ represents the inverted dark channel. The reasons for removing the minimum filter are two. One is that it reduces the required computational resources, as performing the sliding window algorithm is time-consuming. The other is that the minimum filter may lead to unclear boundaries of the objects in the defogging results because it exploits all the surrounding pixels.

Then, the min-max normalization is performed on the $I_{idc}$ to get the prior map,
\begin{equation}
\label{deqn_ex1a}
I_{p}=\frac{I_{idc}-min(I_{idc})}{max(I_{idc})-min(I_{idc})}.
\end{equation}

\begin{figure*}
\centering
\includegraphics[width=5in]{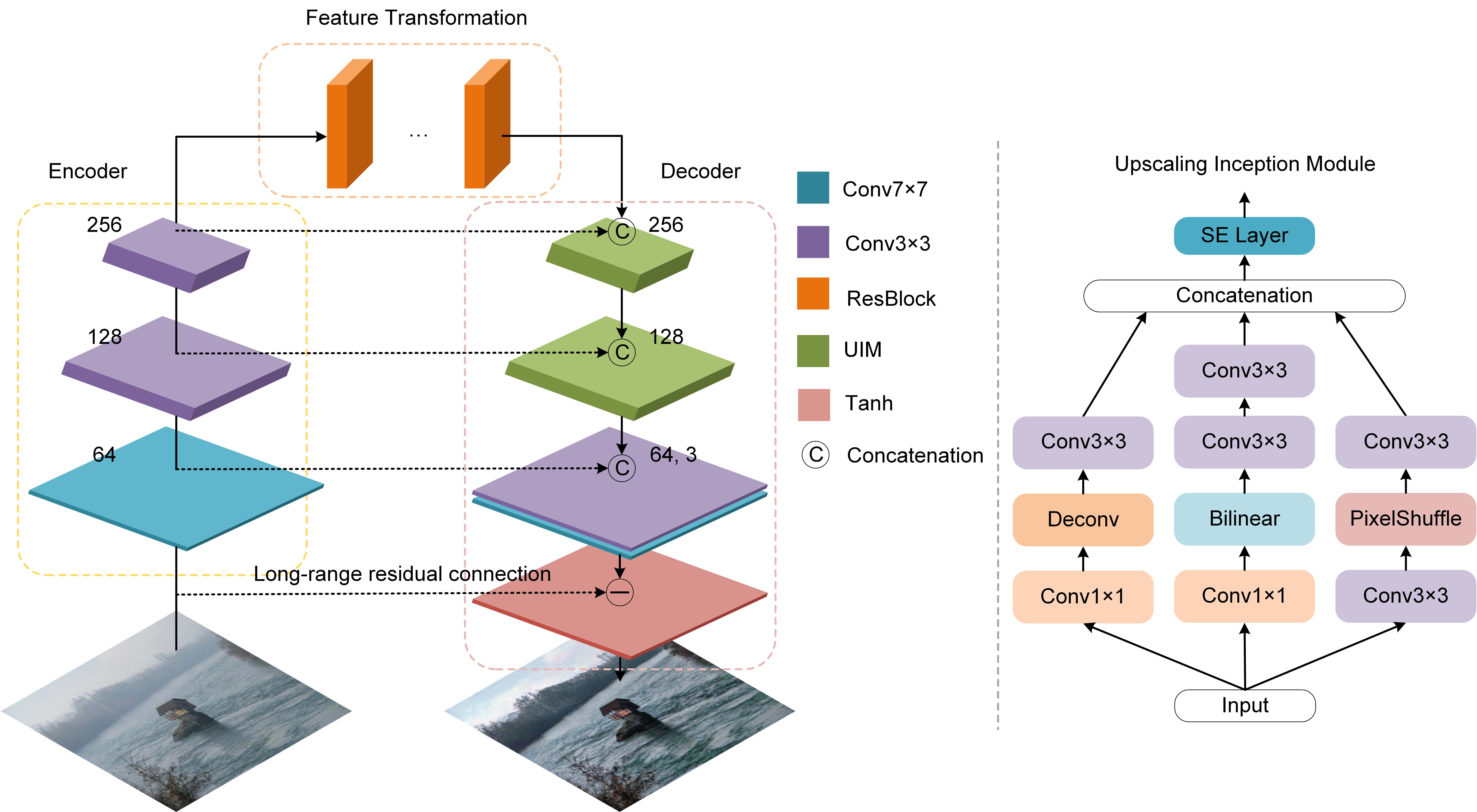}
\caption{The architecture of the defogging network with long-range residual connections. The network has an encoder, a feature transformation that consists of 9 residual blocks, and a decoder that cascades two upscaling inception modules (UIM). The details of UIM are shown on the right side.}
\label{fig_sim}
\end{figure*}

\begin{figure}
\centering
\includegraphics[width=3in]{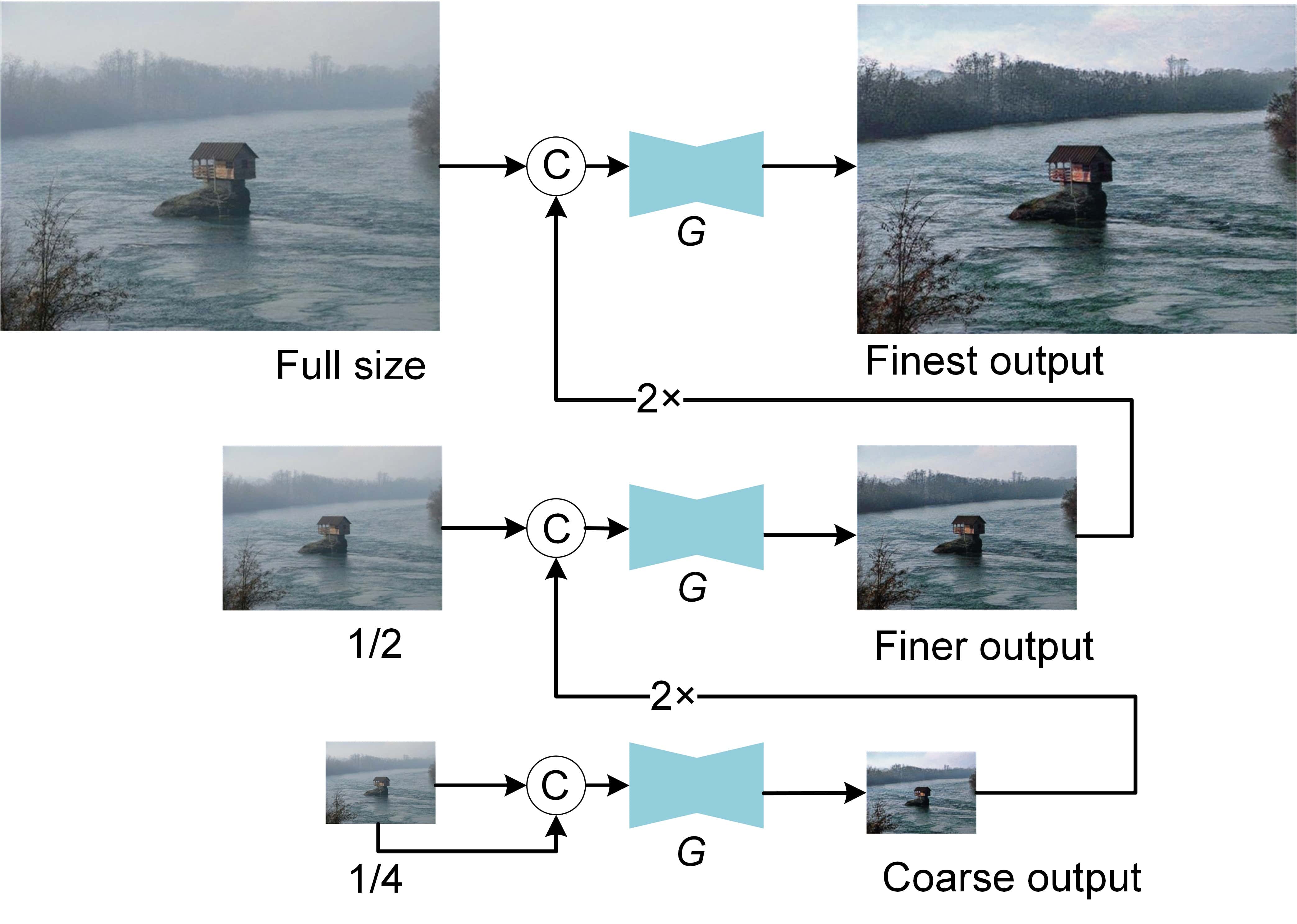}
\caption{The coarse-to-fine framework. The coarse stage takes the same two 1/4 downscaled foggy images as input. At the finer and finest stage, respectively, the 1/2 and full-size defogged images generated by the previous stage are concatenated as input. We take the output from the finest stage as the final defogged results.}
\label{fig_sim}
\end{figure}

After getting the prior map, we further define the prior map guided cycle-consistency loss:
\begin{multline}
\label{deqn_ex1a}
L_{pgcyc}=||(F(G(x))-x)I_p^{foggy}||_1\\
+||(G(F(y))-y)I_p^{clean}||_1,
\end{multline}
and the prior map guided GAN loss:
\begin{multline}
\label{deqn_ex1a}
L_{pgG} = \mathbb{E}_{y\sim p_{data}(y)}[(D_Y(y)-1)^2]\\
+\mathbb{E}_{x\sim p_{data}(x)}[D_Y(G(x)I_p^{foggy})^2],
\end{multline}
where $x$ and $y$ are foggy image and clean image, $I_p^{foggy}$ and $I_p^{clean}$ are the corresponding prior maps, and $||\cdot||_1$ is an $L_1$ norm. In $L_{pgcyc}$ and $L_{pgG}$, the loss of objects is increased and the loss of sky is reduced, as a result, the network is guided to recover the objects.

\subsection{Network Architecture}
The details of the defogging network $G$ are shown in Fig. 5. We employ an encoder-decoder structure that includes an encoder for downscaling, a feature transformation module for enhancing the representation capability, and a decoder that consists of two UIMs for upscaling the feature maps while reducing the loss of details. In the forward propagation phase, the network learns a residual from the input image and enhance the defogging effect through the coarse-to-fine framework.

In the encoder, the first layer has 64 filters with a kernel size 7$\times$7 and a stride of 1, the second and the third layers have 128 and 256 filters, respectively, with a kernel size of 3$\times$3 and a stride of 2. The feature transformation module contains 9 residual blocks \cite{he2016deep} to enhance the representation capability of the network. The decoder cascades two upscaling inception modules and two convolutional layers. The first convolutional layer in the decoder has 64 filters with a kernel size 3$\times$3 and a stride of 1, the second convolutional layer has 3 filters, each with a kernel size of 7$\times$7 and a stride of 1. It should be noted that the architecture of the fogging net $F$ is the same as $G$.

\textit{1) Upscaling Inception Module:} As the name suggests, the UIM is inspired by the Inception module \cite{szegedy2015going}, it is a combination of three parallel upscaling branches: deconvolution, bilinear interpolation, and PixelShuffle \cite{shi2016real}, with their output filter banks concatenated into a single output. As shown in Fig. 5, the 1$\times$1 convolution is applied before the deconvolution and bilinear interpolation, which offers a channel-wise pooling to decrease the number of feature maps whilst retaining their salient features. Due to the bilinear interpolation being a fixed upscaling method, we cascade two 3$\times$3 convolutional layers after it to enhance its representation ability, while the other two branches cascade one 3$\times$3 convolutional layer after the upscaling layer. In the two UIMs in the decoder, the number of feature maps of the final 3$\times$3 convolutional layer of the deconvolution, bilinear interpolation, and PixelShuffle \cite{shi2016real} are (64, 32, 32) and (32, 16, 16) respectively. After obtaining all the upscaled feature maps, we adopt the Squeeze-and-Excitation (SE) layer \cite{hu2018squeeze}  to selectively emphasize informative features and suppress less useful ones, which consists of a global average pooling layer to generate channel-wise statistics and a bottleneck layer with two fully connected (FC) layers.

\textit{2) Long-range Residual Coarse-to-fine framework:} The two complementary mechanisms: namely, the long-range residual connection and coarse-to-fine framework are adopted in our network for enhancing the defogging performance. As shown in Fig. 5, we first add a long-range residual connection between the input foggy image and the final output layer, which enables the network to preserve detailed information by learning the residual from the input foggy image. Nonetheless, this will also limit the representation capability of the network. Thus, we extend the defogging procedure into three stages, namely, the coarse, finer, and finest stage. As shown in Fig. 6, except that the coarse stage takes the same two 1/4 downscaled foggy images as input, in the finer and finest stage, respectively, the 1/2 and full-size defogged images generated by the previous stage are integrated as additional information to enhance the defogging performance. As a consequence of these two mutual compensations, the final defogged image retains the detailed information of the input image while also having a desired defogging effect.

\textit{3) Discriminator:} In our network, we have two discriminators. As shown in Fig. 2, $D_X$ is used to distinguish between the input foggy images and the generated foggy images from the generator $F$; in the same way, $D_Y$ is used to discriminate between the generated clean images from $G$ and the input clean images. We adopt PatchGAN \cite{isola2017image} discriminator as $D_X$ which consists of 5 convolutional layers. Each layer has the same kernel size 4$\times$4 with a stride of 2, and the number of filters is 64, 128, 256, 512, 1 from the lowest to highest.

For the discriminator $D_Y$, we employ the pixel level rather than the patch level discriminator, since the size of the prior map is equal to the input image, and the former produces an array of the same size as the input image, while the latter produces an array with reduced size. The discriminator $D_Y$ consists of 3 convolutional layers, where each layer has the same kernel size 1$\times$1 with a stride of 1, and the number of filters is 64, 128, 1 from the lowest to highest. After the output is obtained from $D_Y$, it is further weighted by the prior map with a high penalty for the objects.

\subsection{Loss Function}
The loss function of PG-CycleGAN includes 4 items, i.e., the prior map guided cycle-consistency loss $L_{pgcyc}$, the prior map guided GAN loss $L_{pgG}$, the cycle-consistency loss $L_{cyc}$, and the perceptual loss $L_{VGG}$.

\textbf{Prior map guided cycle-consistency loss} and \textbf{prior map guided GAN loss} are defined in Equations (5) and (6), they give higher penalties to objects on the water and lower penalties to the sky.



\textbf{Cycle-consistency loss} aims to minimize the objective between the foggy images $X$ and its reconstructed foggy images $F(G(x))$, which can be expressed as:

\begin{equation}
\label{deqn_ex1a}
L_{cyc}=||F(G(x))-x||_1+||G(F(y))-y||_1.
\end{equation}

\textbf{Perceptual loss} is also used to enhance the textural information of the recovery image, which is based on VGG16 \cite{simonyan2014very} to further constrain the generators, defined as:
\begin{multline}
\label{deqn_ex1a}
L_{VGG}=||\phi(x)-\phi(F(G(x)))||_2^2\\+||\phi(y)-\phi(G(F(y)))||_2^2,
\end{multline}
where $\phi$ represents the feature maps of the second and fifth pooling layer of the VGG16 network, and $||\cdot||_2$ is an $L_2$ norm.

\textbf{Overall loss function.} Combining all the above losses, the whole loss function is formulated as,
\begin{equation}
\label{deqn_ex1a}
L=\lambda_1 (L_{cyc}+ L_{pgcyc}+ L_{VGG})+\lambda_2 L_{pgG}.
\end{equation}
where $\lambda_i$ ($i=1,2$) is a hyperparameter indicating the weight. In our empirical experiments, it is found that the network is sensitive to $L_{pgG}$ which can significantly improve the defogging performance after fine-tuning. Thus, we set $\lambda_1$ to 1 which can stabilize the training, and set $\lambda_2$ to a value that is slightly larger 2.

\section{Experiments and Discussions}
In this section, we evaluate the qualitative and quantitative results of our proposed method against six state-of-the-art methods, including the prior-based method DCP \cite{he2010single}, the supervised method MSBDN \cite{Dong_2020_CVPR}, the semi-supervised and unsupervised methods, e.g. DAD \cite{shao2020domain}, PSD \cite{chen2021psd}, RefineDNet \cite{zhao2021refinednet}, and SLAD \cite{Liang_2022_IJCAI}.

\begin{figure*}[!t]
\centering
\subfloat{\includegraphics[width=0.123\linewidth]{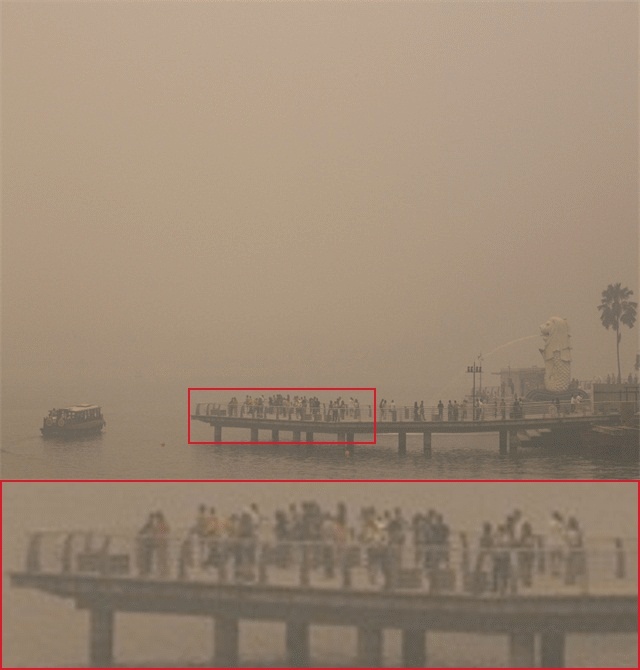}%
\label{}}
\hfil
\subfloat{\includegraphics[width=0.123\linewidth]{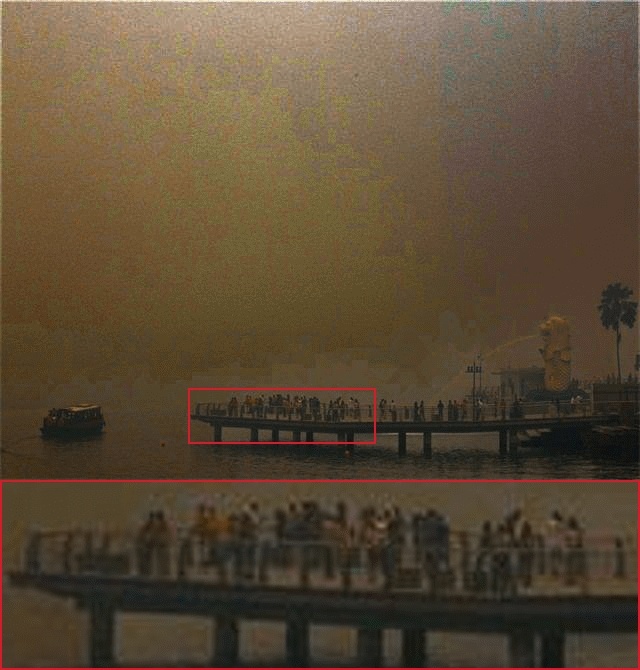}%
\label{}}
\hfil
\subfloat{\includegraphics[width=0.123\linewidth]{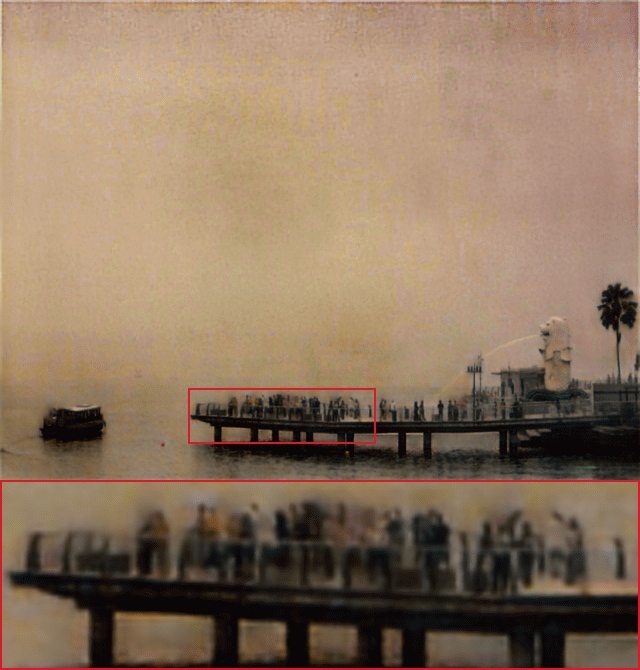}%
\label{}}
\hfil
\subfloat{\includegraphics[width=0.123\linewidth]{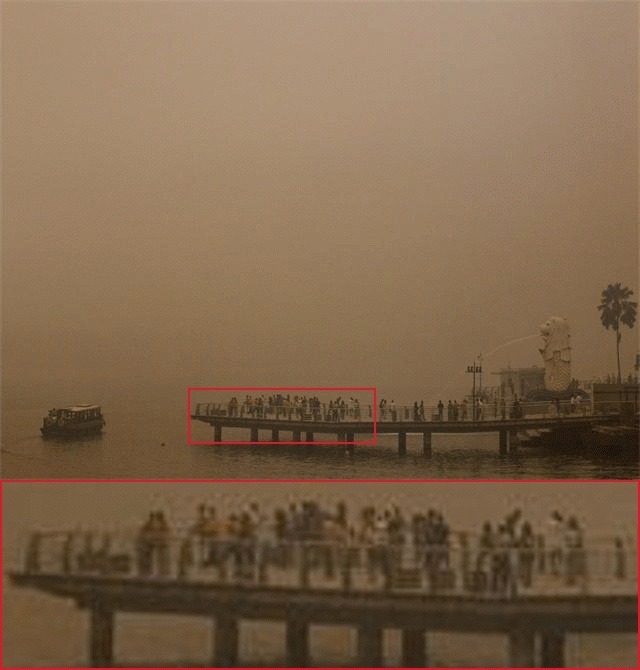}%
\label{}}
\hfil
\subfloat{\includegraphics[width=0.123\linewidth]{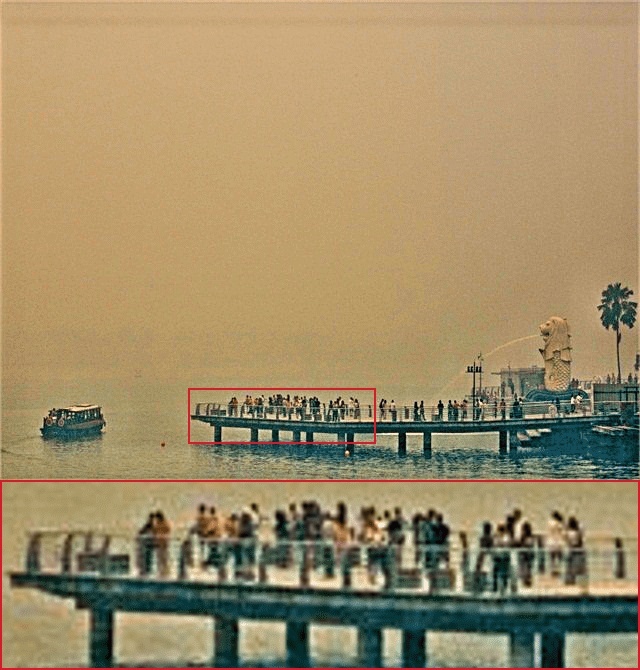}%
\label{}}
\hfil
\subfloat{\includegraphics[width=0.123\linewidth]{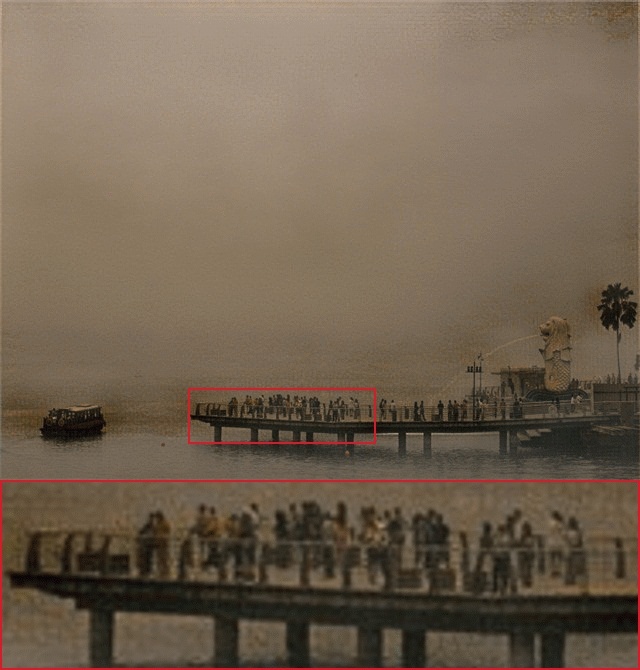}%
\label{}}
\hfil
\subfloat{\includegraphics[width=0.123\linewidth]{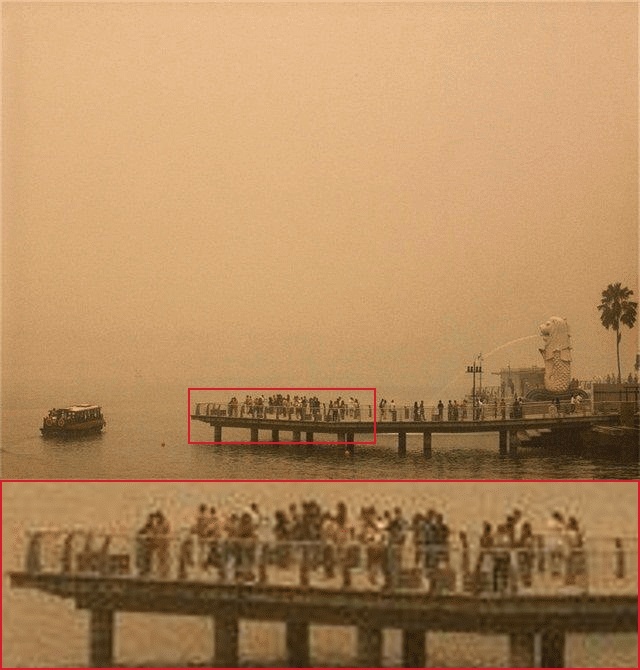}%
\label{}}
\hfil
\subfloat{\includegraphics[width=0.123\linewidth]{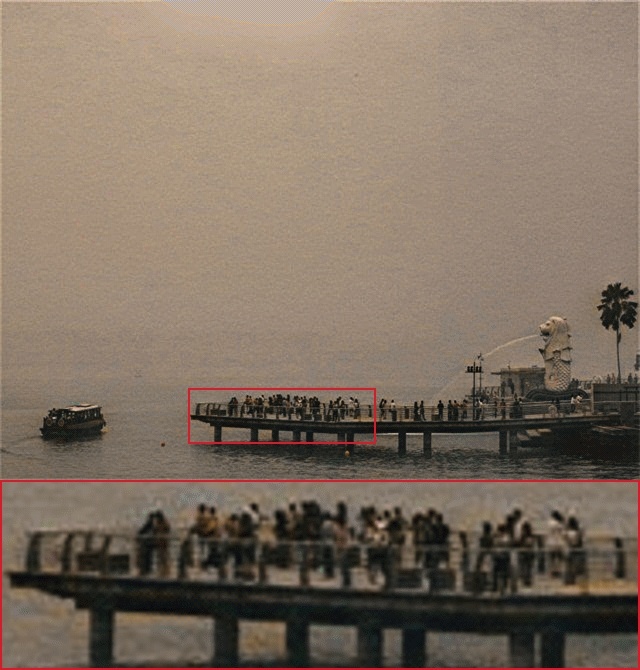}%
\label{}}\\\vspace{-2mm}
\makebox[0.118\linewidth][c]{\scriptsize 2.84 / 0.80}
\makebox[0.118\linewidth][c]{\scriptsize 0.75 / 0.58}
\makebox[0.118\linewidth][c]{\scriptsize 0.93 / 0.78}
\makebox[0.118\linewidth][c]{\scriptsize 2.30 / 0.77}
\makebox[0.118\linewidth][c]{\scriptsize 0.79 / 0.74}
\makebox[0.118\linewidth][c]{\scriptsize 1.23 / 0.64}
\makebox[0.118\linewidth][c]{\scriptsize 0.81 / 0.80}
\makebox[0.118\linewidth][c]{\scriptsize 0.92 / 0.60}\\\vspace{-3mm}
\subfloat{\includegraphics[width=0.123\linewidth]{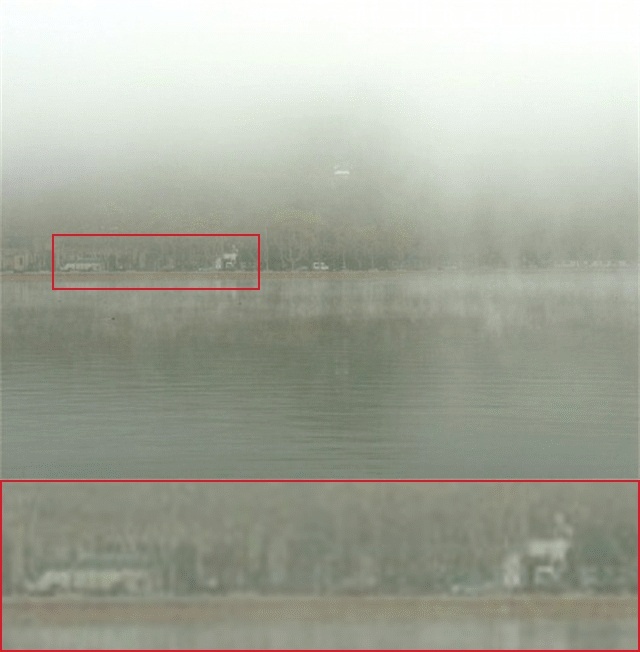}%
\label{}}
\hfil
\subfloat{\includegraphics[width=0.123\linewidth]{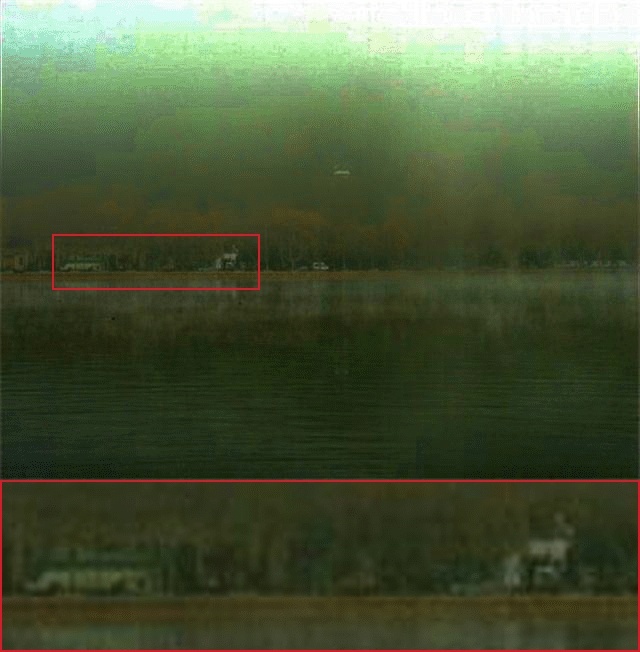}%
\label{}}
\hfil
\subfloat{\includegraphics[width=0.123\linewidth]{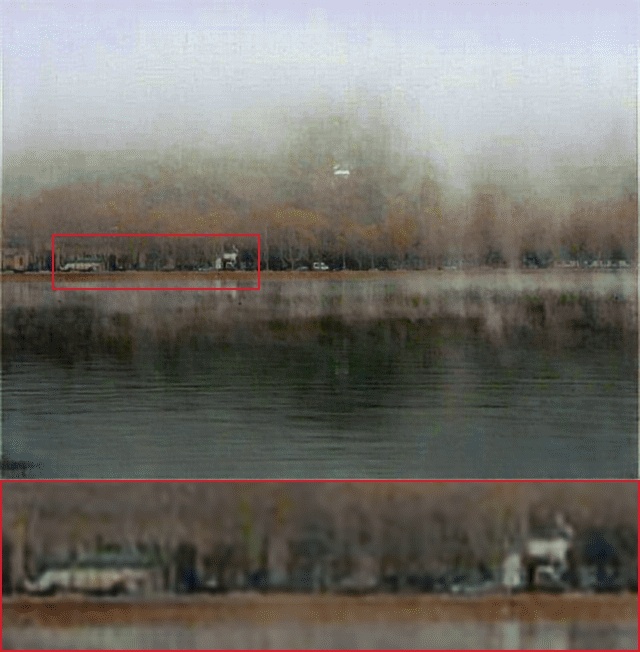}%
\label{}}
\hfil
\subfloat{\includegraphics[width=0.123\linewidth]{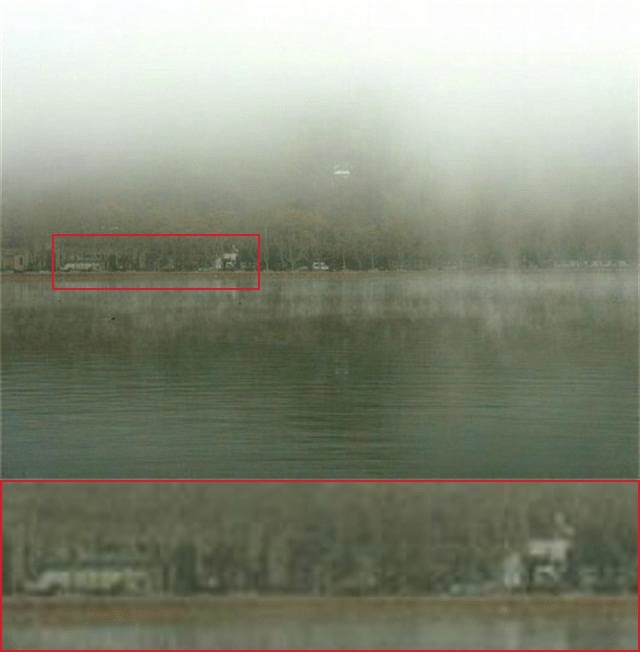}%
\label{}}
\hfil
\subfloat{\includegraphics[width=0.123\linewidth]{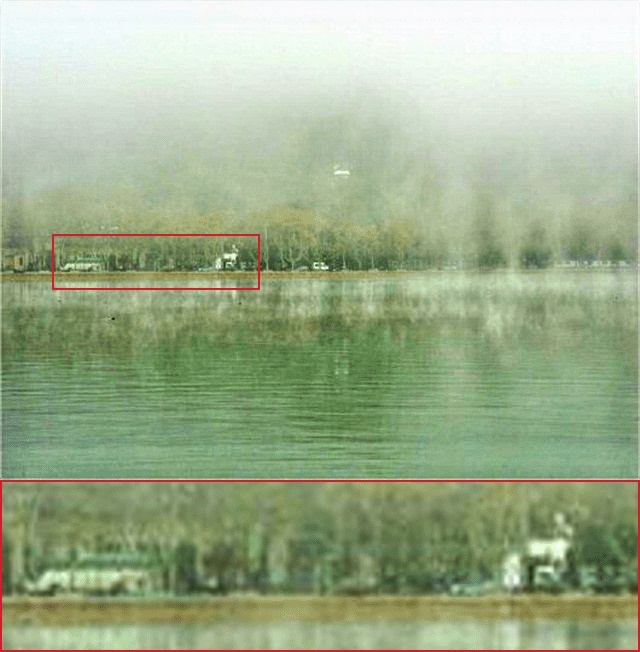}%
\label{}}
\hfil
\subfloat{\includegraphics[width=0.123\linewidth]{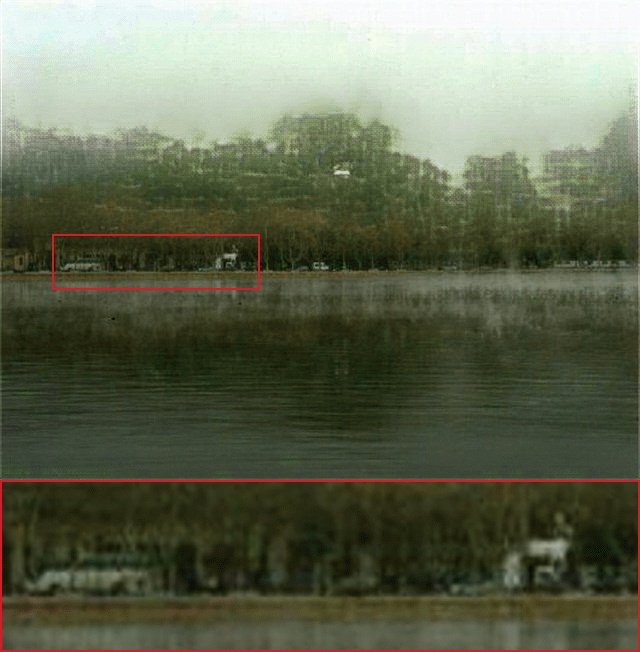}%
\label{}}
\hfil
\subfloat{\includegraphics[width=0.123\linewidth]{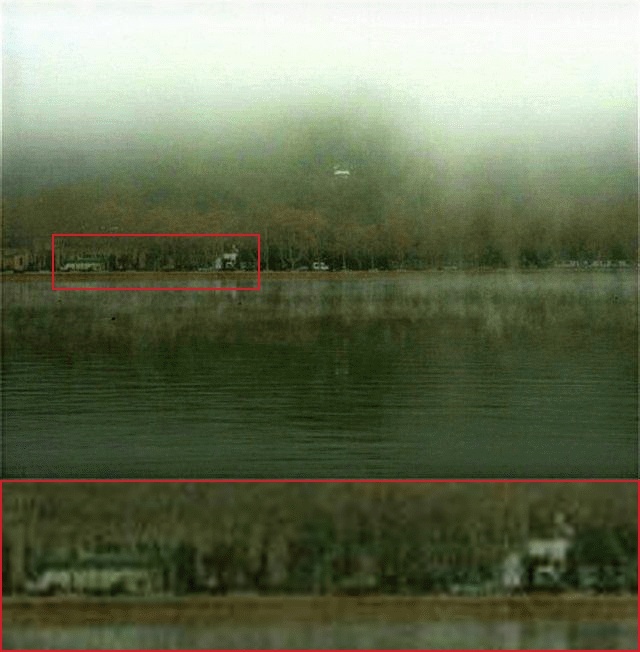}%
\label{}}
\hfil
\subfloat{\includegraphics[width=0.123\linewidth]{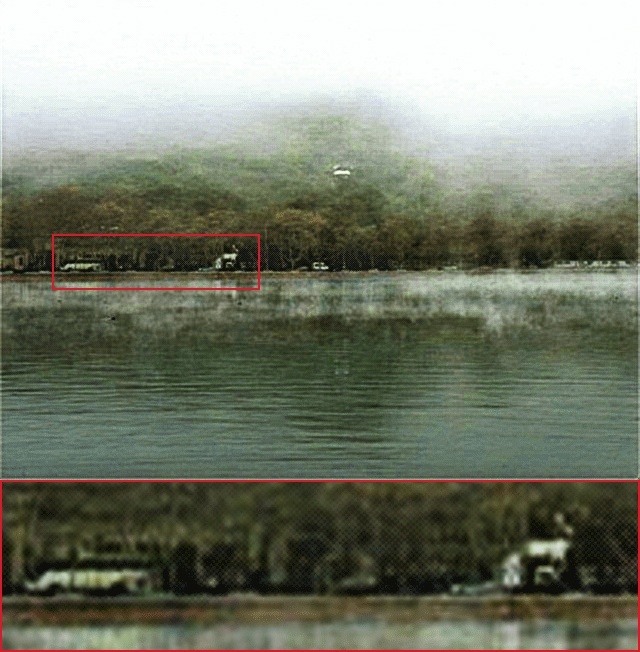}%
\label{}}\\\vspace{-2mm}
\makebox[0.118\linewidth][c]{\scriptsize 4.88 / 0.73}
\makebox[0.118\linewidth][c]{\scriptsize 0.86 / 0.34}
\makebox[0.118\linewidth][c]{\scriptsize 1.51 / 0.48}
\makebox[0.118\linewidth][c]{\scriptsize 2.85 / 0.61}
\makebox[0.118\linewidth][c]{\scriptsize 1.48 / 0.49}
\makebox[0.118\linewidth][c]{\scriptsize 0.88 / 0.35}
\makebox[0.118\linewidth][c]{\scriptsize 0.96 / 0.41}
\makebox[0.118\linewidth][c]{\scriptsize 0.83 / 0.33}\\\vspace{-3mm}
\subfloat{\includegraphics[width=0.123\linewidth]{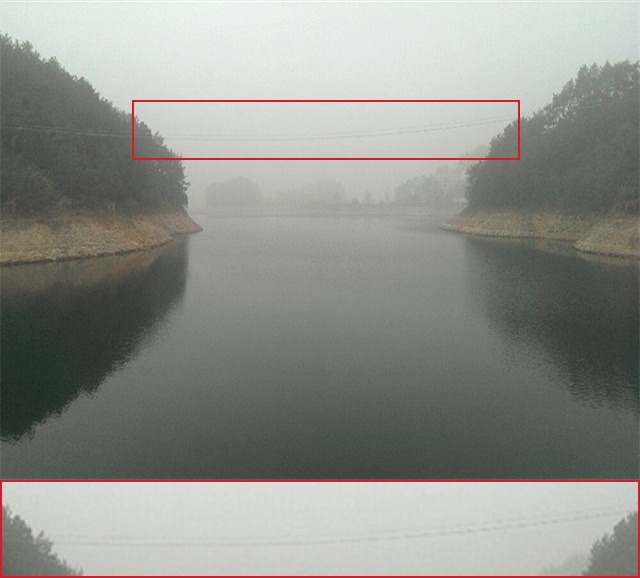}%
\label{}}
\hfil
\subfloat{\includegraphics[width=0.123\linewidth]{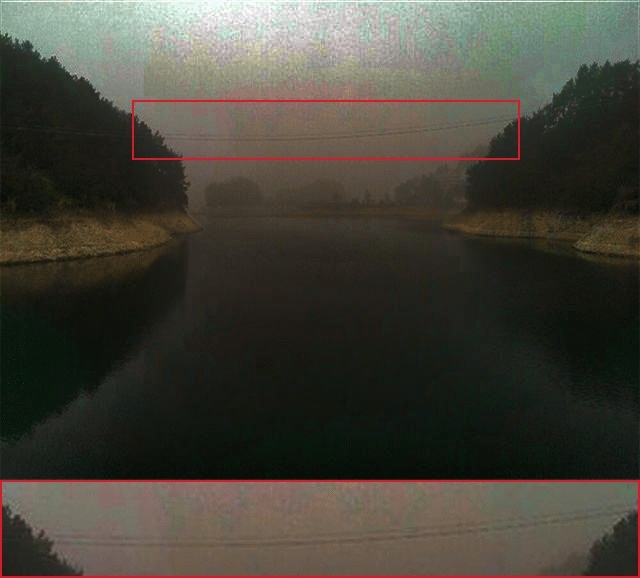}%
\label{}}
\hfil
\subfloat{\includegraphics[width=0.123\linewidth]{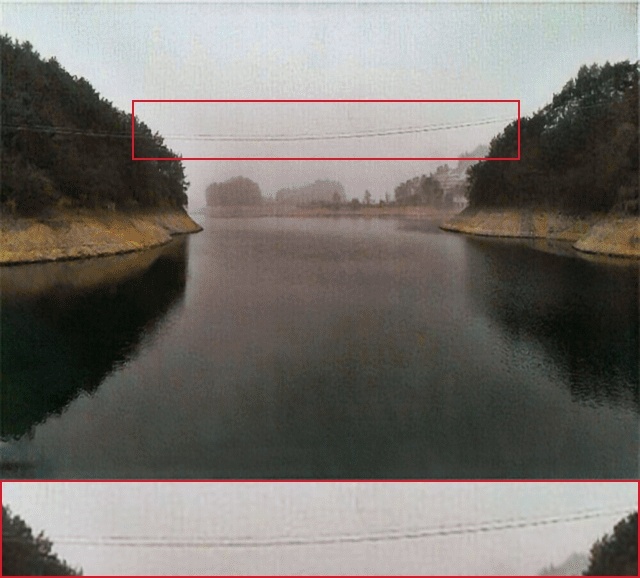}%
\label{}}
\hfil
\subfloat{\includegraphics[width=0.123\linewidth]{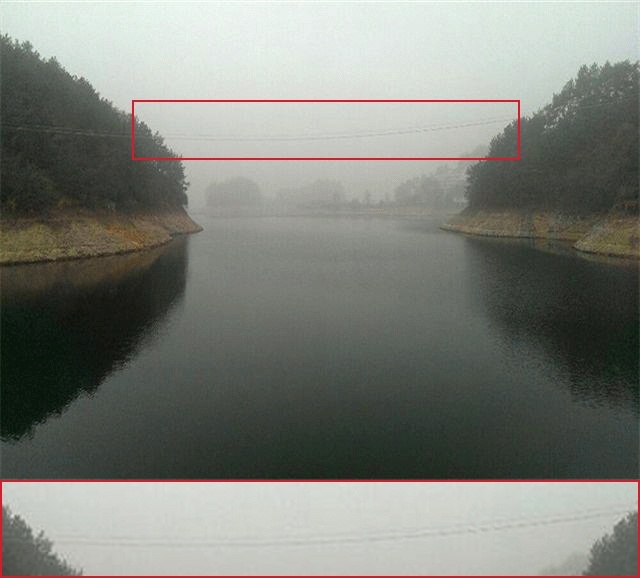}%
\label{}}
\hfil
\subfloat{\includegraphics[width=0.123\linewidth]{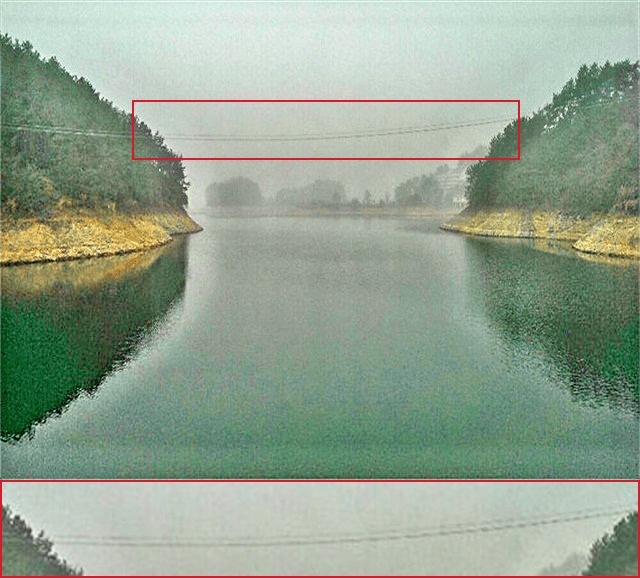}%
\label{}}
\hfil
\subfloat{\includegraphics[width=0.123\linewidth]{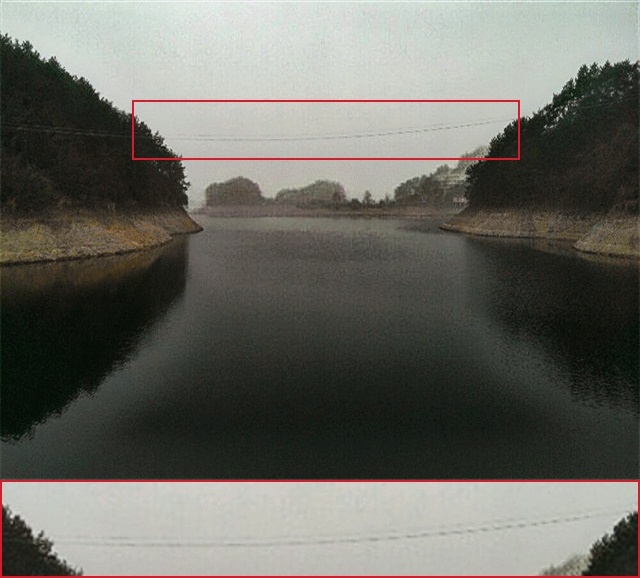}%
\label{}}
\hfil
\subfloat{\includegraphics[width=0.123\linewidth]{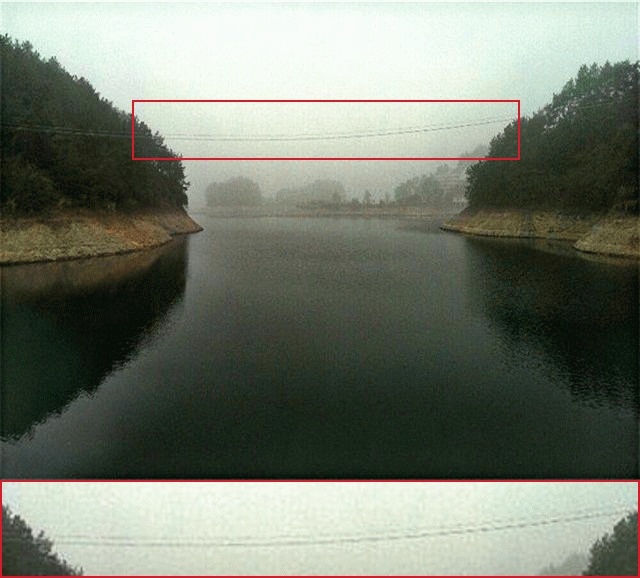}%
\label{}}
\hfil
\subfloat{\includegraphics[width=0.123\linewidth]{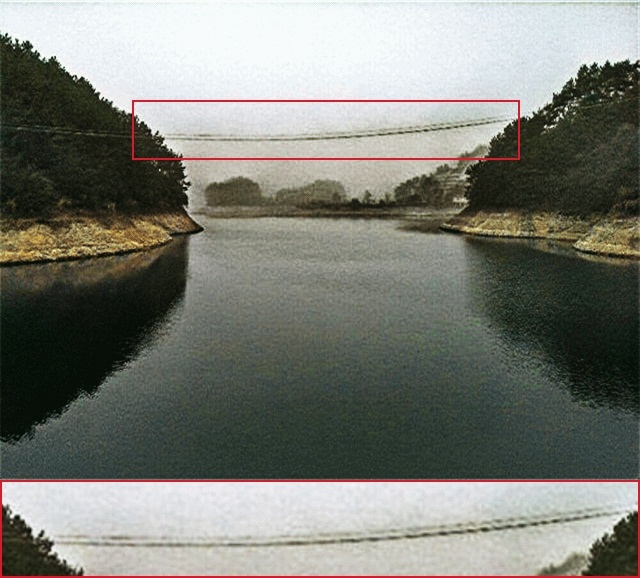}%
\label{}}\\\vspace{-2mm}
\makebox[0.118\linewidth][c]{\scriptsize 2.90 / 0.56}
\makebox[0.118\linewidth][c]{\scriptsize 1.31 / 0.35}
\makebox[0.118\linewidth][c]{\scriptsize 1.13 / 0.45}
\makebox[0.118\linewidth][c]{\scriptsize 1.96 / 0.49}
\makebox[0.118\linewidth][c]{\scriptsize 0.82 / 0.33}
\makebox[0.118\linewidth][c]{\scriptsize 0.98 / 0.37}
\makebox[0.118\linewidth][c]{\scriptsize 0.86 / 0.35}
\makebox[0.118\linewidth][c]{\scriptsize 0.54 / 0.26}\\\vspace{-3mm}
\subfloat{\includegraphics[width=0.123\linewidth]{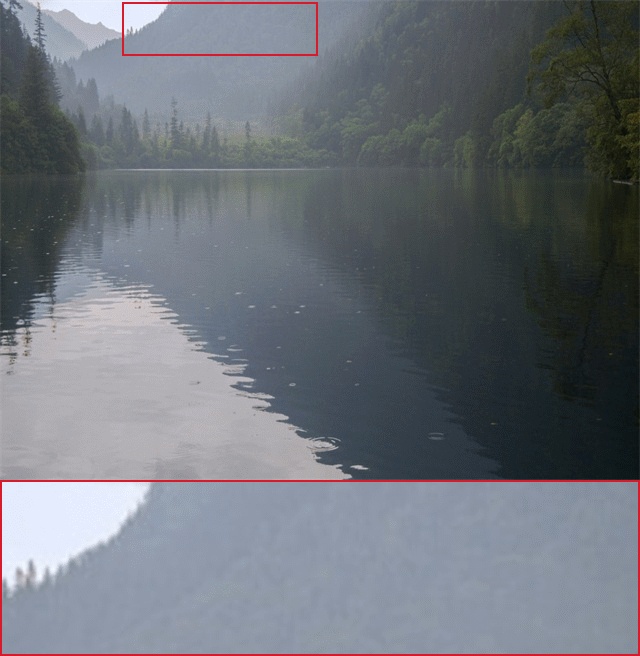}%
\label{}}
\hfil
\subfloat{\includegraphics[width=0.123\linewidth]{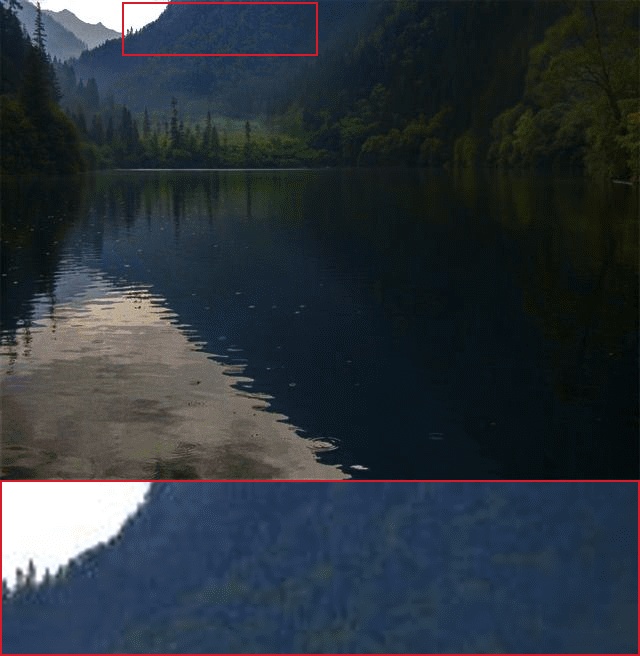}%
\label{}}
\hfil
\subfloat{\includegraphics[width=0.123\linewidth]{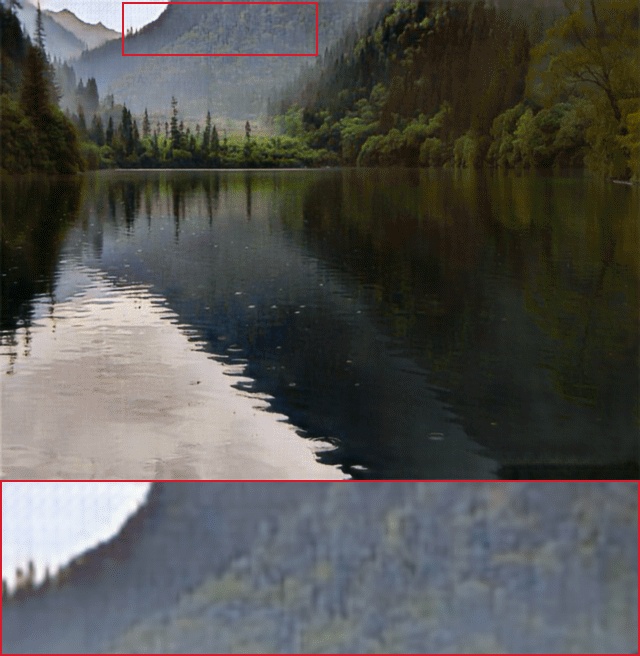}%
\label{}}
\hfil
\subfloat{\includegraphics[width=0.123\linewidth]{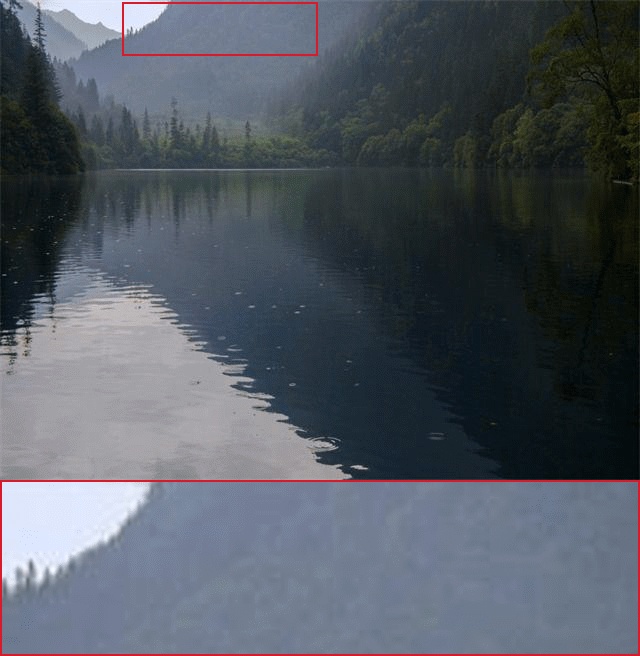}%
\label{}}
\hfil
\subfloat{\includegraphics[width=0.123\linewidth]{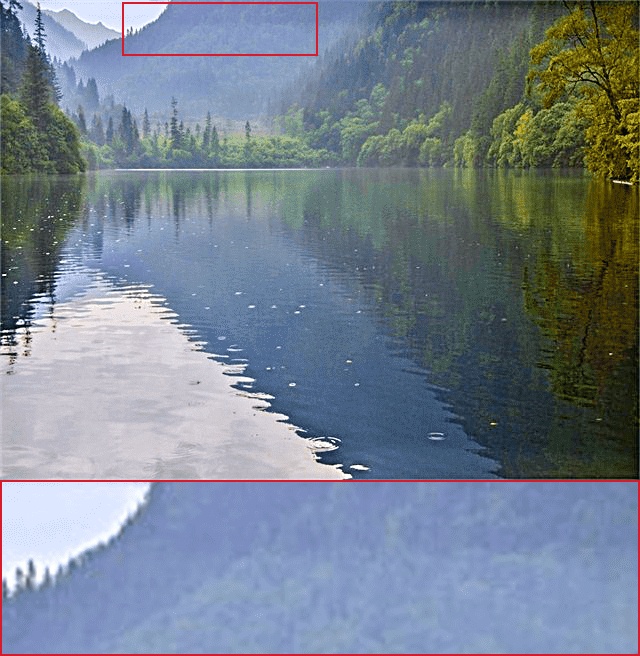}%
\label{}}
\hfil
\subfloat{\includegraphics[width=0.123\linewidth]{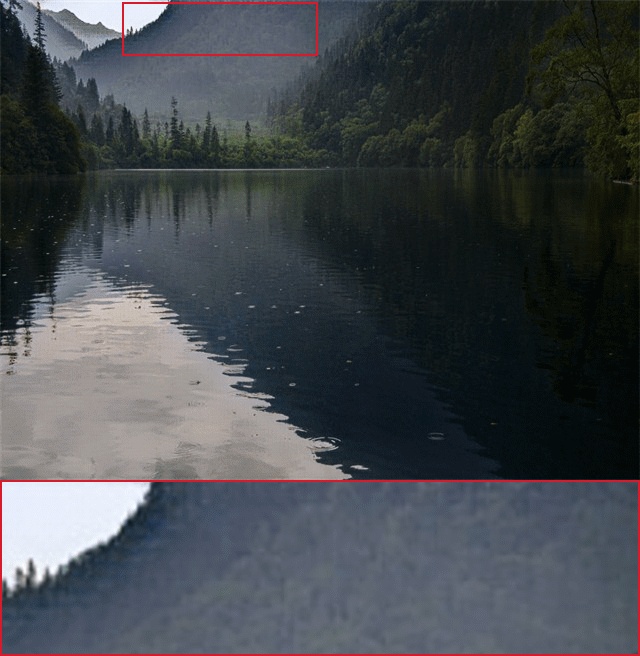}%
\label{}}
\hfil
\subfloat{\includegraphics[width=0.123\linewidth]{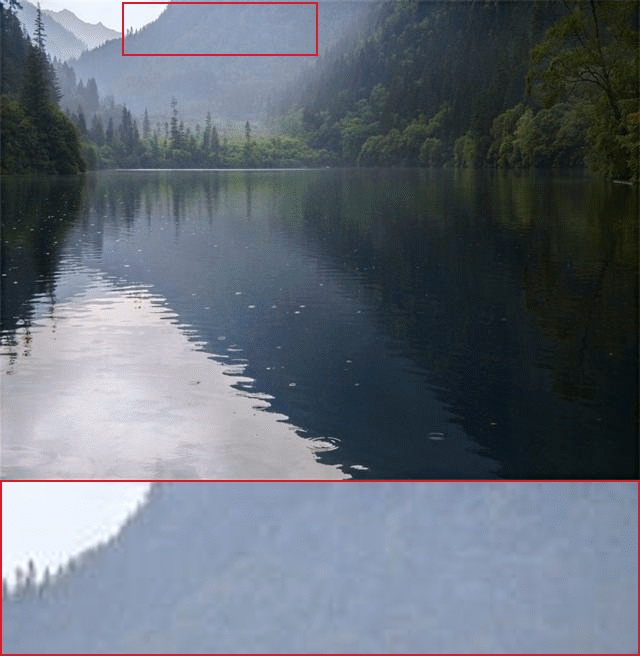}%
\label{}}
\hfil
\subfloat{\includegraphics[width=0.123\linewidth]{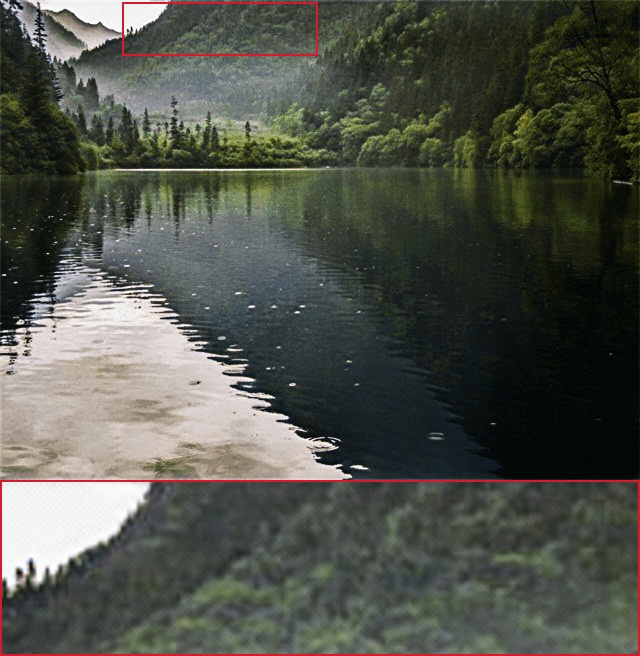}%
\label{}}\\\vspace{-2mm}
\makebox[0.118\linewidth][c]{\scriptsize 1.66 / 0.41}
\makebox[0.118\linewidth][c]{\scriptsize 0.91 / 0.24}
\makebox[0.118\linewidth][c]{\scriptsize 0.65 / 0.35}
\makebox[0.118\linewidth][c]{\scriptsize 0.98 / 0.30}
\makebox[0.118\linewidth][c]{\scriptsize 0.48 / 0.22}
\makebox[0.118\linewidth][c]{\scriptsize 0.60 / 0.24}
\makebox[0.118\linewidth][c]{\scriptsize 0.92 / 0.30}
\makebox[0.118\linewidth][c]{\scriptsize 0.36 / 0.16}\\\vspace{-3mm}
\subfloat{\includegraphics[width=0.123\linewidth]{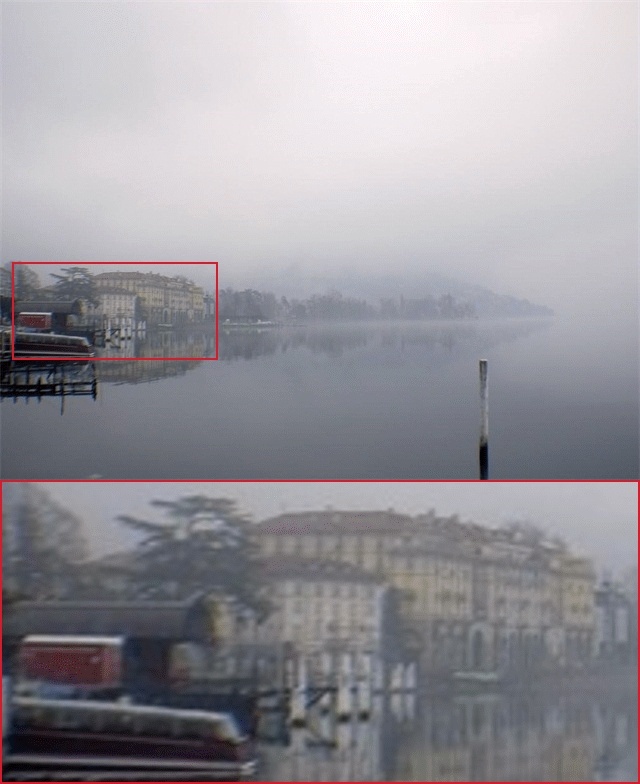}%
\label{}}
\hfil
\subfloat{\includegraphics[width=0.123\linewidth]{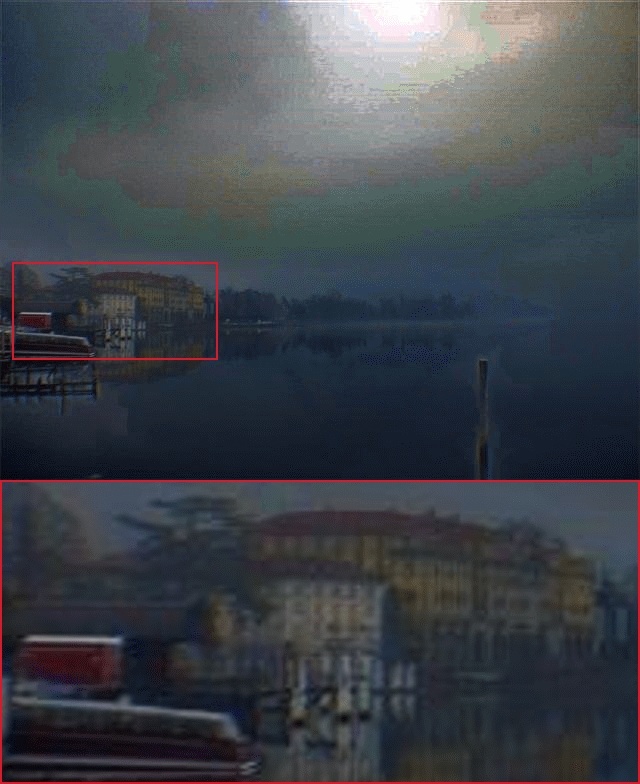}%
\label{}}
\hfil
\subfloat{\includegraphics[width=0.123\linewidth]{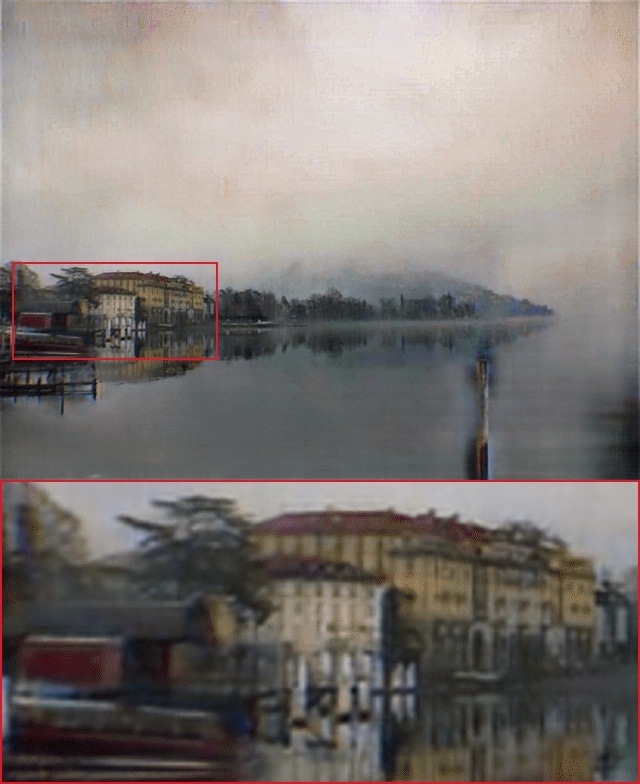}%
\label{}}
\hfil
\subfloat{\includegraphics[width=0.123\linewidth]{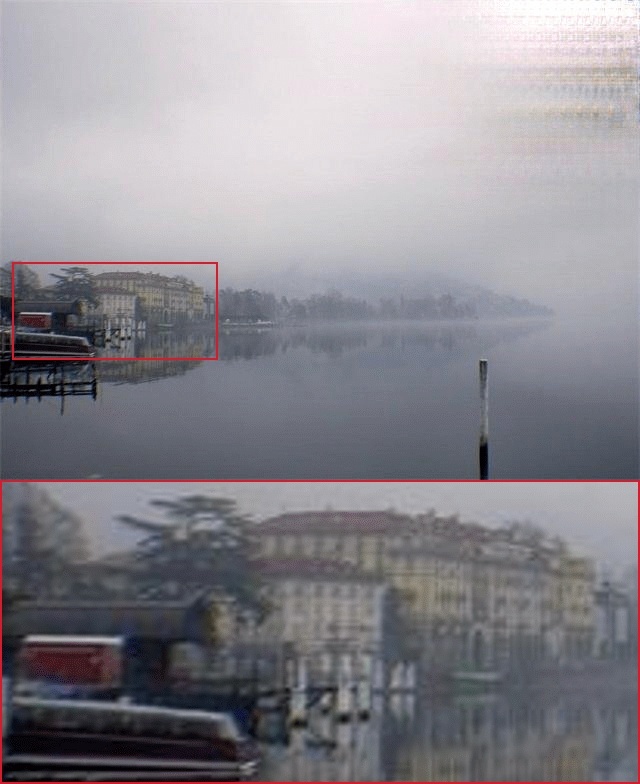}%
\label{}}
\hfil
\subfloat{\includegraphics[width=0.123\linewidth]{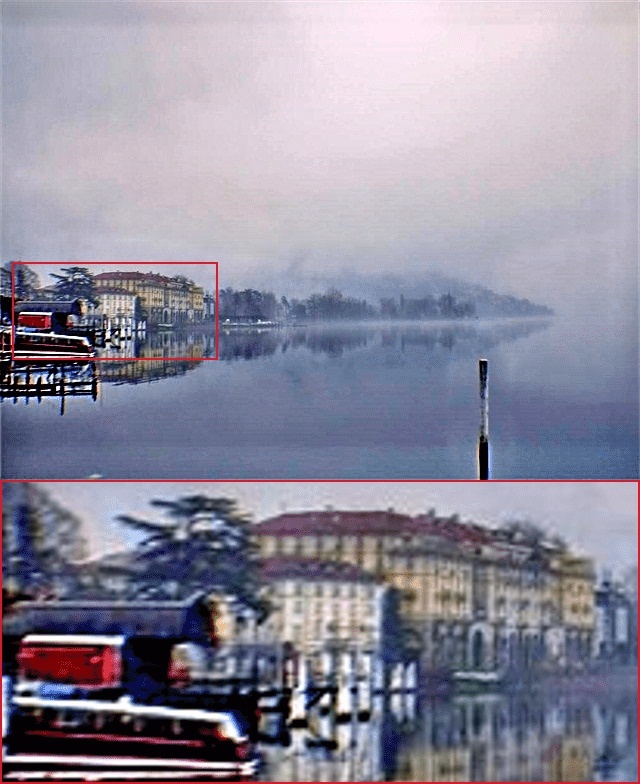}%
\label{}}
\hfil
\subfloat{\includegraphics[width=0.123\linewidth]{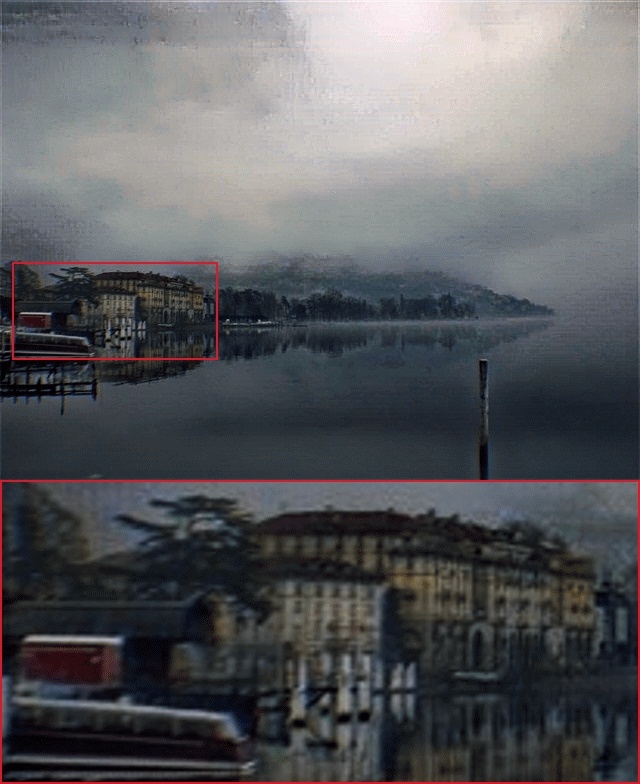}%
\label{}}
\hfil
\subfloat{\includegraphics[width=0.123\linewidth]{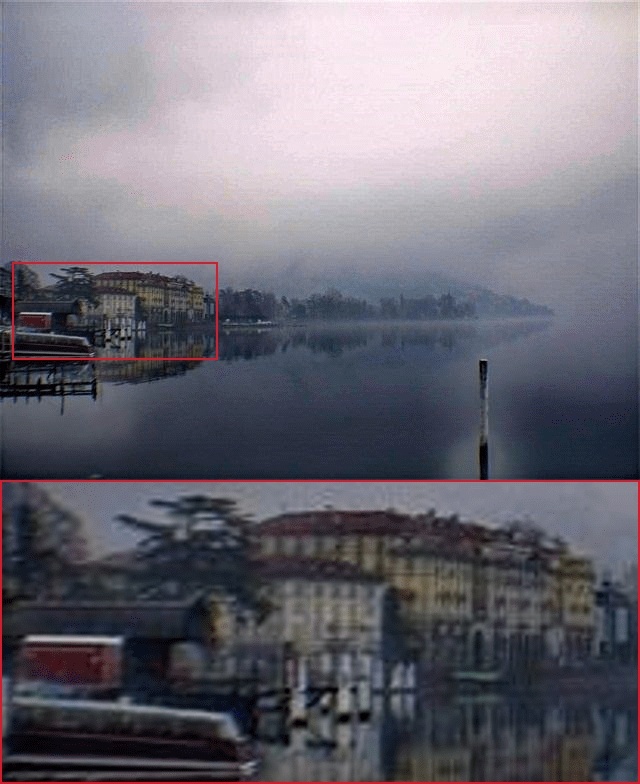}%
\label{}}
\hfil
\subfloat{\includegraphics[width=0.123\linewidth]{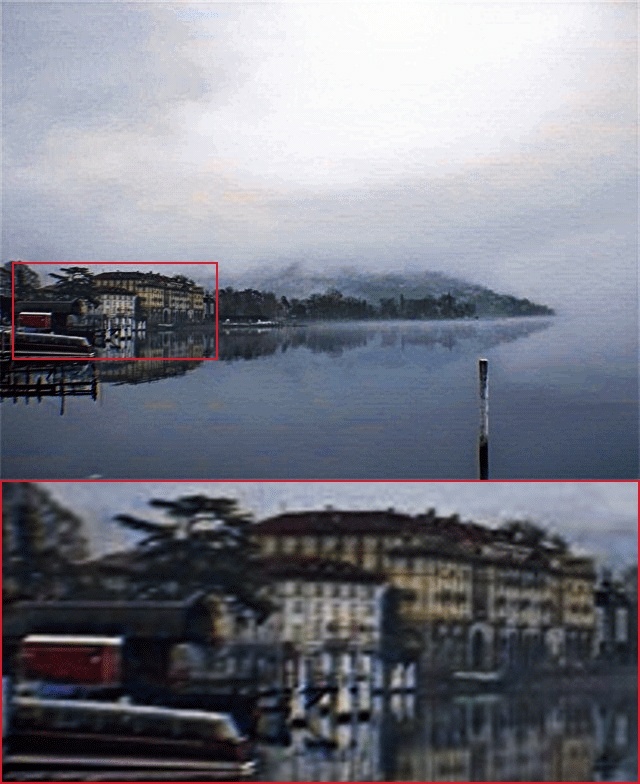}%
\label{}}\\\vspace{-2mm}
\makebox[0.118\linewidth][c]{\scriptsize 4.49 / 0.72}
\makebox[0.118\linewidth][c]{\scriptsize 1.31 / 0.32}
\makebox[0.118\linewidth][c]{\scriptsize 2.38 / 0.64}
\makebox[0.118\linewidth][c]{\scriptsize 3.50 / 0.68}
\makebox[0.118\linewidth][c]{\scriptsize 1.31 / 0.58}
\makebox[0.118\linewidth][c]{\scriptsize 1.28 / 0.44}
\makebox[0.118\linewidth][c]{\scriptsize 1.51 / 0.54}
\makebox[0.118\linewidth][c]{\scriptsize 1.21 / 0.49}\\\vspace{-3mm}
\subfloat{\includegraphics[width=0.123\linewidth]{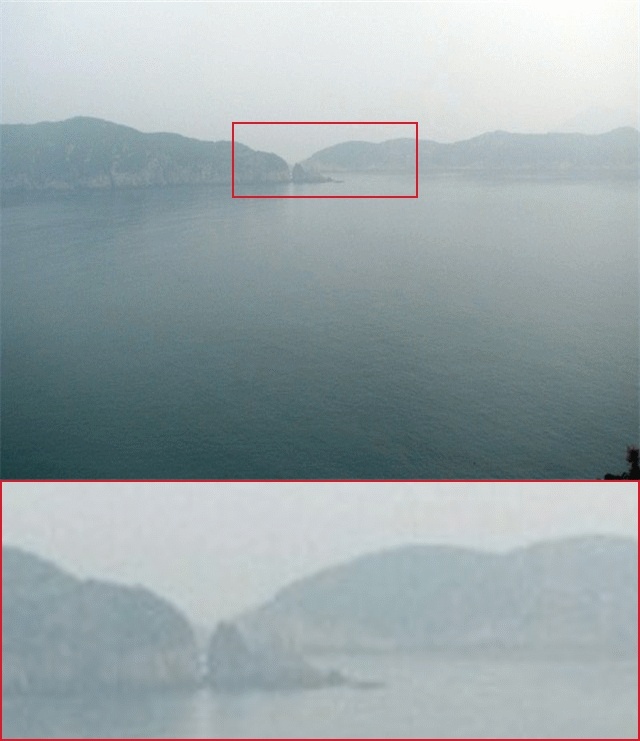}%
\label{}}
\hfil
\subfloat{\includegraphics[width=0.123\linewidth]{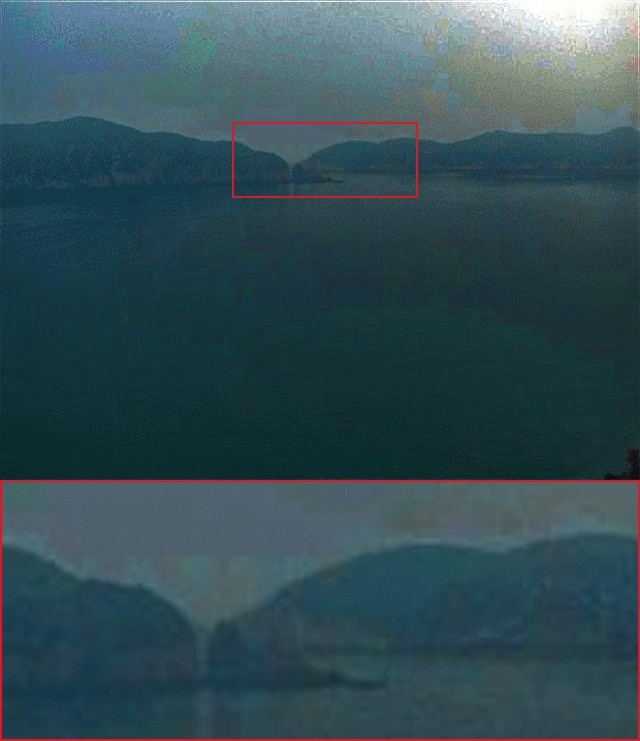}%
\label{}}
\hfil
\subfloat{\includegraphics[width=0.123\linewidth]{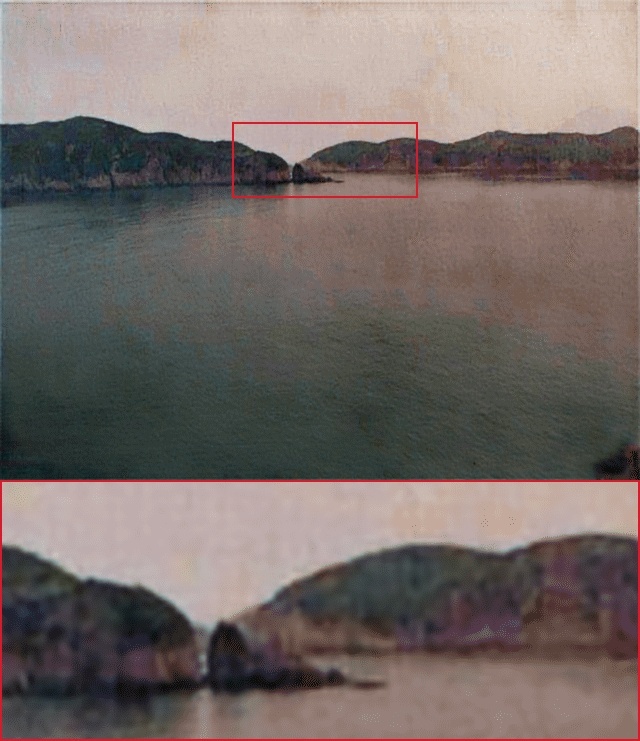}%
\label{}}
\hfil
\subfloat{\includegraphics[width=0.123\linewidth]{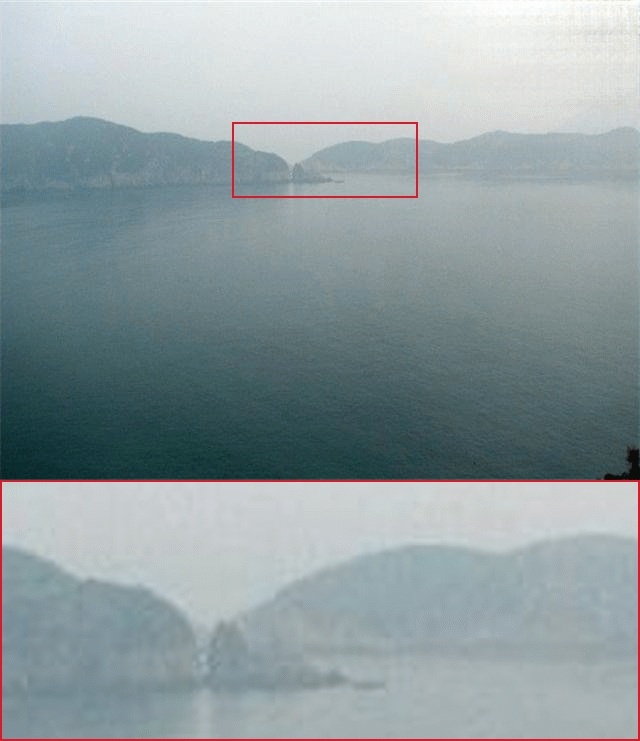}%
\label{}}
\hfil
\subfloat{\includegraphics[width=0.123\linewidth]{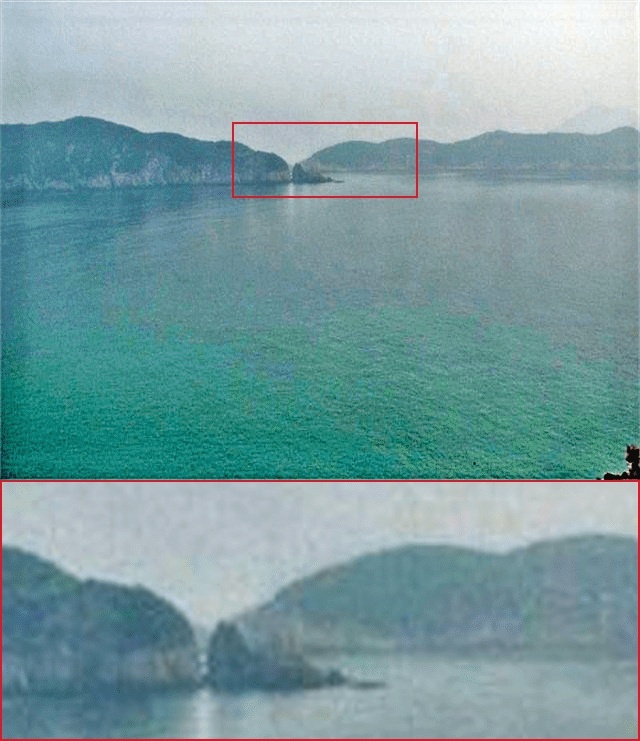}%
\label{}}
\hfil
\subfloat{\includegraphics[width=0.123\linewidth]{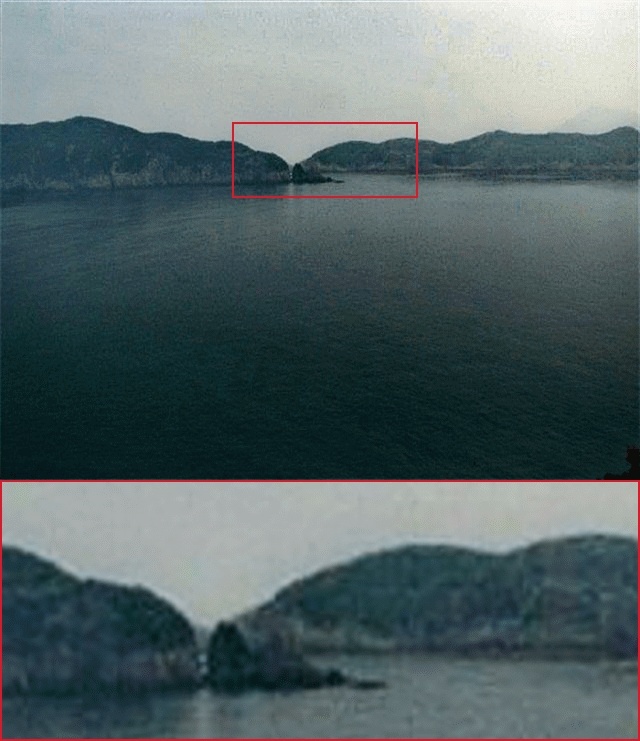}%
\label{}}
\hfil
\subfloat{\includegraphics[width=0.123\linewidth]{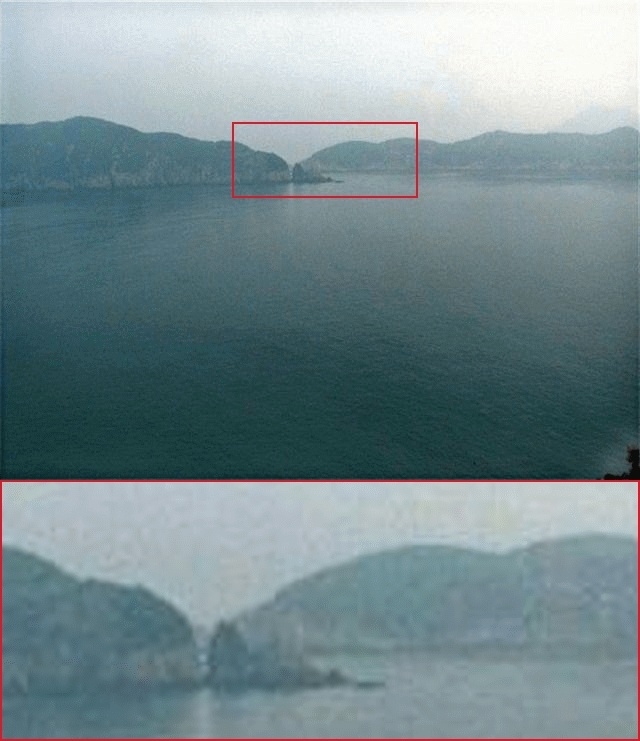}%
\label{}}
\hfil
\subfloat{\includegraphics[width=0.123\linewidth]{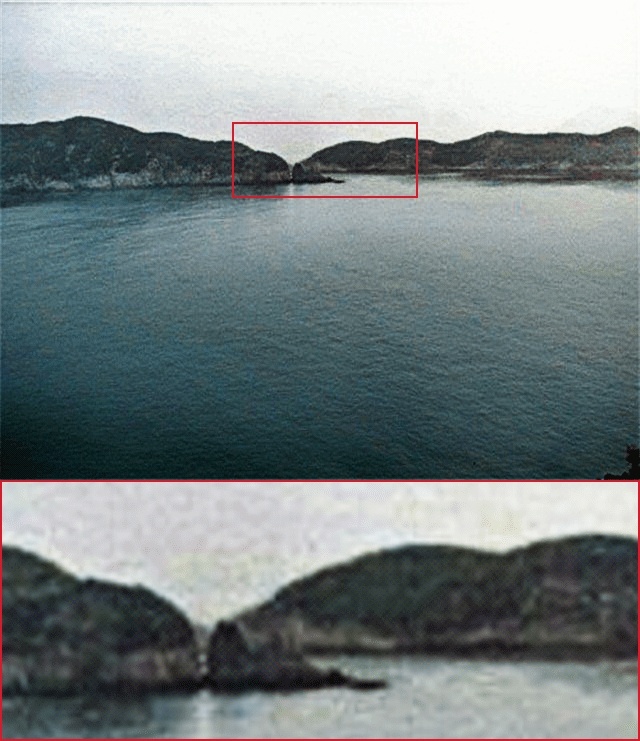}%
\label{}}\\\vspace{-2mm}
\makebox[0.118\linewidth][c]{\scriptsize 2.99 / 0.57}
\makebox[0.118\linewidth][c]{\scriptsize 1.11 / 0.23}
\makebox[0.118\linewidth][c]{\scriptsize 1.31 / 0.50}
\makebox[0.118\linewidth][c]{\scriptsize 2.17 / 0.52}
\makebox[0.118\linewidth][c]{\scriptsize 0.73 / 0.29}
\makebox[0.118\linewidth][c]{\scriptsize 0.74 / 0.28}
\makebox[0.118\linewidth][c]{\scriptsize 1.16 / 0.37}
\makebox[0.118\linewidth][c]{\scriptsize 0.82 / 0.29}\\
\makebox[0.118\linewidth][c]{\scriptsize (a)}
\makebox[0.118\linewidth][c]{\scriptsize (b)}
\makebox[0.118\linewidth][c]{\scriptsize (c)}
\makebox[0.118\linewidth][c]{\scriptsize (d)}
\makebox[0.118\linewidth][c]{\scriptsize (e)}
\makebox[0.118\linewidth][c]{\scriptsize (f)}
\makebox[0.118\linewidth][c]{\scriptsize (g)}
\makebox[0.118\linewidth][c]{\scriptsize (h)}\vspace{-2mm}
\caption{Qualitative evaluation of defogging performance on real-world overwater images. The corresponding indicators under the images are FADE / HazDesNet. (a) Foggy (b) DCP (c) DAD (d) MSBDN (e) PSD (f) RefineDNet (g) SLAD (h) Ours.}
\label{}
\end{figure*}

\begin{table}[h]
\renewcommand\arraystretch{1.2}
\begin{center}
\caption{Comparison of training set sizes for different defogging methods}
\label{tab1}
\begin{tabular}{ l  c  c c}
\toprule
 & \multicolumn{3}{c}{Samples}\\
 & Image pairs & Unpaired foggy & Unpaired clean \\
\hline
DAD  & 6000 & 1000 & -\\
RefineDNet  & 13,990 & 2903 & 3557\\
MSBDN  & 16,000 & - & - \\
Ours & - & 2090 & 2441 \\
\bottomrule
\end{tabular}
\end{center}
\end{table}

\subsection{Datasets}
In this work, we use unpaired real-world images for training instead of the widely used paired synthetic images. Table \uppercase\expandafter{\romannumeral1} shows the comparison of training set sizes for different defogging methods. Unlike other methods that employ a large number of paired and unpaired images for training, our training set only contains 2,090 unpaired foggy and 2,441 clean overwater images. The used dataset is proposed by Zheng \textit{et al.} \cite{zheng2020overwater}, which also contains 188 real-world overwater foggy images for testing. 

To demonstrate the applicability of PG-CycleGAN in wider contexts, two land image datasets O-HAZE \cite{O-HAZE_2018} and NH-HAZE \cite{NH-Haze_2020} are also employed as test sets. O-HAZE \cite{O-HAZE_2018} contains 45 different outdoor scenes depicting the same visual content recorded in haze-free and hazy conditions, under the same illumination parameters. NH-HAZE \cite{NH-Haze_2020} is a non-homogeneous real dataset with pairs of real hazy and corresponding haze-free images, from 55 outdoor scenes.

\subsection{Implementation Details}
In our network, all the training samples are resized to 512$\times$512, and then randomly cropped to 256$\times$256 for data augmentation. During training, ADAM optimizer \cite{kingma2014adam} is employed for the generators and discriminators with a learning rate of 0.0001 and batch size of 1. We adopt cosine annealing schedule where the initial learning rate is decreased to 0 after 200 epochs according to the cosine function. The network is trained on Pytorch deep learning platform with the acceleration of an Nvidia GeForce RTX 3090 GPU, and it takes about 30 hours for training.

\begin{figure*}[!t]
\centering
\subfloat{\includegraphics[width=0.12\linewidth]{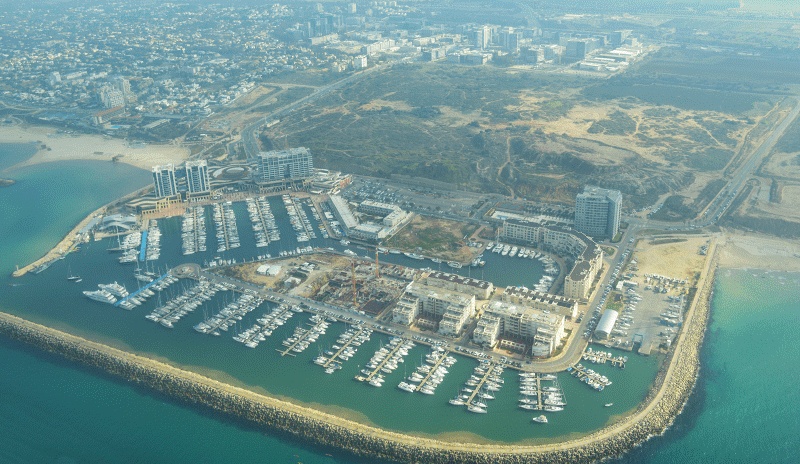}
\label{}}
\subfloat{\includegraphics[width=0.12\linewidth]{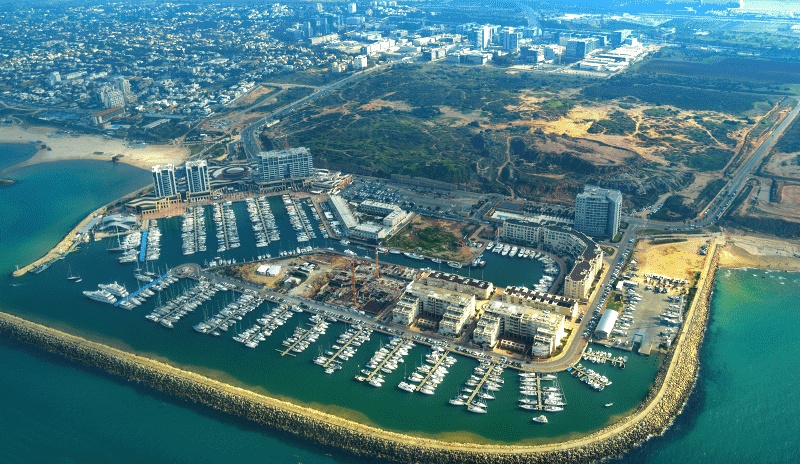}
\label{}}
\subfloat{\includegraphics[width=0.12\linewidth]{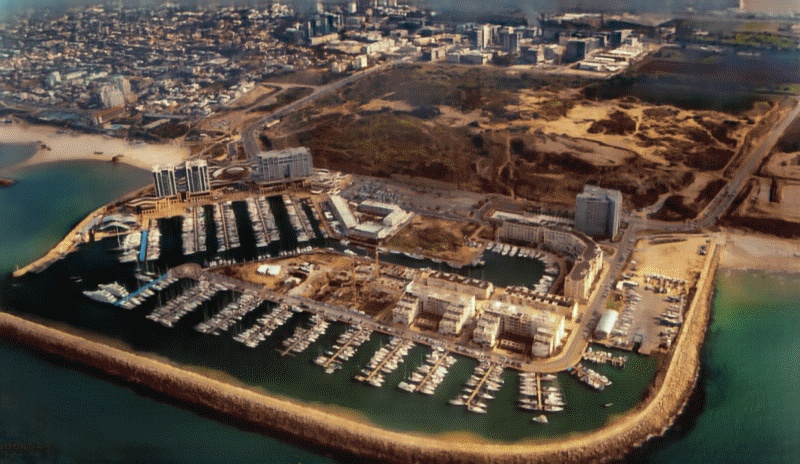}
\label{}}
\subfloat{\includegraphics[width=0.12\linewidth]{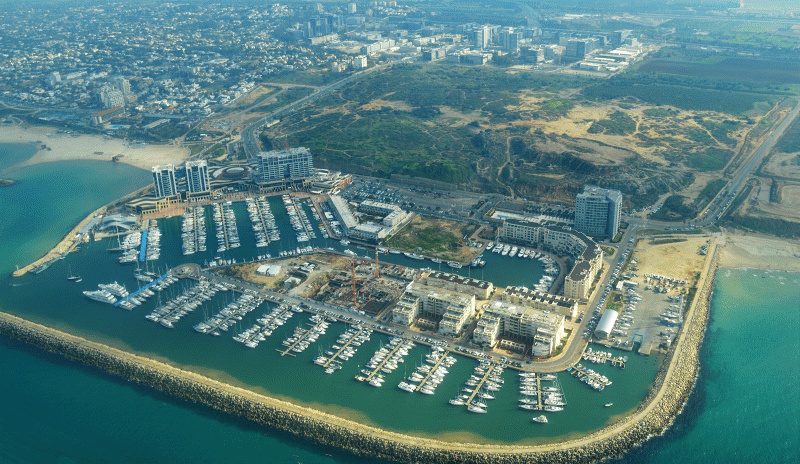}
\label{}}
\subfloat{\includegraphics[width=0.12\linewidth]{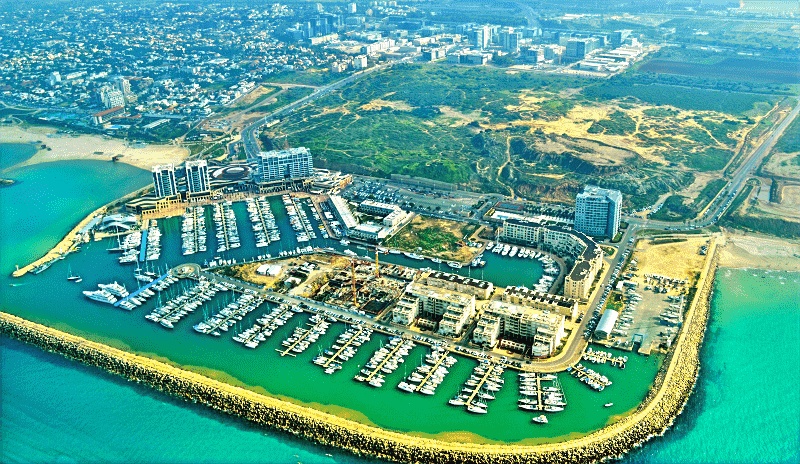}
\label{}}
\subfloat{\includegraphics[width=0.12\linewidth]{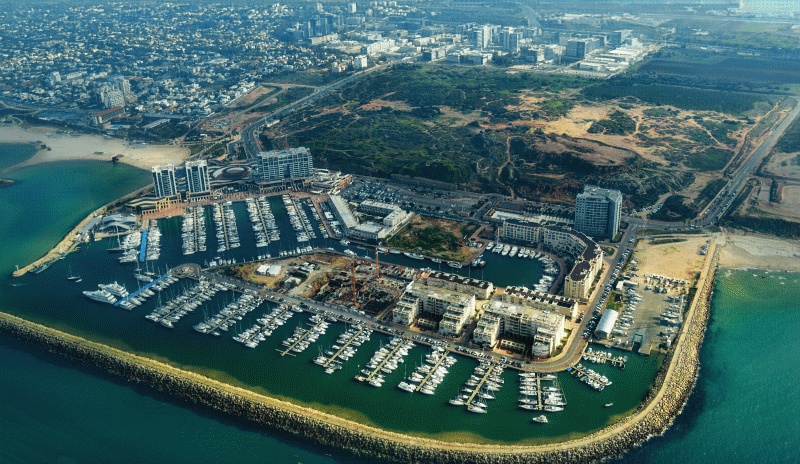}
\label{}}
\subfloat{\includegraphics[width=0.12\linewidth]{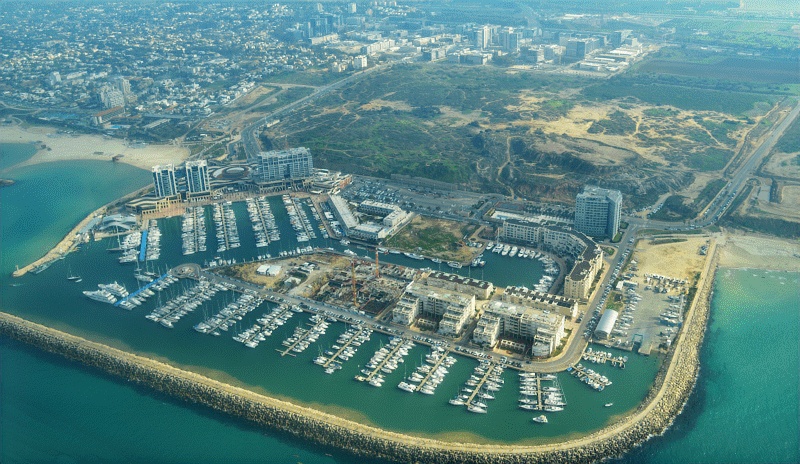}
\label{}}
\subfloat{\includegraphics[width=0.12\linewidth]{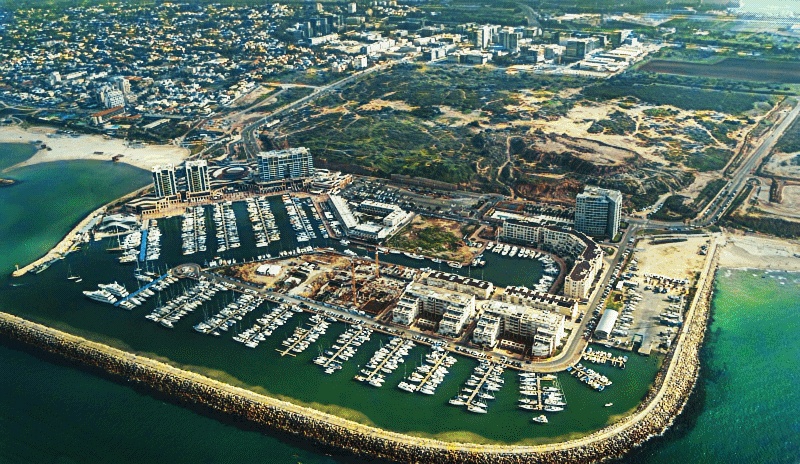}
\label{}}\\\vspace{-2mm}
\makebox[0.118\linewidth][c]{\scriptsize 0.57 / 0.07}
\makebox[0.118\linewidth][c]{\scriptsize 0.24 / 0.01}
\makebox[0.118\linewidth][c]{\scriptsize 0.31 / 0.10}
\makebox[0.118\linewidth][c]{\scriptsize 0.33 / 0.04}
\makebox[0.118\linewidth][c]{\scriptsize 0.17 / 0.01}
\makebox[0.118\linewidth][c]{\scriptsize 0.22 / 0.02}
\makebox[0.118\linewidth][c]{\scriptsize 0.34 / 0.03}
\makebox[0.118\linewidth][c]{\scriptsize 0.18 / 0.02}\\\vspace{-3mm}
\subfloat{\includegraphics[width=0.12\linewidth]{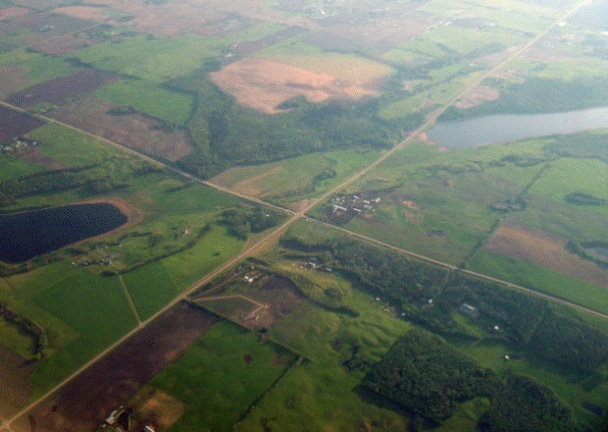}
\label{}}
\subfloat{\includegraphics[width=0.12\linewidth]{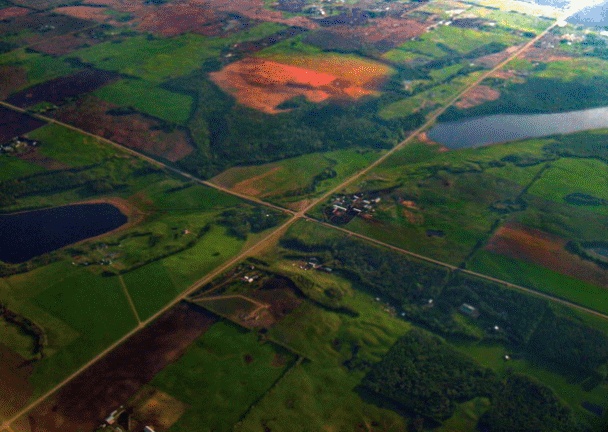}
\label{}}
\subfloat{\includegraphics[width=0.12\linewidth]{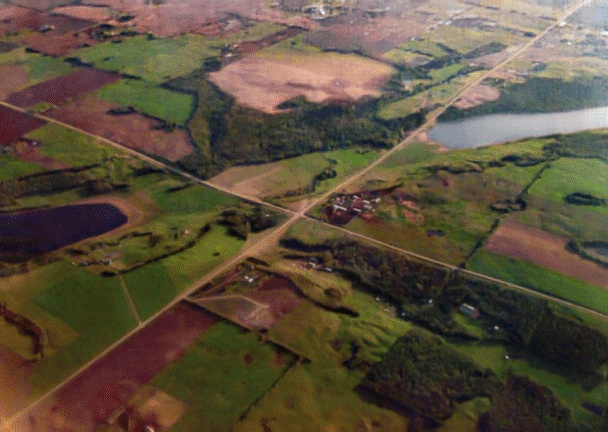}
\label{}}
\subfloat{\includegraphics[width=0.12\linewidth]{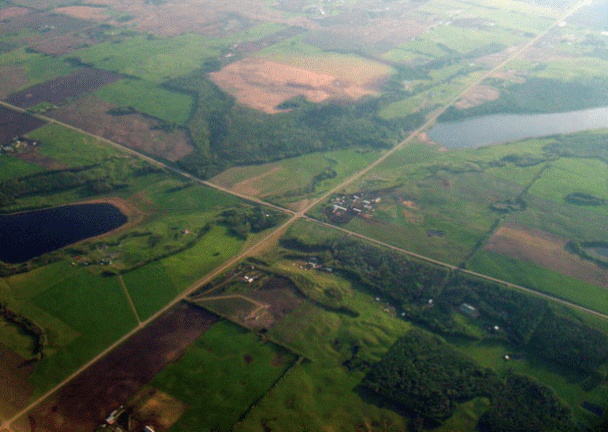}
\label{}}
\subfloat{\includegraphics[width=0.12\linewidth]{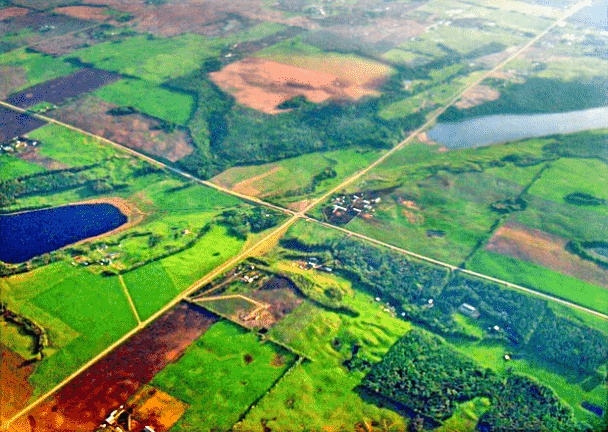}
\label{}}
\subfloat{\includegraphics[width=0.12\linewidth]{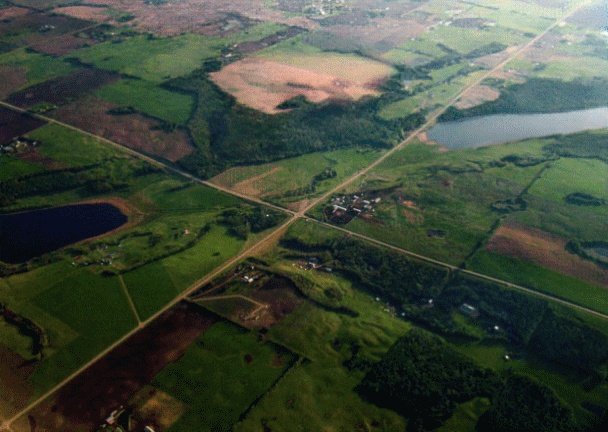}
\label{}}
\subfloat{\includegraphics[width=0.12\linewidth]{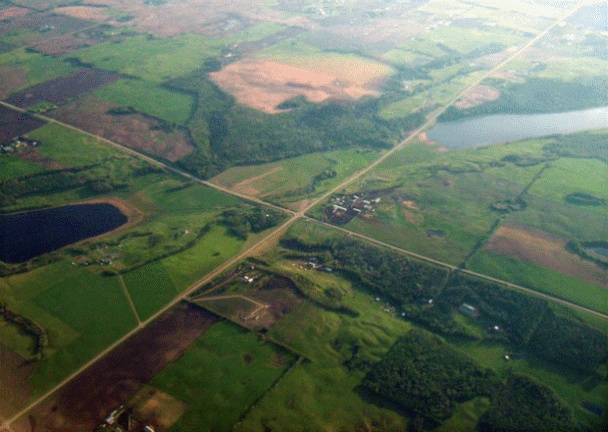}
\label{}}
\subfloat{\includegraphics[width=0.12\linewidth]{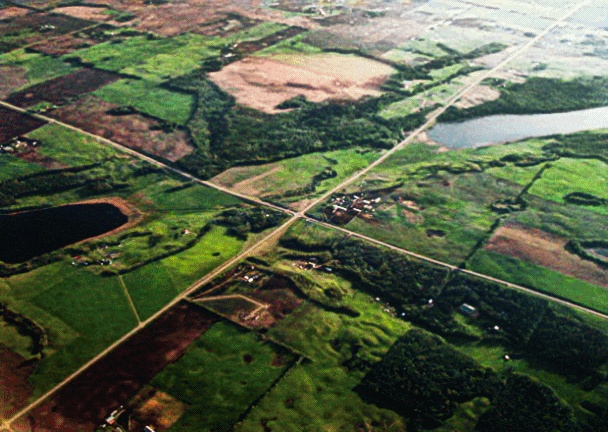}
\label{}}\\\vspace{-2mm}
\makebox[0.118\linewidth][c]{\scriptsize 0.78 / 0.23}
\makebox[0.118\linewidth][c]{\scriptsize 0.32 / 0.10}
\makebox[0.118\linewidth][c]{\scriptsize 0.37 / 0.24}
\makebox[0.118\linewidth][c]{\scriptsize 0.56 / 0.20}
\makebox[0.118\linewidth][c]{\scriptsize 0.21 / 0.05}
\makebox[0.118\linewidth][c]{\scriptsize 0.32 / 0.10}
\makebox[0.118\linewidth][c]{\scriptsize 0.51 / 0.17}
\makebox[0.118\linewidth][c]{\scriptsize 0.20 / 0.06}\\\vspace{-3mm}
\subfloat{\includegraphics[width=0.12\linewidth]{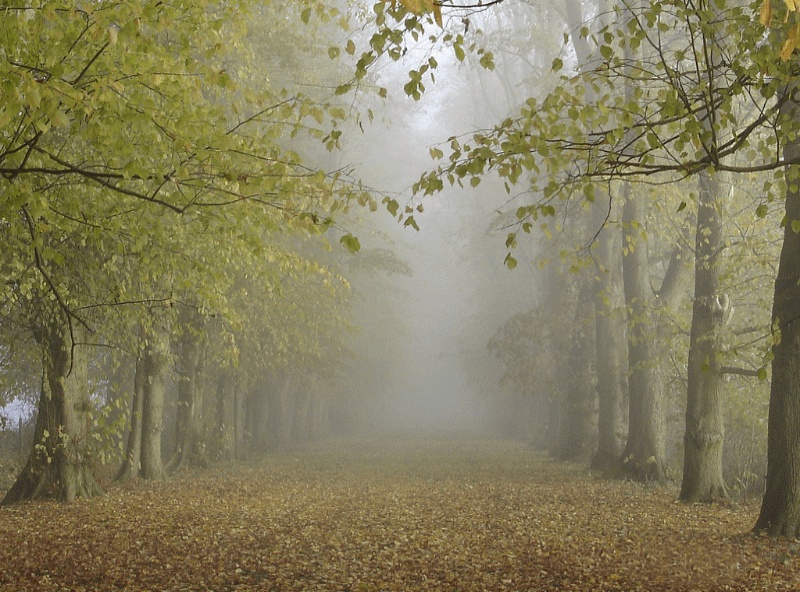}
\label{}}
\subfloat{\includegraphics[width=0.12\linewidth]{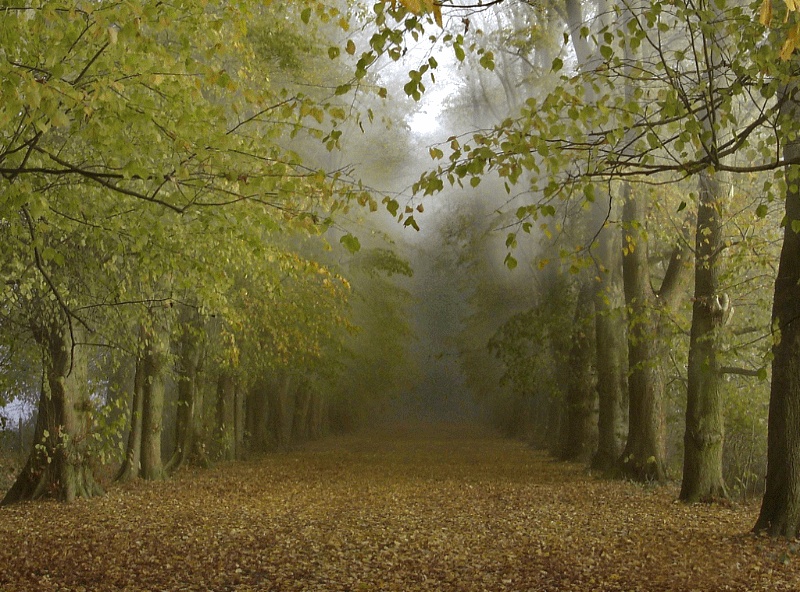}
\label{}}
\subfloat{\includegraphics[width=0.12\linewidth]{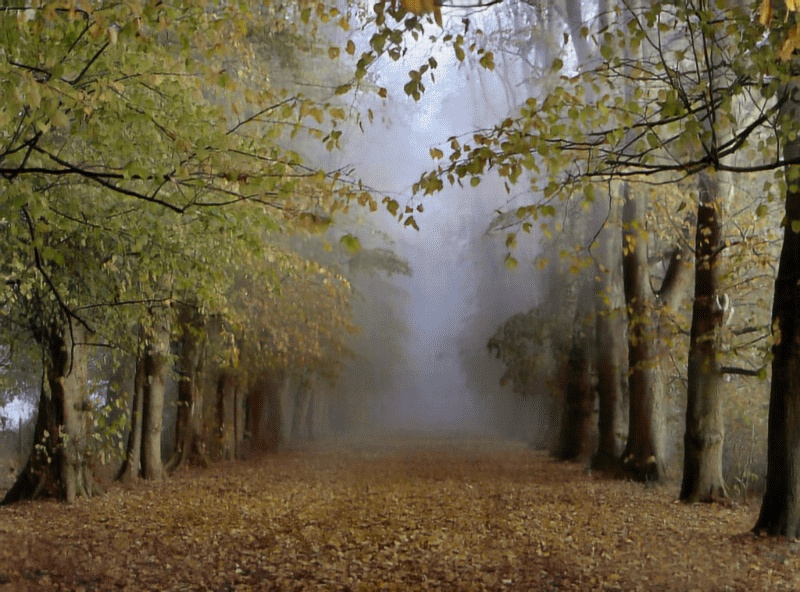}
\label{}}
\subfloat{\includegraphics[width=0.12\linewidth]{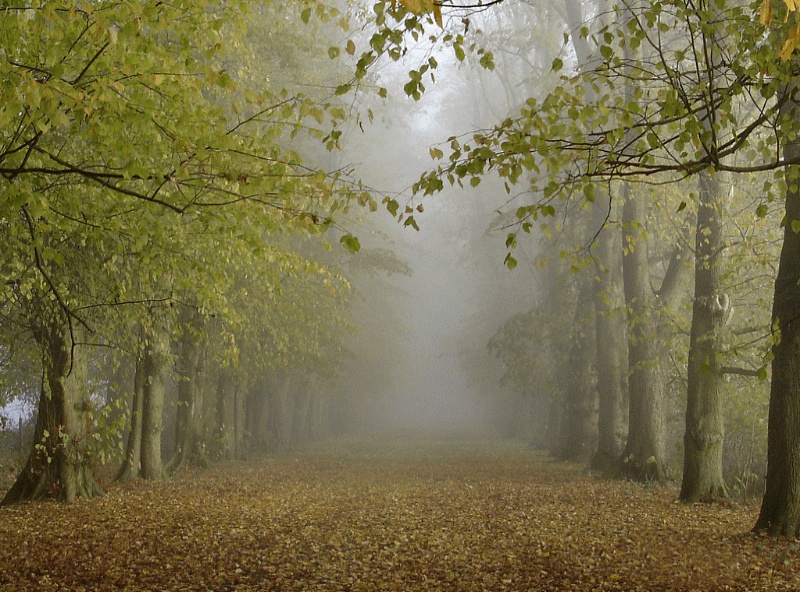}
\label{}}
\subfloat{\includegraphics[width=0.12\linewidth]{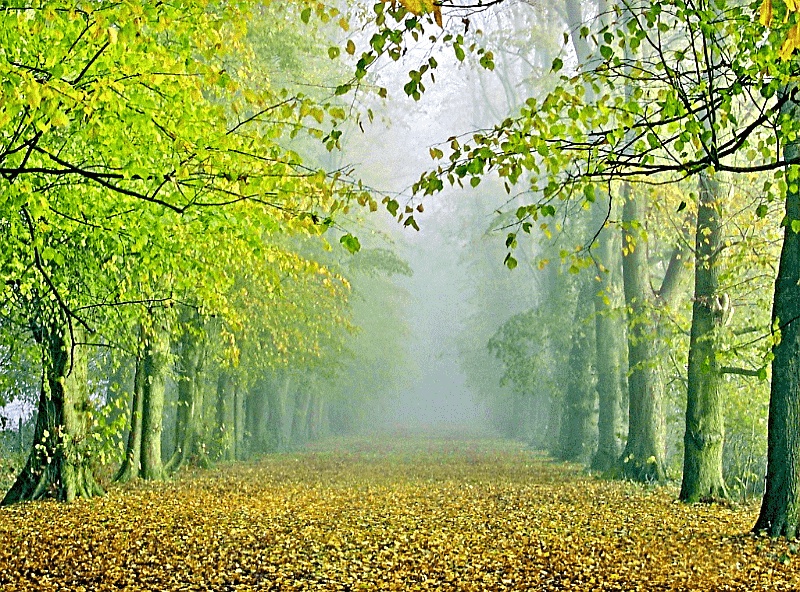}
\label{}}
\subfloat{\includegraphics[width=0.12\linewidth]{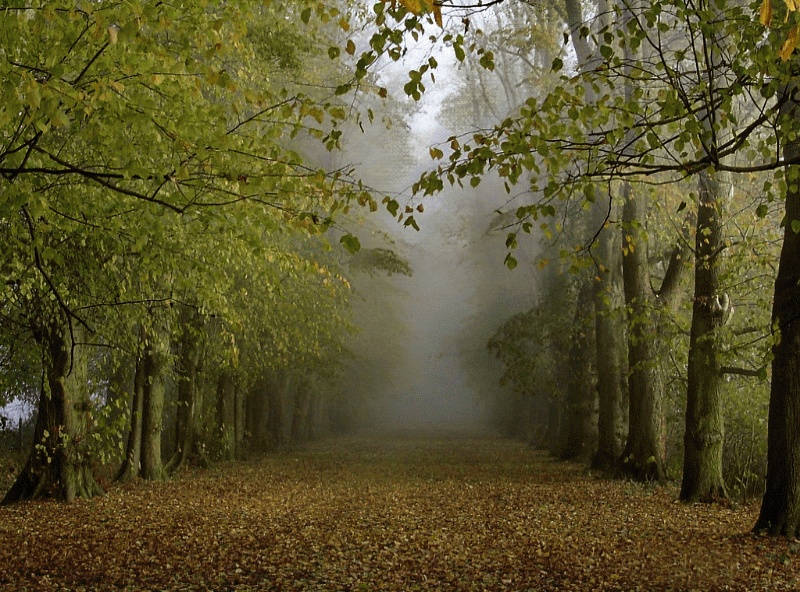}
\label{}}
\subfloat{\includegraphics[width=0.12\linewidth]{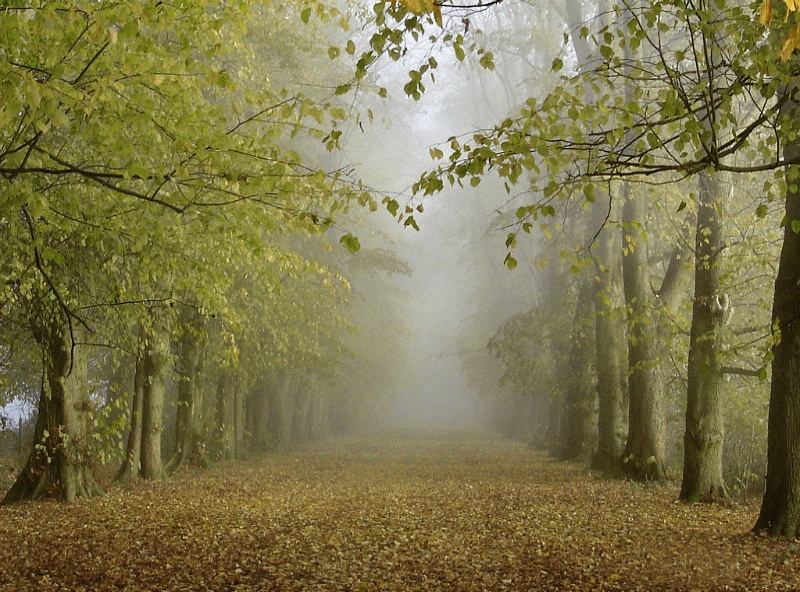}
\label{}}
\subfloat{\includegraphics[width=0.12\linewidth]{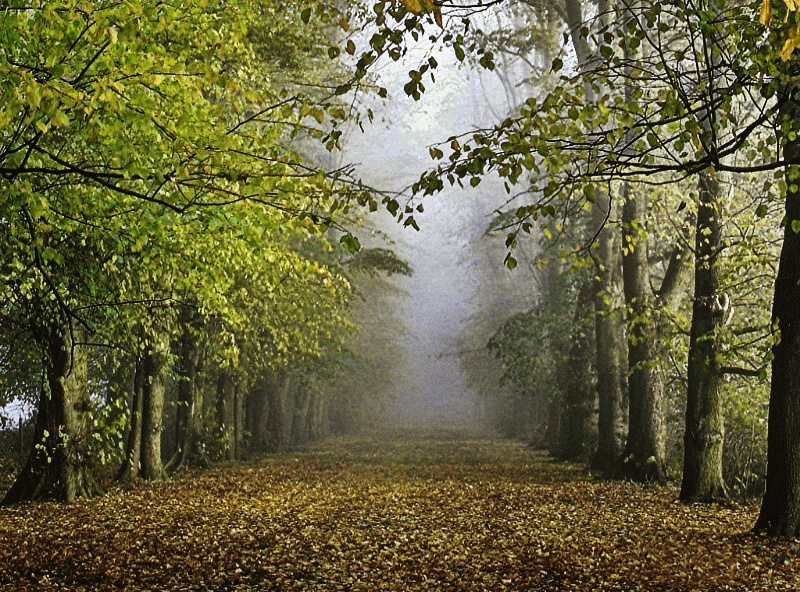}
\label{}}\\\vspace{-2mm}
\makebox[0.118\linewidth][c]{\scriptsize 0.44 / 0.42}
\makebox[0.118\linewidth][c]{\scriptsize 0.21 / 0.31}
\makebox[0.118\linewidth][c]{\scriptsize 0.32 / 0.36}
\makebox[0.118\linewidth][c]{\scriptsize 0.31 / 0.39}
\makebox[0.118\linewidth][c]{\scriptsize 0.17 / 0.17}
\makebox[0.118\linewidth][c]{\scriptsize 0.17 / 0.20}
\makebox[0.118\linewidth][c]{\scriptsize 0.25 / 0.36}
\makebox[0.118\linewidth][c]{\scriptsize 0.13 / 0.10}\\\vspace{-3mm}
\subfloat{\includegraphics[width=0.12\linewidth]{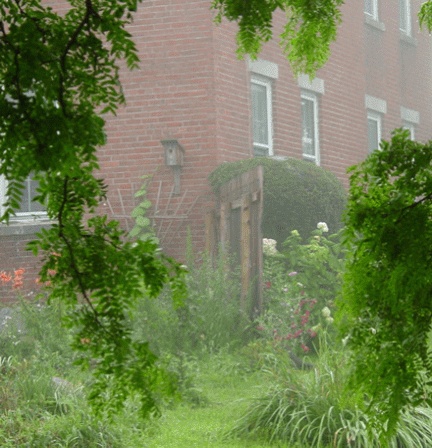}
\label{}}
\subfloat{\includegraphics[width=0.12\linewidth]{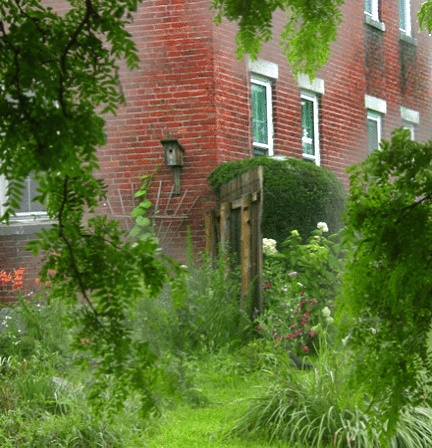}
\label{}}
\subfloat{\includegraphics[width=0.12\linewidth]{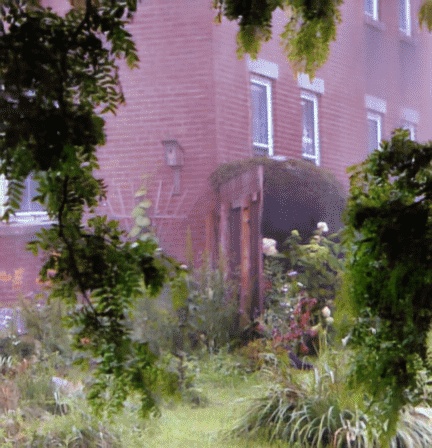}
\label{}}
\subfloat{\includegraphics[width=0.12\linewidth]{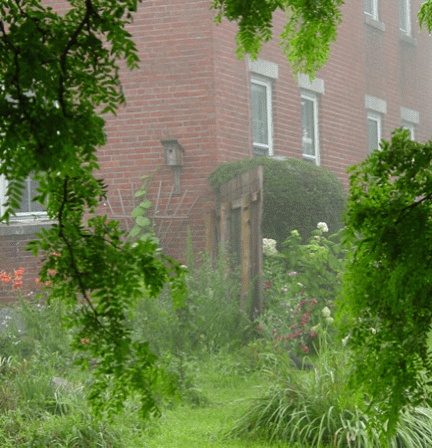}
\label{}}
\subfloat{\includegraphics[width=0.12\linewidth]{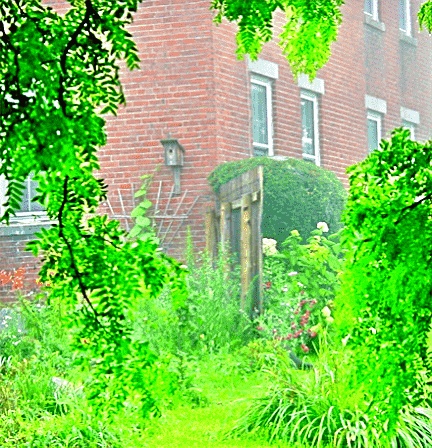}
\label{}}
\subfloat{\includegraphics[width=0.12\linewidth]{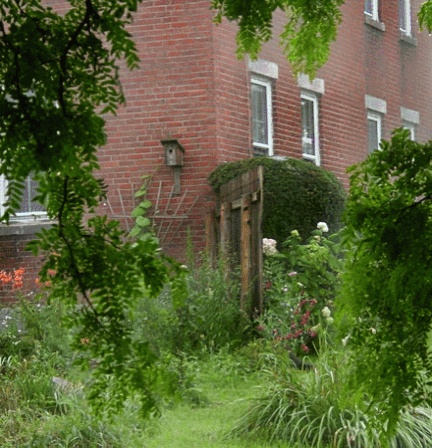}
\label{}}
\subfloat{\includegraphics[width=0.12\linewidth]{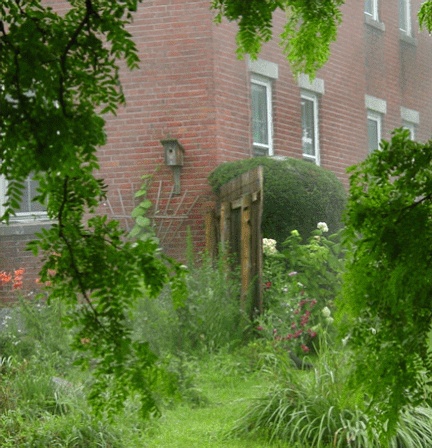}
\label{}}
\subfloat{\includegraphics[width=0.12\linewidth]{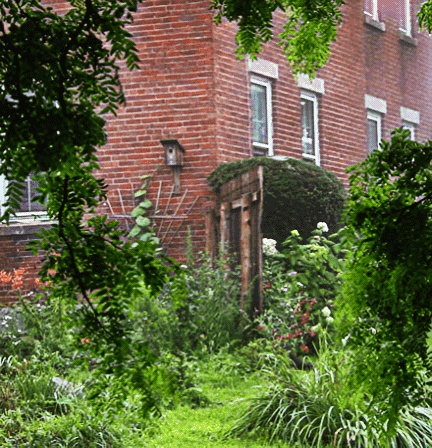}
\label{}}\\\vspace{-2mm}
\makebox[0.118\linewidth][c]{\scriptsize 0.31 / 0.32}
\makebox[0.118\linewidth][c]{\scriptsize 0.16 / 0.19}
\makebox[0.118\linewidth][c]{\scriptsize 0.34 / 0.42}
\makebox[0.118\linewidth][c]{\scriptsize 0.26 / 0.29}
\makebox[0.118\linewidth][c]{\scriptsize 0.17 / 0.07}
\makebox[0.118\linewidth][c]{\scriptsize 0.17 / 0.17}
\makebox[0.118\linewidth][c]{\scriptsize 0.21 / 0.27}
\makebox[0.118\linewidth][c]{\scriptsize 0.11 / 0.07}\\
\makebox[0.118\linewidth][c]{\scriptsize (a)}
\makebox[0.118\linewidth][c]{\scriptsize (b)}
\makebox[0.118\linewidth][c]{\scriptsize (c)}
\makebox[0.118\linewidth][c]{\scriptsize (d)}
\makebox[0.118\linewidth][c]{\scriptsize (e)}
\makebox[0.118\linewidth][c]{\scriptsize (f)}
\makebox[0.118\linewidth][c]{\scriptsize (g)}
\makebox[0.118\linewidth][c]{\scriptsize (h)}\vspace{-2mm}
\caption{Qualitative evaluation of defogging performance on real-world foggy images. The corresponding performance scores under the images are FADE / HazDesNet. (a) Foggy (b) DCP (c) DAD (d) MSBDN (e) PSD (f) RefineDNet (g) SLAD (h) Ours.}
\label{}
\end{figure*}

\begin{figure*}[!h]
\centering
\subfloat{\includegraphics[width=0.2\linewidth]{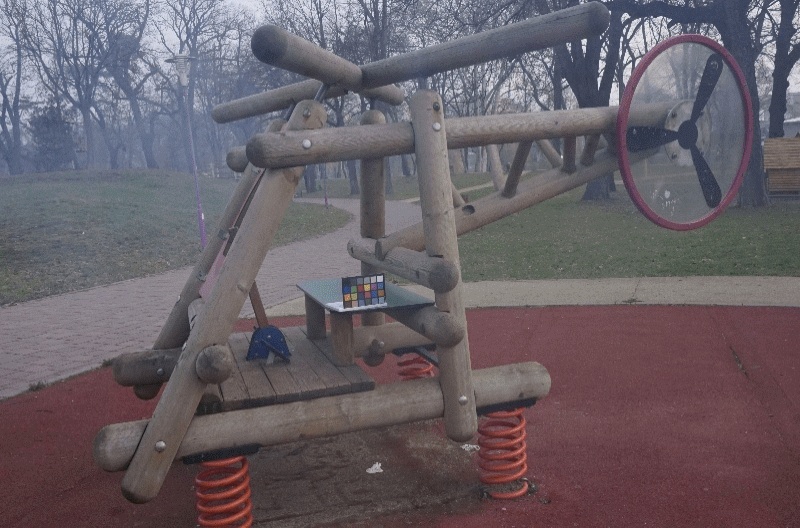}%
\label{}}\hspace{1mm}
\subfloat{\includegraphics[width=0.2\linewidth]{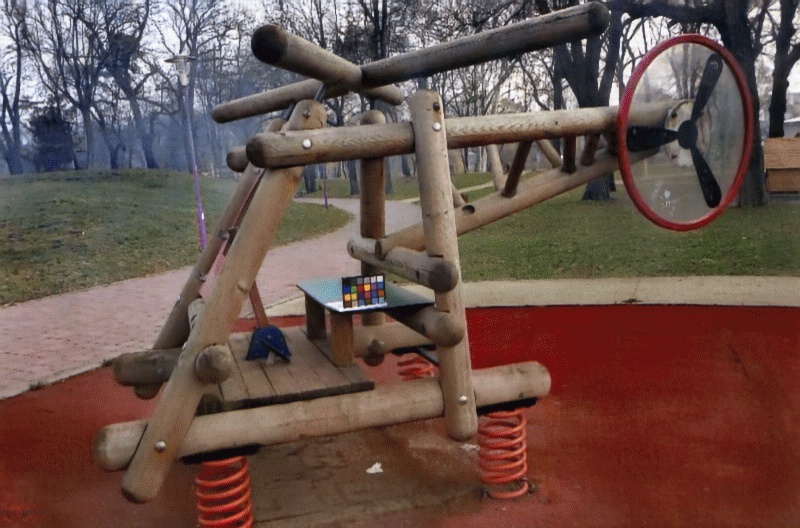}%
\label{}}\hspace{1mm}
\subfloat{\includegraphics[width=0.2\linewidth]{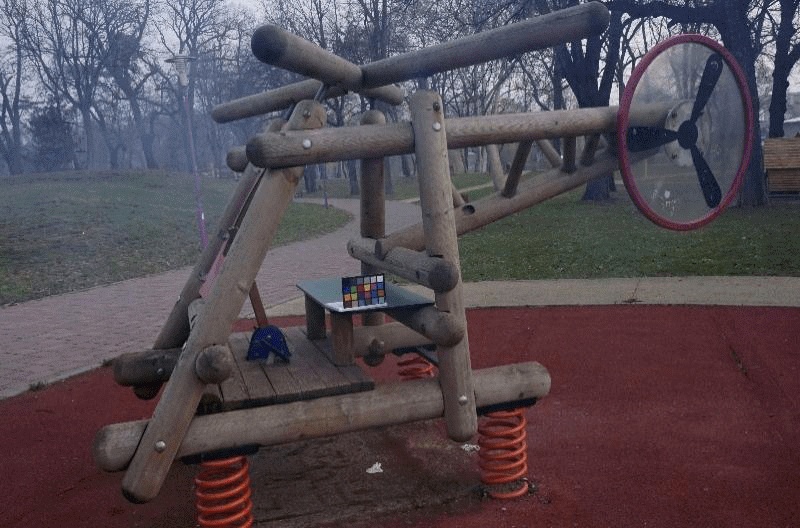}%
\label{}}\hspace{1mm}
\subfloat{\includegraphics[width=0.2\linewidth]{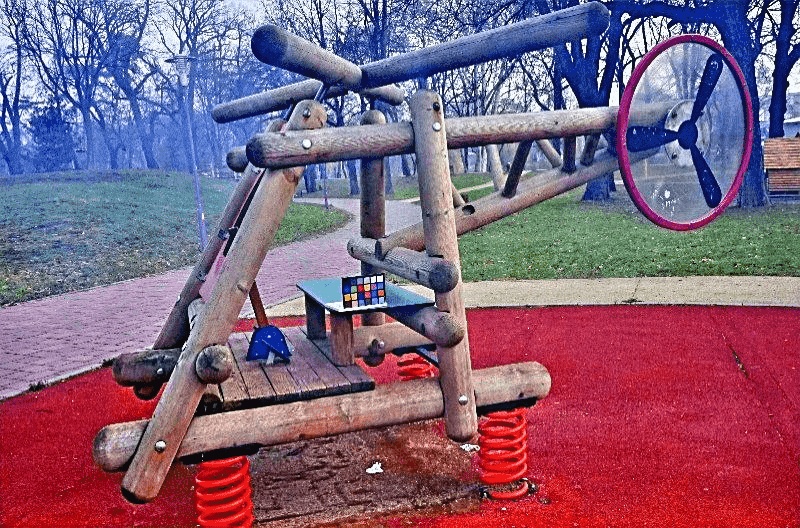}%
\label{}}
\\
\makebox[0.2\linewidth][c]{\footnotesize (a) Foggy}\hspace{1mm}
\makebox[0.2\linewidth][c]{\footnotesize (b) DAD}\hspace{1mm}
\makebox[0.2\linewidth][c]{\footnotesize (c) MSBDN}\hspace{1mm}
\makebox[0.2\linewidth][c]{\footnotesize (d) PSD}
\\\vspace{-2mm}
\subfloat{\includegraphics[width=0.2\linewidth]{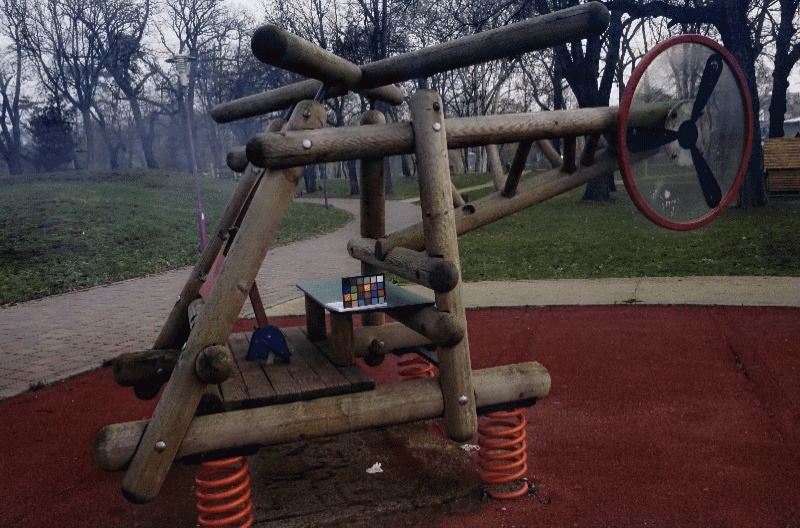}%
\label{}}\hspace{1mm}
\subfloat{\includegraphics[width=0.2\linewidth]{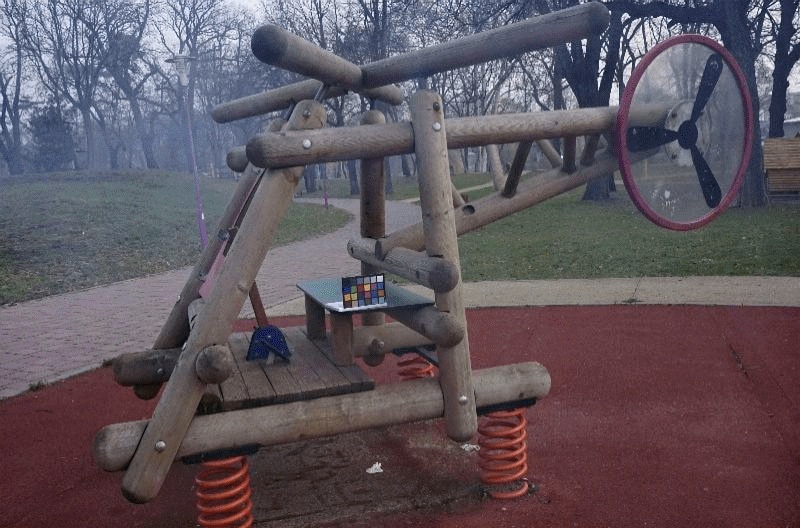}%
\label{}}\hspace{1mm}
\subfloat{\includegraphics[width=0.2\linewidth]{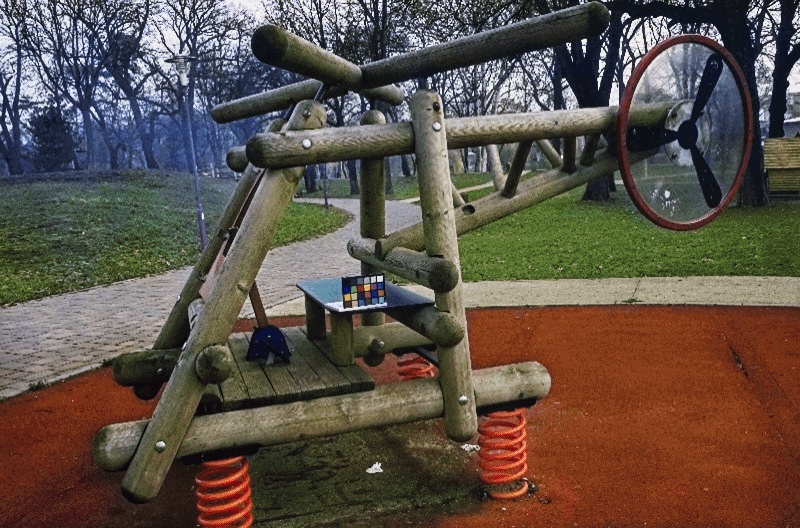}%
\label{}}\hspace{1mm}
\subfloat{\includegraphics[width=0.2\linewidth]{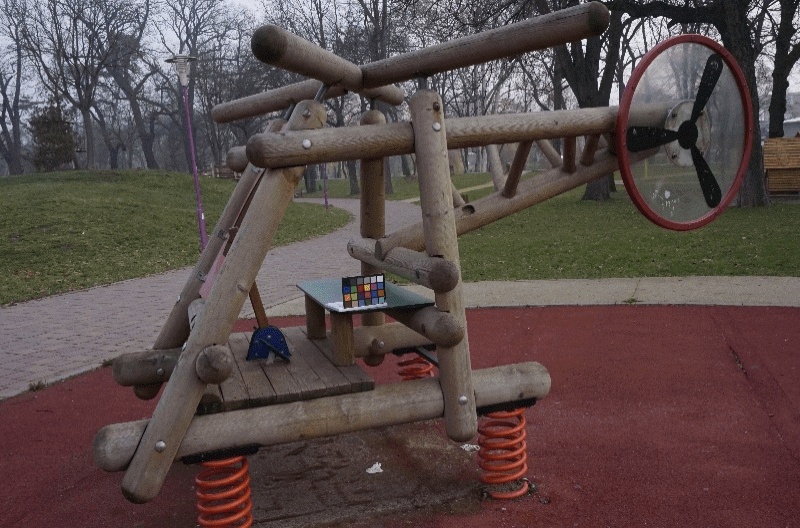}%
\label{}}
\\
\makebox[0.2\linewidth][c]{\footnotesize (e) RefineDNet}\hspace{1mm}
\makebox[0.2\linewidth][c]{\footnotesize (f) SLAD}\hspace{1mm}
\makebox[0.2\linewidth][c]{\footnotesize (g) Ours}\hspace{1mm}
\makebox[0.2\linewidth][c]{\footnotesize (h) GT}\vspace{1mm}
\subfloat{\includegraphics[width=0.2\linewidth]{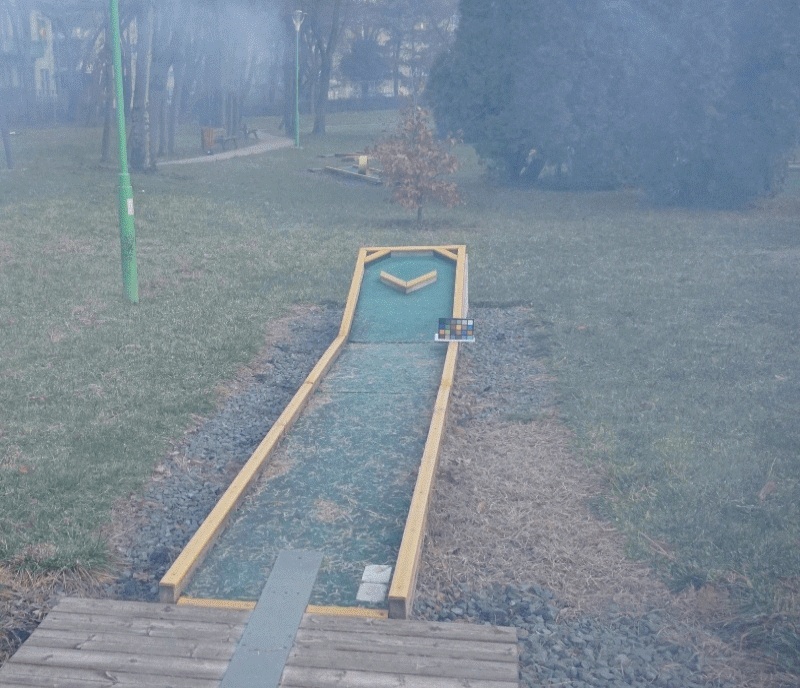}%
\label{}}\hspace{1mm}
\subfloat{\includegraphics[width=0.2\linewidth]{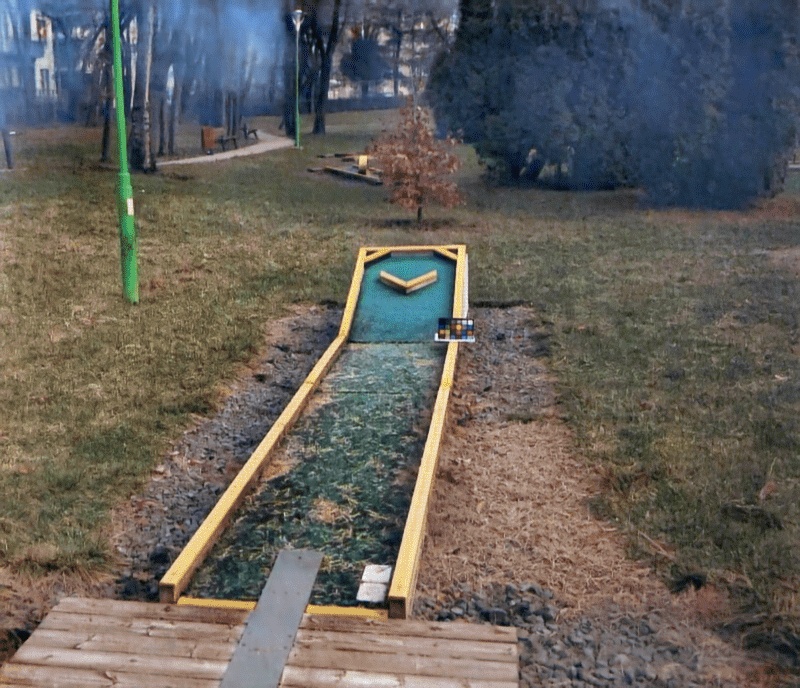}%
\label{}}\hspace{1mm}
\subfloat{\includegraphics[width=0.2\linewidth]{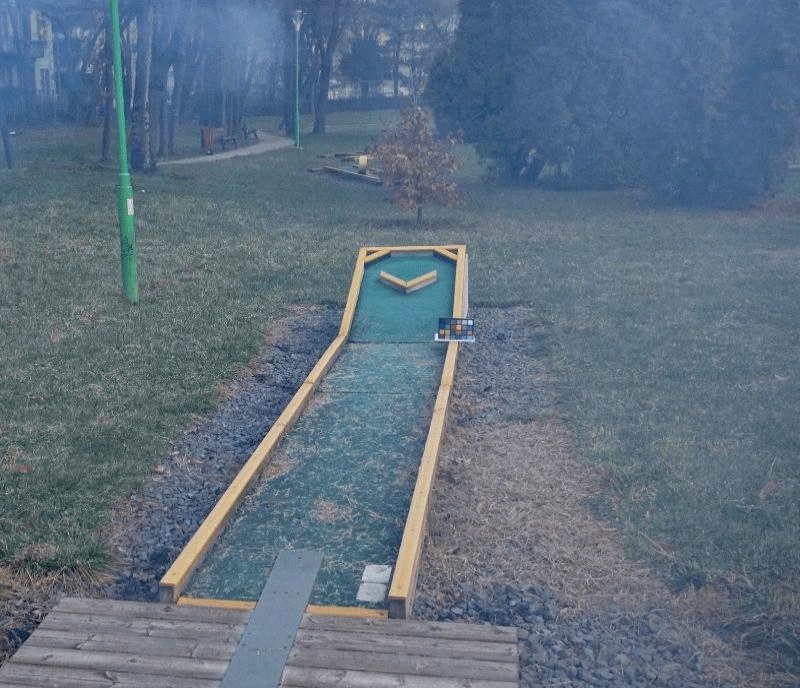}%
\label{}}\hspace{1mm}
\subfloat{\includegraphics[width=0.2\linewidth]{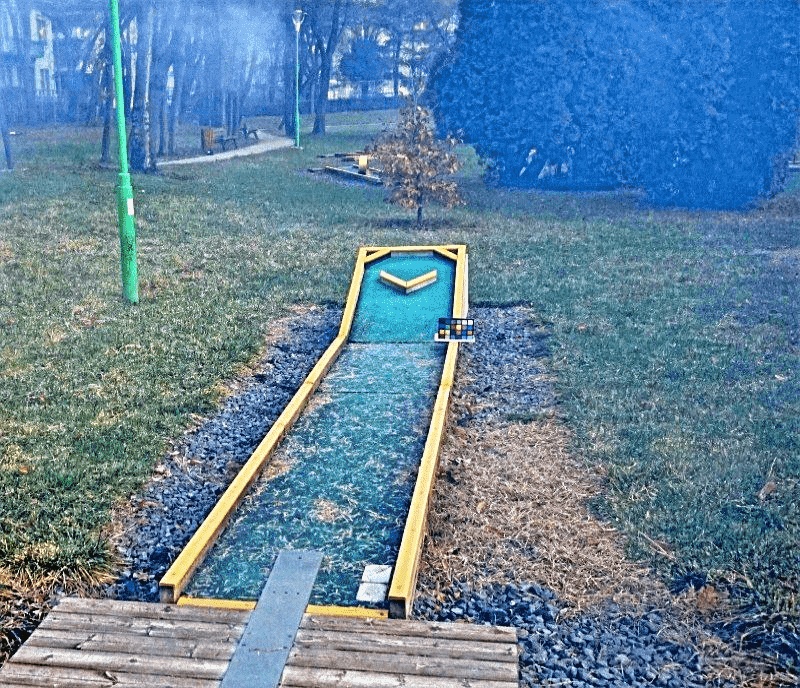}%
\label{}}
\\
\makebox[0.2\linewidth][c]{\footnotesize (a) Foggy}\hspace{1mm}
\makebox[0.2\linewidth][c]{\footnotesize (b) DAD}\hspace{1mm}
\makebox[0.2\linewidth][c]{\footnotesize (c) MSBDN}\hspace{1mm}
\makebox[0.2\linewidth][c]{\footnotesize (d) PSD}
\\\vspace{-2mm}
\subfloat{\includegraphics[width=0.2\linewidth]{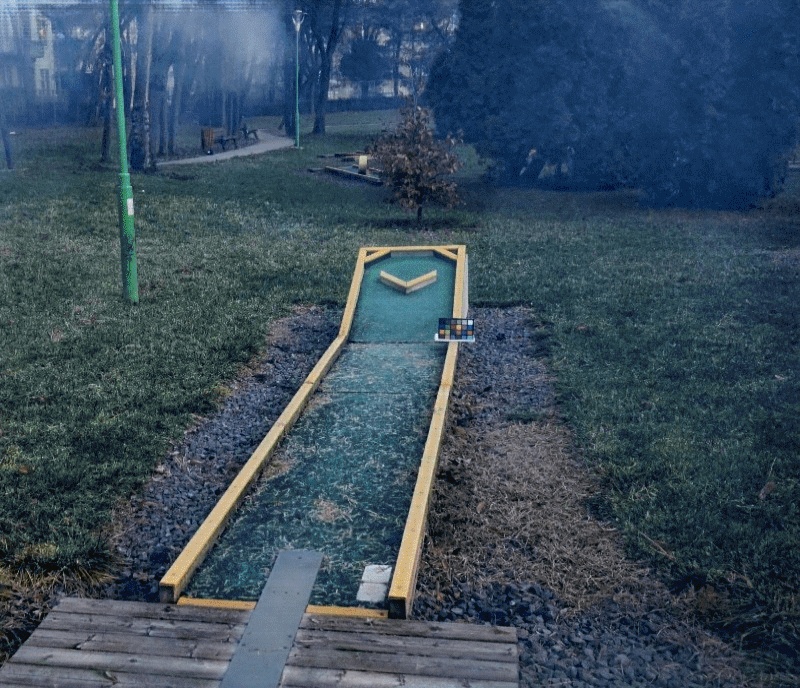}%
\label{}}\hspace{1mm}
\subfloat{\includegraphics[width=0.2\linewidth]{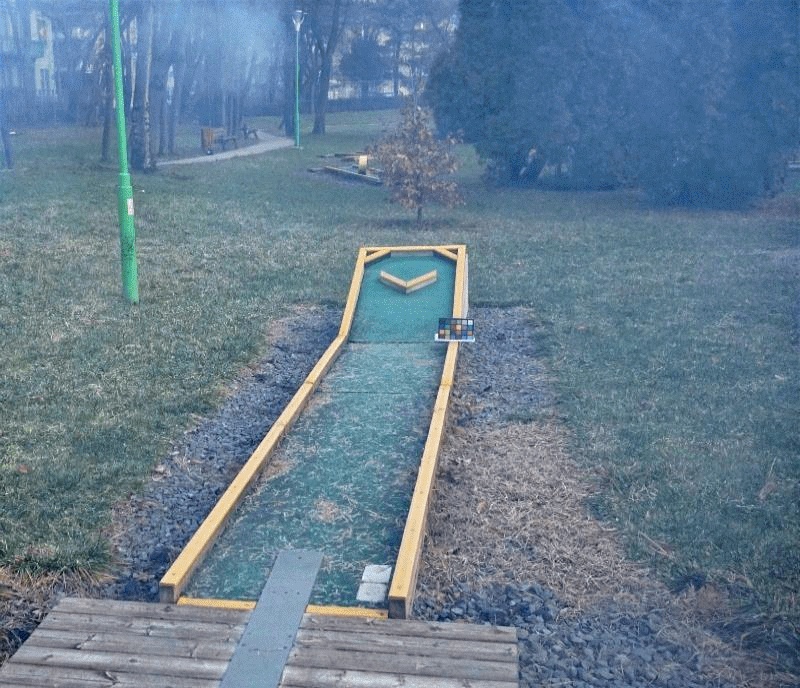}%
\label{}}\hspace{1mm}
\subfloat{\includegraphics[width=0.2\linewidth]{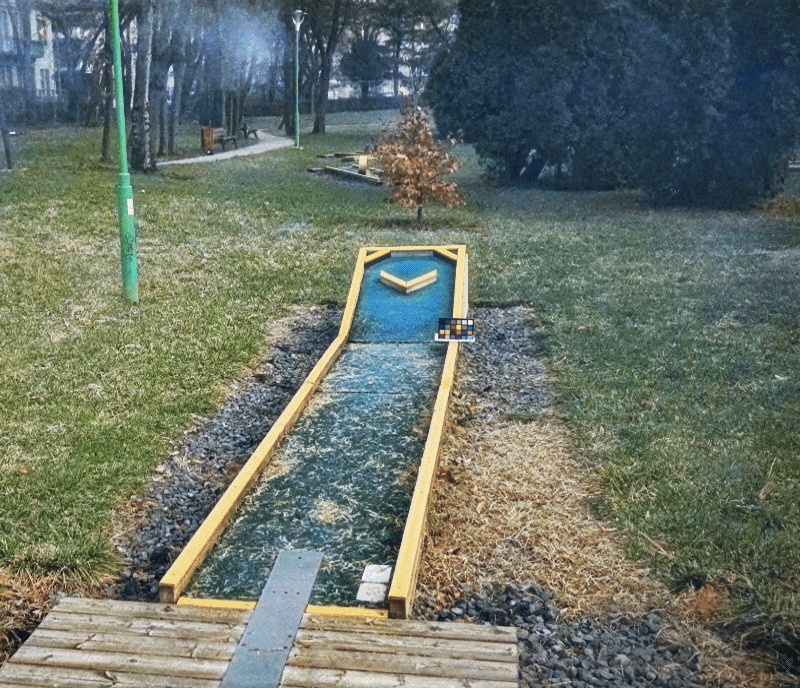}%
\label{}}\hspace{1mm}
\subfloat{\includegraphics[width=0.2\linewidth]{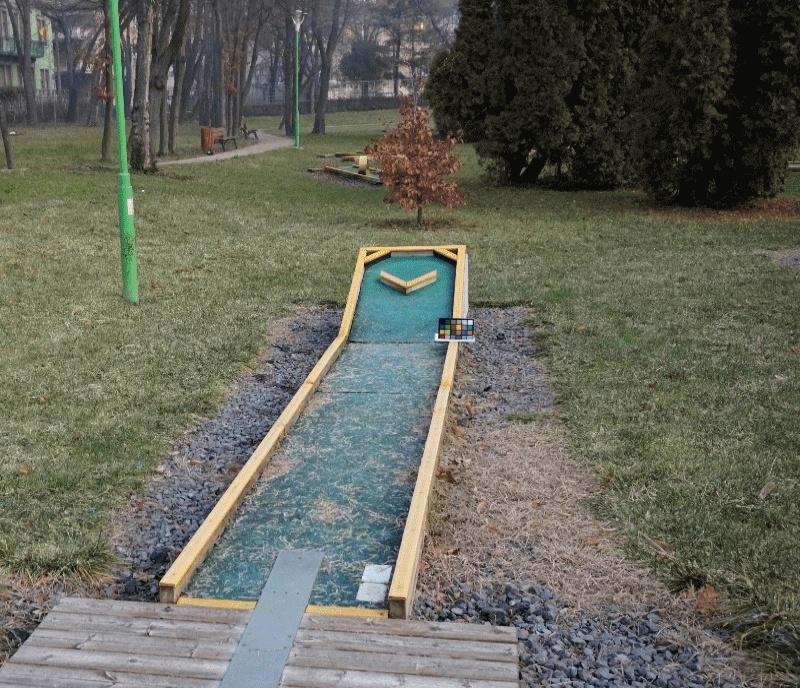}%
\label{}}
\\
\makebox[0.2\linewidth][c]{\footnotesize (e) RefineDNet}\hspace{1mm}
\makebox[0.2\linewidth][c]{\footnotesize (f) SLAD}\hspace{1mm}
\makebox[0.2\linewidth][c]{\footnotesize (g) Ours}\hspace{1mm}
\makebox[0.2\linewidth][c]{\footnotesize (h) GT}\vspace{-2mm}
\caption{Qualitative evaluation of defogging performance on the O-HAZE dataset.}
\label{}
\end{figure*}

\subsection{Comparison With State-of-the-Art Methods}
We compare the performance of the PG-CycleGAN with several state-of-the-art defogging algorithms as mentioned earlier, and a set of experiments including qualitative and quantitative on overwater and land images is conducted.

\textit{1) Qualitative comparisons on overwater images:} In this experiment, we evaluate the visual quality of the defogged images obtained by PG-CycleGAN on real-world overwater images.

Fig. 7 shows the results of different methods on overwater images. We can observe that, in general, the resulting images given by our method has higher brightness than those by DCP \cite{he2010single}. This is because of the inaccurate estimation of atmospheric light with the DCP method. The results of the DAD \cite{shao2020domain} often suffer from color distortion and unrealistic looking. Due to the domain gap between the synthetic and real-world foggy images, images restored by MSBDN \cite{Dong_2020_CVPR} remain somewhat foggy, especially in distant areas. PSD \cite{chen2021psd} sometimes fails to handle the overwater scene, which leads to severe color shifting. Although RefineDNet \cite{zhao2021refinednet} shows pretty good results, it tends to darken the images, and the sky regions in the 5th row of Fig. 7 are blurred. This may be due to the drawback that the RefineDNet \cite{zhao2021refinednet} adopts the DCP \cite{he2010single} in the first stage to restore visibility, while DCP is difficult to restore sky regions of the foggy images. As can be seen in the 4th and 6th rows of Fig. 7, the defogged images of the self-supervised method SLAD \cite{Liang_2022_IJCAI} also remain somewhat foggy.

\textit{2) Qualitative comparisons on land images:} In this experiment, we evaluate the visual quality of the defogged images by PG-CycleGAN on land images.

Figs. 8 and 9 show the results of different methods on real-world land images on the O-HAZE dataset. From the 4th row of Fig. 8, we can observe that with the DCP method, fog around the leaves still remains. The MSBDN, PSD, and SLAD degrade substantially when dealing with thick fog, as shown in Fig. 9, where certain amount of fog still remains. Although DAD can remove the fog well, the resulting images have severe color distortion. Due to the use of the DCP in the first defogging stage, the resulting images by RefineDNet are too dark as compared to the ground-truth.

In comparison, our network can generate cleaner images from the overwater and land foggy images. As demonstrated in the zoom-in regions in Fig. 7, e.g. the 1st and 2nd rows, the crowd and cars are well recovered due to the prior map guided losses. Similarly, as shown in the 3rd row of Fig. 7, other methods neglect the fine rope, yet our model can still effectively defog it. Furthermore, the 4th-6th rows in Fig. 7 show that our results have less remaining fog and artifacts. From Fig. 9, we can observe that the results of PG-CycleGAN are closer to the ground-truth than other methods. Considering those results together, we can see that PG-CycleGAN restores more details and obtains images that are visually clearer.

\textit{3) Quantitative comparisons on overwater images:} Because the overwater images in the test set have no corresponding clean images, full-reference image quality assessment metrics are not suitable for our experiments. Recently, some non-reference image quality assessment metrics \cite{choi2015referenceless}\cite{guan2022visibility}\cite{hautiere2008blind}\cite{zhang2020hazdesnet}  have been proposed for quantitatively evaluating defogging performance. We adopt three non-reference evaluation criteria: FADE \cite{choi2015referenceless}, HazDesNet \cite{zhang2020hazdesnet}, and BAVE  \cite{hautiere2008blind}, in which FADE and HazDesNet are designed to evaluate the defogging performance by predicting the density of the fog. Lower values of the FADE and HazDesNet denote better defogging performance. The BAVE is an evaluator of visibility enhancement that contains three indicators, which are the rate of visible edges $e$, the quality value of the contrast restoration $r$, and the normalized saturate value of pixels $\sigma$. For a defogged image, the higher values of $e$ and $r$, 
and the lower value of $\sigma$ denote the better quality of the recovered image.

A quantitative comparison of the test set which contains 188 real-world overwater images is reported in Table \uppercase\expandafter{\romannumeral2}. It can be seen that our method gives the best results in both FADE and the HazDesNet, which demonstrate that the PG-CycleGAN can remove the fog more effectively. Although PG-CycleGAN is ranked the second in terms of visible edges $e$ and the quality value of the contrast restoration $r$, and the third in terms of the indicator $\sigma$, it is not far from the best results. Combining all the performance indicators, it can be argued that our results have less fog remaining and better visual quality.

\begin{table}[!h]
\renewcommand\arraystretch{1.2}
\begin{center}
\caption{Quantitative comparison with state-of-the-art methods on 188 real-world overwater images.}
\label{}
\begin{tabular}{ c  c  c  c  c  c  c  c }
\toprule
 & FADE $\downarrow$ & HazDesNet $\downarrow$ & $e$ $\uparrow$ & $\sigma(\%)$ $\downarrow$ & $r$ $\uparrow$\\
 \hline
DCP         & 1.08 & $0.33^2$ & 2.85 & 0.006 & 1.68 \\
DAD         & 1.36 & 0.49 & 2.03 & $0.000^1$ & 2.18 \\
MSBDN       & 2.28 & 0.53 & 1.01 & 0.087 & 1.50 \\
PSD         & $1.01^1$ & 0.41 & 2.36 & 1.146 & $3.57^1$ \\
RefineDNet  & 1.06 & 0.36 & $3.94^1$ & $0.001^1$ & 2.16 \\
SLAD        & 1.07 & 0.40 & 3.30 & 0.037 & 2.31 \\     
Ours        & $0.75^1$ & $0.32^1$ & $3.81^2$ & 0.005 & $2.94^2$ \\
\bottomrule
\end{tabular}
\end{center}
\end{table}

We further present the fog density map predicted by HazDesNet in Fig. 10. It can be seen that benefiting from the prior map guided losses, the fog density map of our result is closer to dark blue compared to other methods, which indicates that our method has better defogging performance.

\begin{figure}[!h]
\centering
\subfloat{\includegraphics[width=0.24\linewidth]{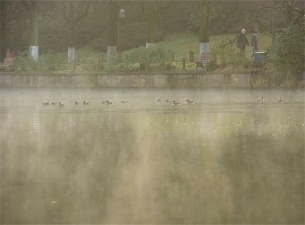}%
\label{}}
\hfill
\subfloat{\includegraphics[width=0.24\linewidth]{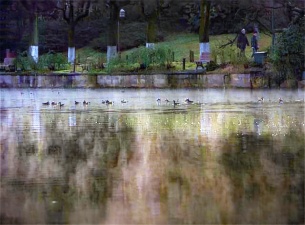}%
\label{}}
\hfill
\subfloat{\includegraphics[width=0.24\linewidth]{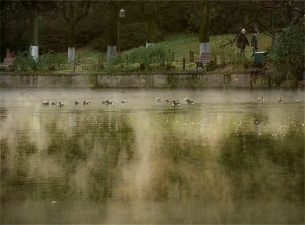}%
\label{}}
\hfill
\subfloat{\includegraphics[width=0.24\linewidth]{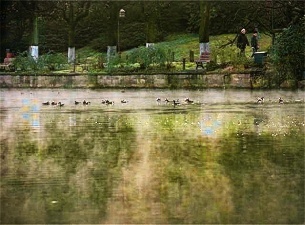}%
\label{}}
\hfill\\\vspace{-3mm}
\subfloat{\includegraphics[width=0.24\linewidth]{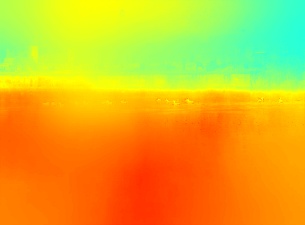}%
\label{}}
\hfill
\subfloat{\includegraphics[width=0.24\linewidth]{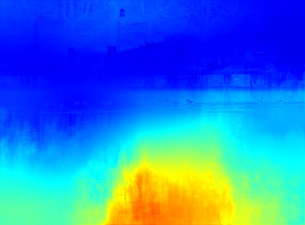}%
\label{}}
\hfill
\subfloat{\includegraphics[width=0.24\linewidth]{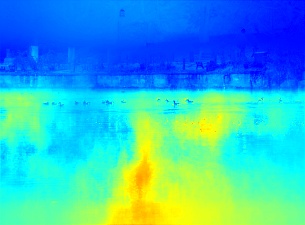}
\label{}}
\hfill
\subfloat{\includegraphics[width=0.24\linewidth]{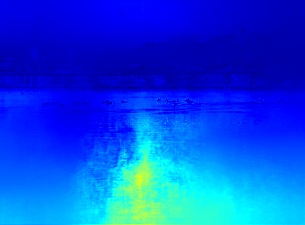}%
\label{}}
\hfill
\\
\hfill\makebox[0.24\linewidth][c]{\footnotesize (a) Foggy}\hfill\makebox[0.24\linewidth][c]{\footnotesize (b) DAD}\hfill\makebox[0.24\linewidth][c]{\footnotesize (c) RefineDNet }\hfill\makebox[0.24\linewidth][c]{\footnotesize (d) Ours}
\caption{Comparison of the HazDesNet predicted fog density maps on the defogging results. Dark blue indicates less fog and dark red indicates dense fog.}
\label{}
\end{figure}

\begin{table}[!h]
\renewcommand\arraystretch{1.2}
\begin{center}
\caption{Quantitative comparison with state-of-the-art methods on the O-HAZE and NH-HAZE datasets.}
\label{}
\begin{tabular}{ c  c  c  c  c  c  c  c }
\toprule
 & \multicolumn{2}{c}{O-HAZE} & \multicolumn{2}{c}{NH-HAZE}\\
 & SSIM & PSNR & SSIM & PSNR\\\hline
DAD         & $0.6948^1$ & $18.21^1$ & 0.5603 & $13.92^2$ \\
MSBDN       & 0.6109 & 16.64 & 0.4976 & 12.60 \\
PSD         & 0.5799 & 12.96 & $0.5604^2$ & 11.61 \\
RefineDNet  & 0.6749 & 17.06 & 0.5334 & 13.04 \\
SLAD & 0.6420 & 14.48 & 0.5422 & 11.79 \\
Ours        & $0.6755^2$ & $17.14^2$ & $0.5992^1$ & $13.19^2$ \\
\bottomrule
\end{tabular}
\end{center}
\end{table}

\textit{4) Quantitative comparisons on land images:}
In this experiment, we adopt the full-reference image quality assessment metrics SSIM and PSNR for comprehensively evaluating the performance of our proposed PG-CycleGAN. 

Table \uppercase\expandafter{\romannumeral3} gives the quantitative comparison on the O-HAZE and NH-HAZE datasets. We can observe that although the PG-CycleGAN is trained on overwater images, it achieves the best performance in SSIM on the NH-HAZE dataset, and ranked the second in terms of PSNR on both the O-HAZE and NH-HAZE datasets. Furthermore, as shown in Fig. 8, the values of FADE and HazDesNet of PG-CycleGAN are lower than other methods, which also demonstrates that our method leads to better quality images with less fog remaining than other methods.

Overall, the quantitative and qualitative comparisons show that our proposed method not only generates images with better visual effects on overwater images, but also offers excellent defogging performance for land images.

\section{Analysis and Discussion}
\subsection{Effectiveness of the PG-CycleGAN}
In order to verify the effectiveness of the LRC, UIM, and prior map guided losses, we conduct a series of ablation studies to analyze our method.

Fig. 11 illustrates the effectiveness of the LRC. After the coarse-to-fine framework is removed (as shown in Fig. 11 (b)), we can see that although learning the residual from the foggy input can retain  detailed information, the results tend to be insufficient defogging. In addition, due to the limited representation power of the network, the artifacts are also introduced. When the long-range residual connection is removed (as shown in Fig. 11 (c)), the defogging results show color shift. After the long-range residual connection and coarse-to-fine framework are both removed, the results may include color shift and heavy fog, and the results of CycleGAN also show the clear color shifts. The above discussion shows that the combination of the long-range residual connection and coarse-to-fine framework can result in a good complementarity that facilitates the restoration of detailed information and provides a desirable defogging effect.

\begin{figure}[h]
\centering
\subfloat{\includegraphics[width=0.16\linewidth]{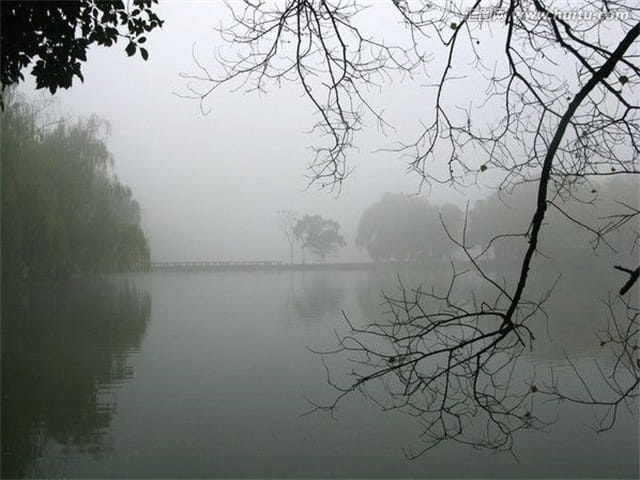}%
\label{}}
\hfil
\subfloat{\includegraphics[width=0.16\linewidth]{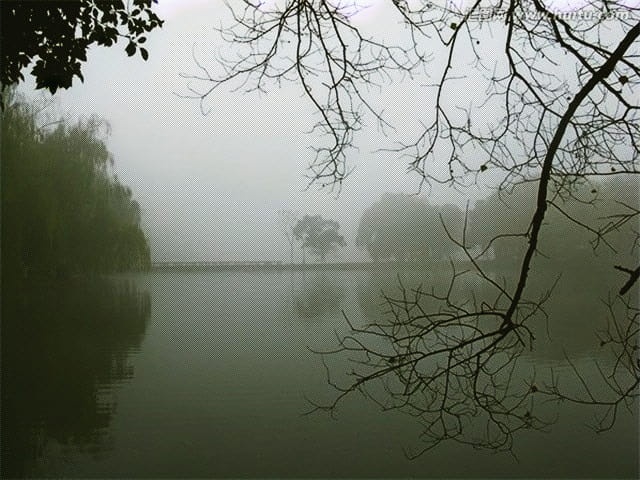}%
\label{}}
\hfil
\subfloat{\includegraphics[width=0.16\linewidth]{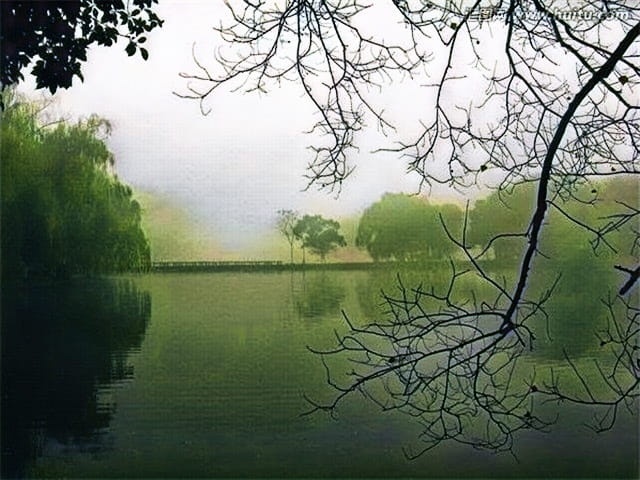}%
\label{}}
\hfil
\subfloat{\includegraphics[width=0.16\linewidth]{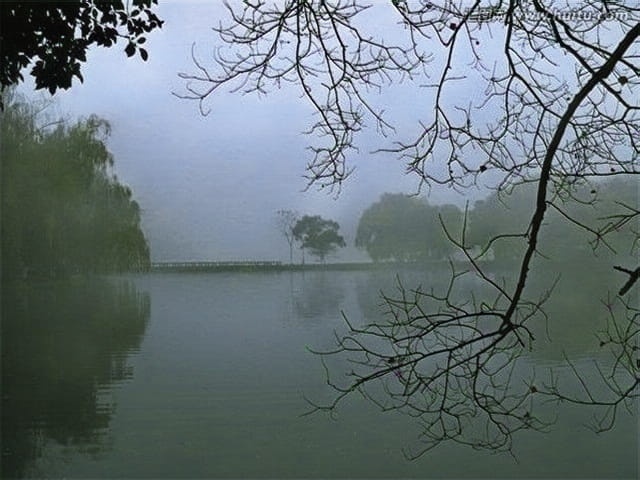}%
\label{}}
\hfil
\subfloat{\includegraphics[width=0.16\linewidth]{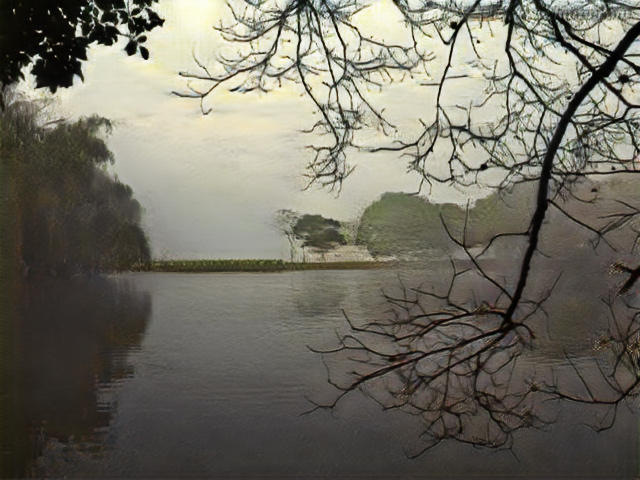}%
\label{}}
\hfil
\subfloat{\includegraphics[width=0.16\linewidth]{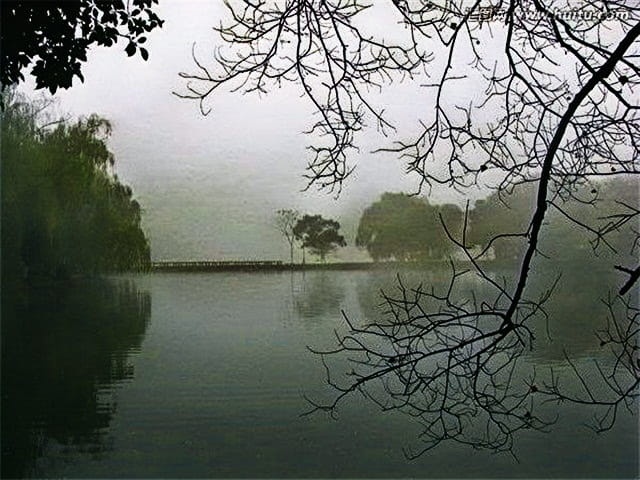}%
\label{}}
\hfil
\\\vspace{-3mm}
\subfloat{\includegraphics[width=0.16\linewidth]{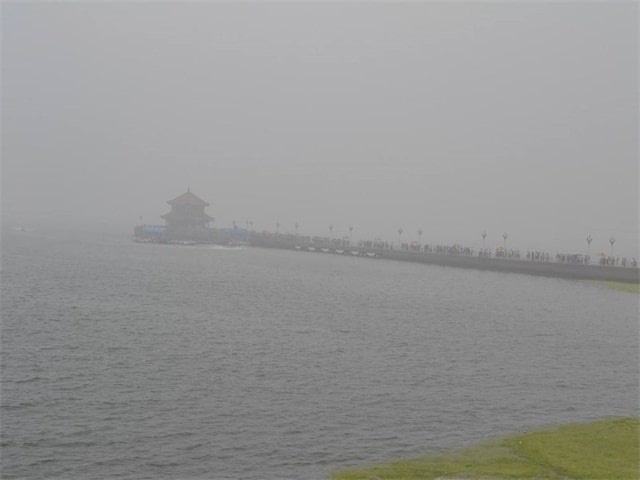}%
\label{}}
\hfil
\subfloat{\includegraphics[width=0.16\linewidth]{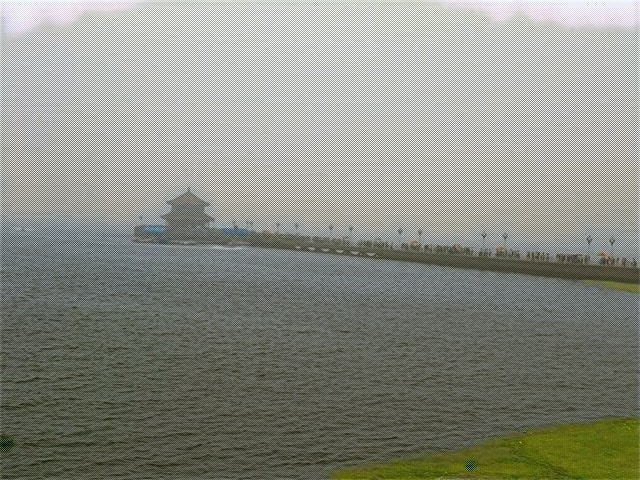}%
\label{}}
\hfil
\subfloat{\includegraphics[width=0.16\linewidth]{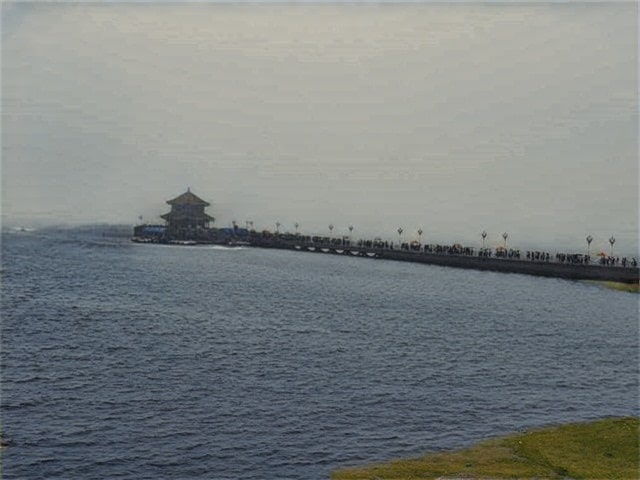}%
\label{}}
\hfil
\subfloat{\includegraphics[width=0.16\linewidth]{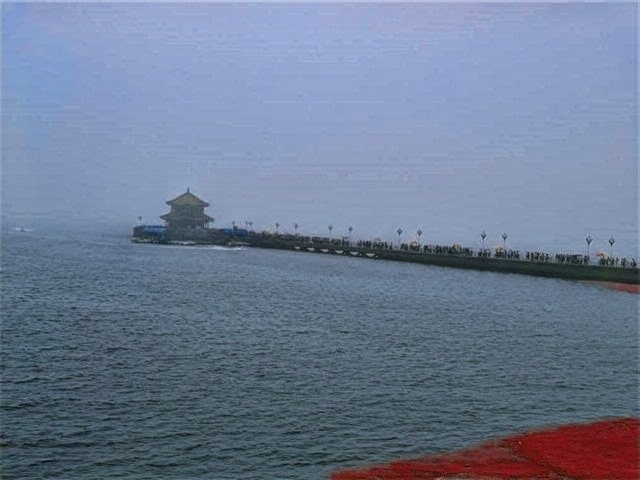}%
\label{}}
\hfil
\subfloat{\includegraphics[width=0.16\linewidth]{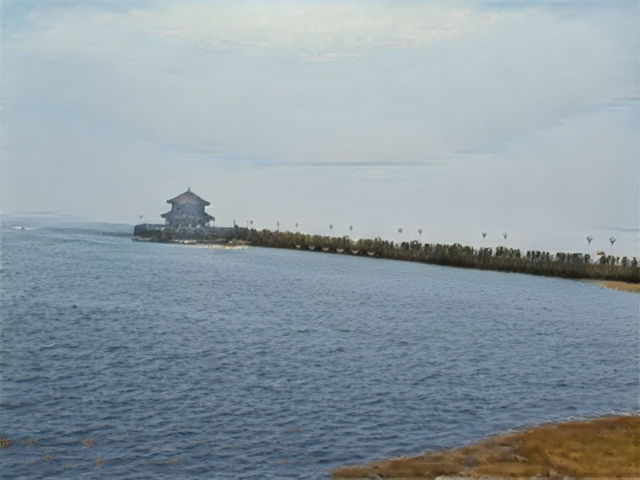}%
\label{}}\hfil
\subfloat{\includegraphics[width=0.16\linewidth]{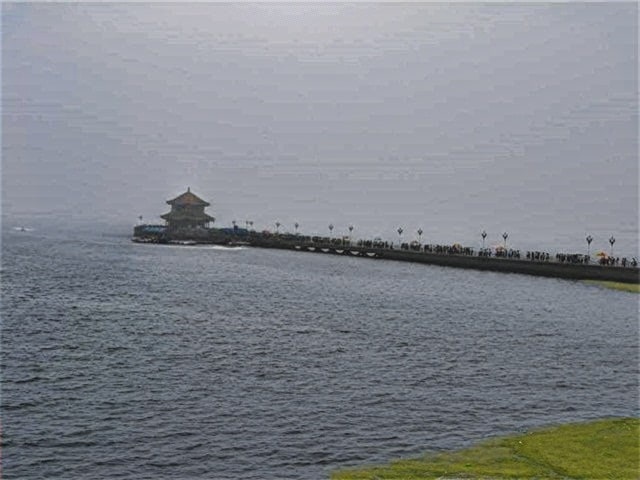}%
\label{}}\hfil
\\
\hfil\makebox[0.16\linewidth][c]{(a)}\hfil\makebox[0.16\linewidth][c]{(b)}\hfil\makebox[0.16\linewidth][c]{(c)}\hfil\makebox[0.16\linewidth][c]{(d)}\hfil\makebox[0.16\linewidth][c]{(e)}\hfil\makebox[0.16\linewidth][c]{(f)}
\caption{Effectiveness of the coarse-to-fine framework and the long-range residual connection. (a) Foggy images. (b) The results of PG-CycleGAN without using the coarse-to-fine framework. (c) The results of PG-CycleGAN without using the long-range residual connection. (d) The results of PG-CycleGAN without using both long-range residual connection and coarse-to-fine framework. (e)  The results of CycleGAN. (f)  The results of PG-CycleGAN.}
\label{}
\end{figure}

\begin{figure}[h]
\centering
\subfloat{\includegraphics[width=0.16\linewidth]{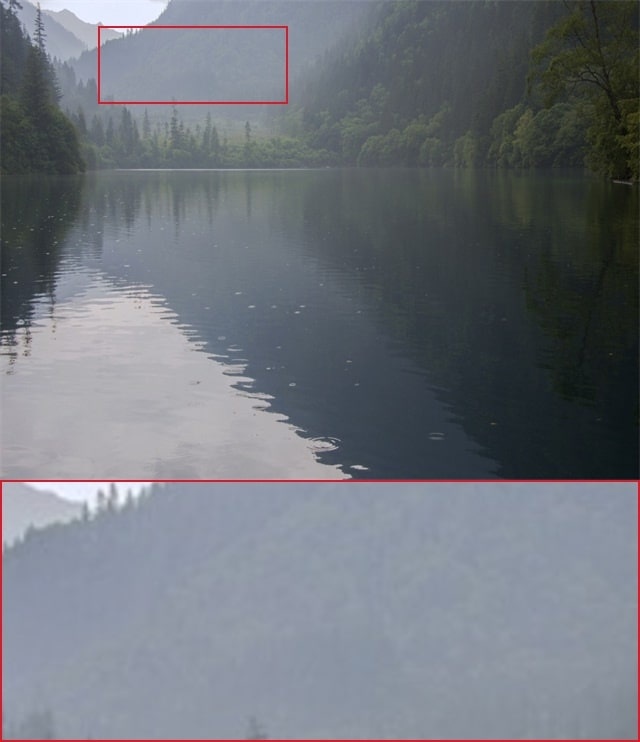}%
\label{}}
\hfil
\subfloat{\includegraphics[width=0.16\linewidth]{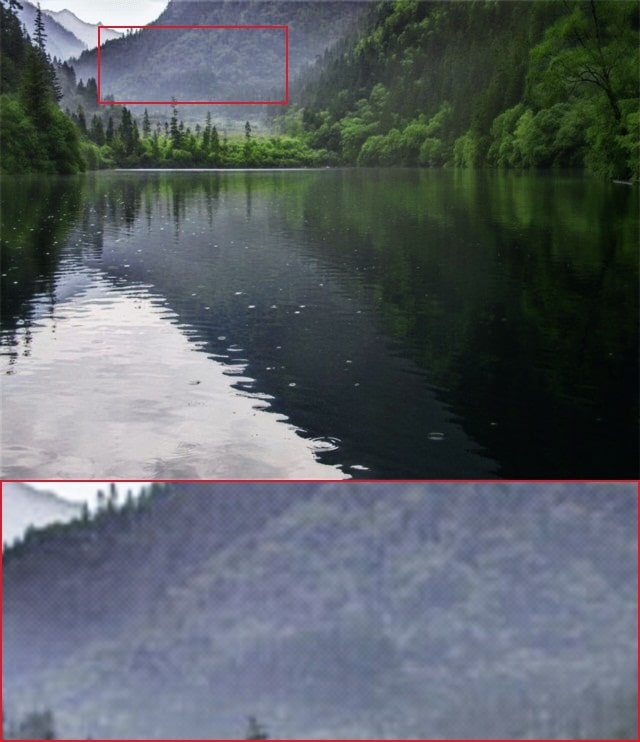}%
\label{}}
\hfil
\subfloat{\includegraphics[width=0.16\linewidth]{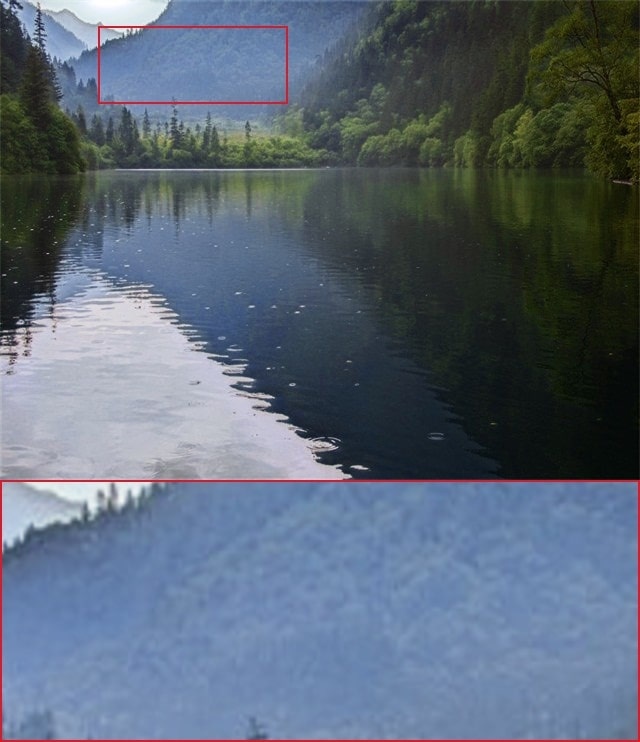}%
\label{}}
\hfil
\subfloat{\includegraphics[width=0.16\linewidth]{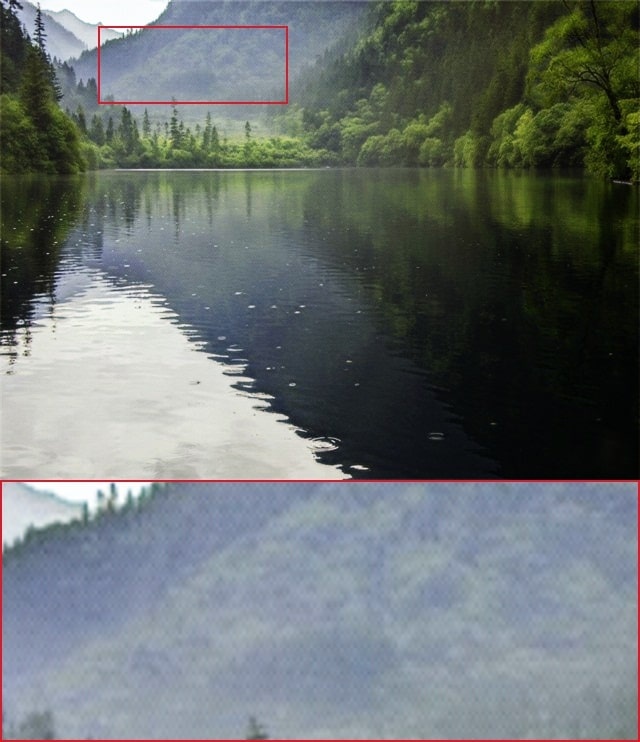}%
\label{}}
\hfil
\subfloat{\includegraphics[width=0.16\linewidth]{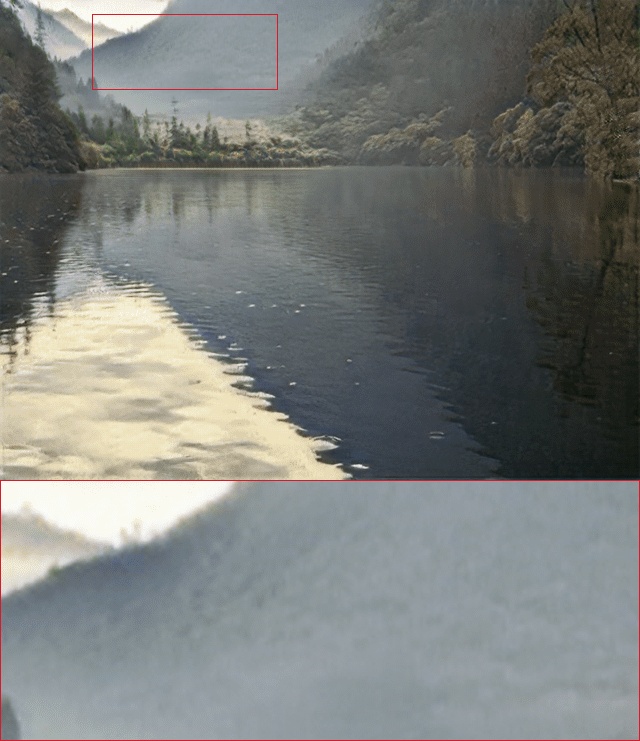}%
\label{}}
\hfil
\subfloat{\includegraphics[width=0.16\linewidth]{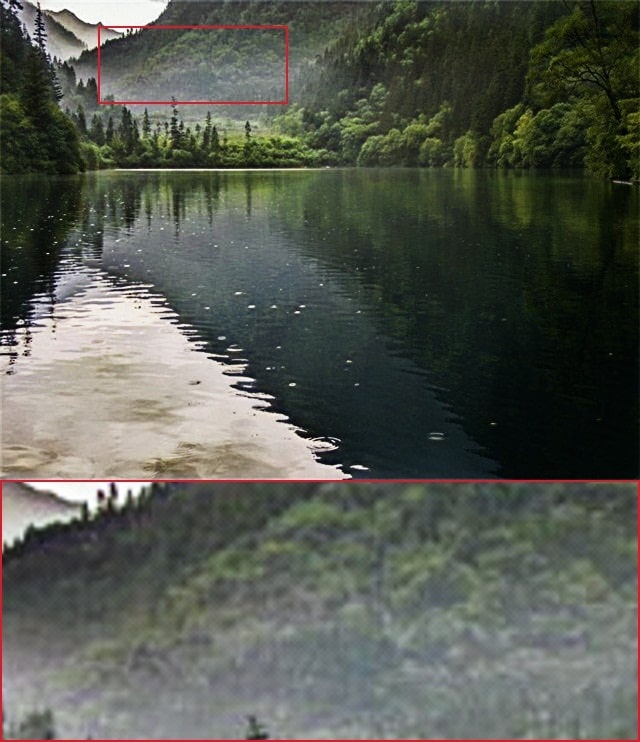}%
\label{}}
\hfil
\\\vspace{-3mm}
\subfloat{\includegraphics[width=0.16\linewidth]{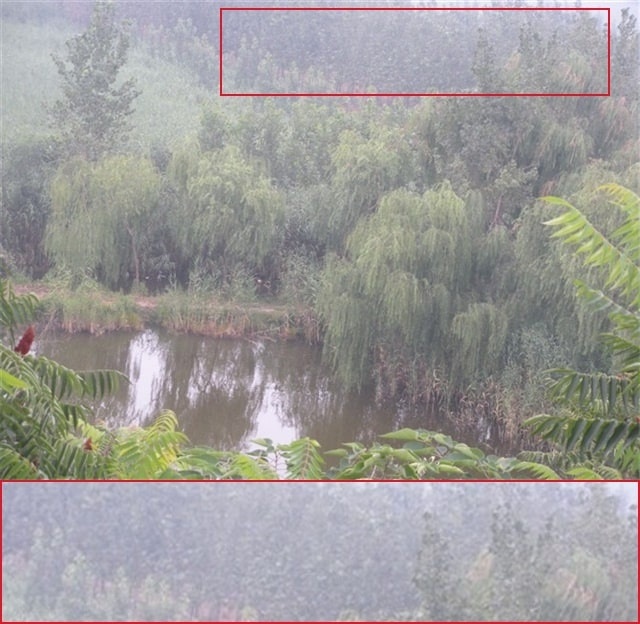}%
\label{}}
\hfil
\subfloat{\includegraphics[width=0.16\linewidth]{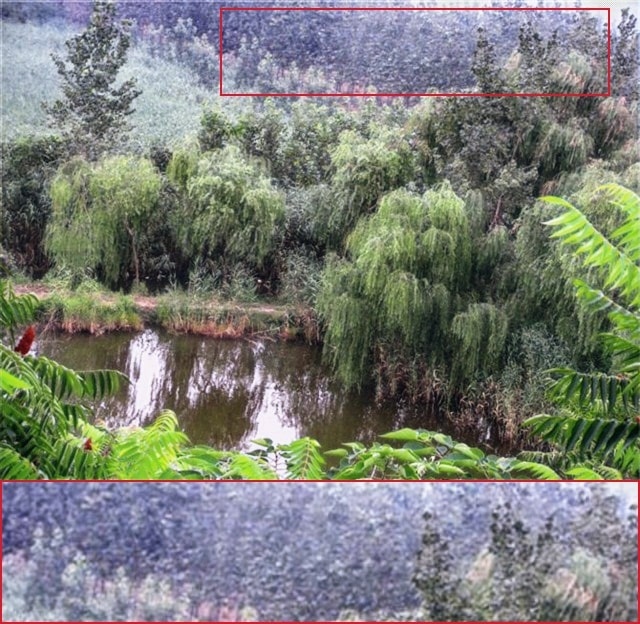}%
\label{}}
\hfil
\subfloat{\includegraphics[width=0.16\linewidth]{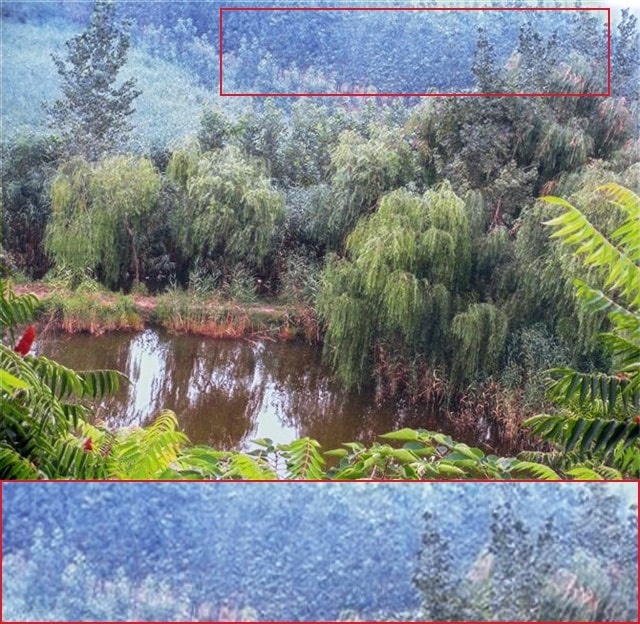}%
\label{}}
\hfil
\subfloat{\includegraphics[width=0.16\linewidth]{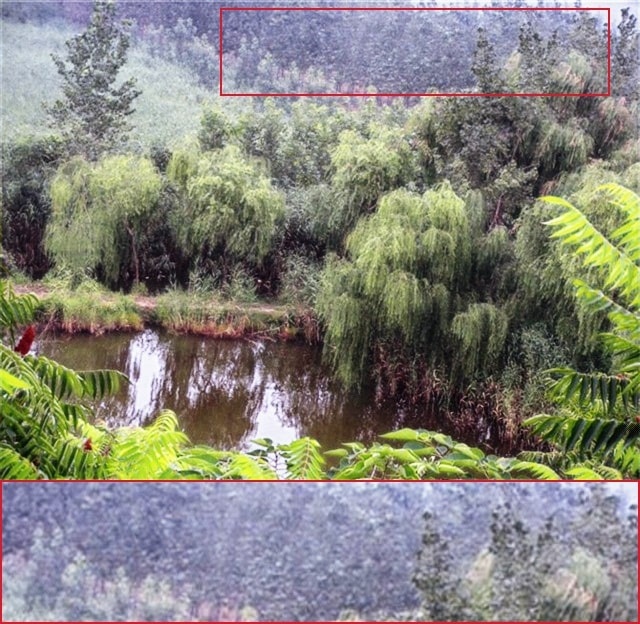}%
\label{}}
\hfil
\subfloat{\includegraphics[width=0.16\linewidth]{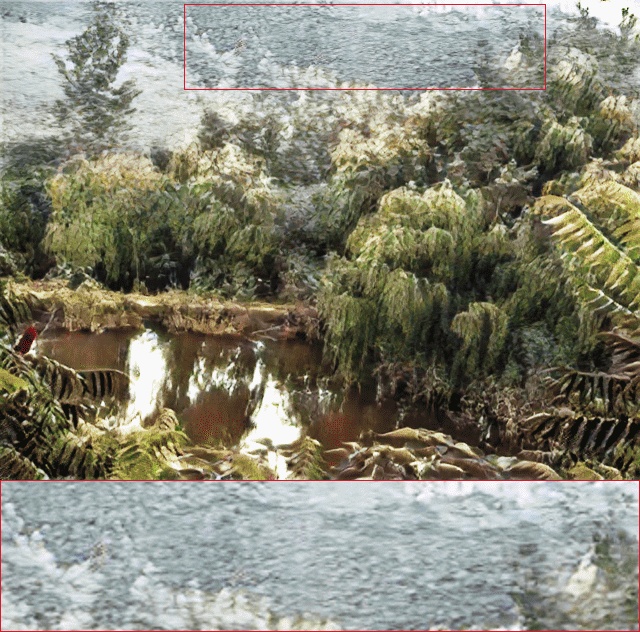}%
\label{}}\hfil
\subfloat{\includegraphics[width=0.16\linewidth]{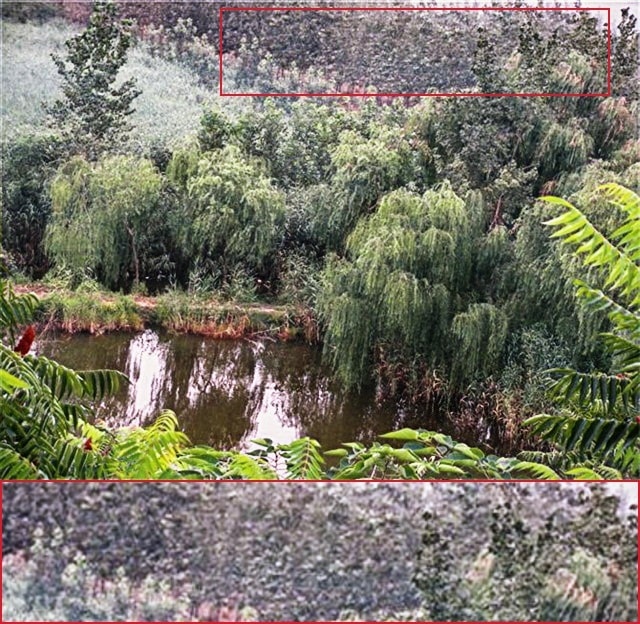}%
\label{}}\hfil\\
\hfil\makebox[0.16\linewidth][c]{(a)}\hfil\makebox[0.16\linewidth][c]{(b)}\hfil\makebox[0.16\linewidth][c]{(c)}\makebox[0.16\linewidth][c]{(d)}\makebox[0.16\linewidth][c]{(e)}\hfil\makebox[0.16\linewidth][c]{(f)}
\caption{Effectiveness of the upscaling inception module. (a) Foggy images. (b)-(d) The results of PG-CycleGAN by using deconvolution, bilinear interpolation, and PixelShuffle alone as the upscaling method, respectively. (e) The results of CycleGAN. (f) The results of PG-CycleGAN.}
\label{}
\end{figure}

In Fig. 12 (b)-(e), the results are generated by using deconvolution, bilinear interpolation, PixelShuffle \cite{shi2016real}, and UIM as the upscaling method, respectively. It can be seen that the zoom-in regions in Fig. 12 (b), (d), and (e) contain more fog than Fig. 12 (f), and the images in Fig. 12 (c) contain color distortion whose color tends to be blue. This figure shows that PG-CycleGAN eliminates color distortion and offers better defogging effect as compared with the baseline methods.

\begin{figure}[h]
\centering
\subfloat{\includegraphics[width=0.16\linewidth]{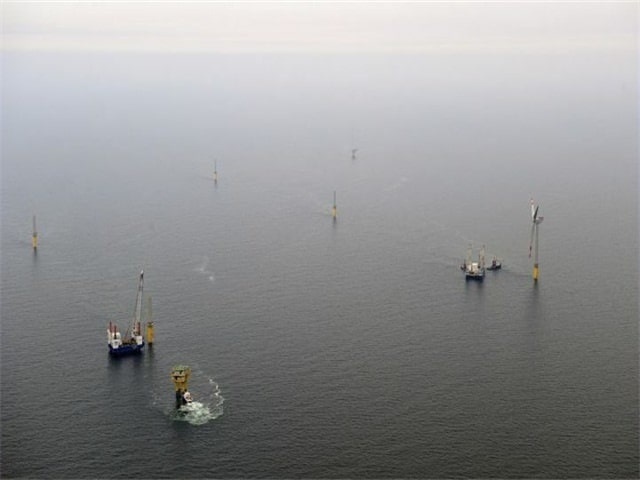}%
\label{}}
\hfil
\subfloat{\includegraphics[width=0.16\linewidth]{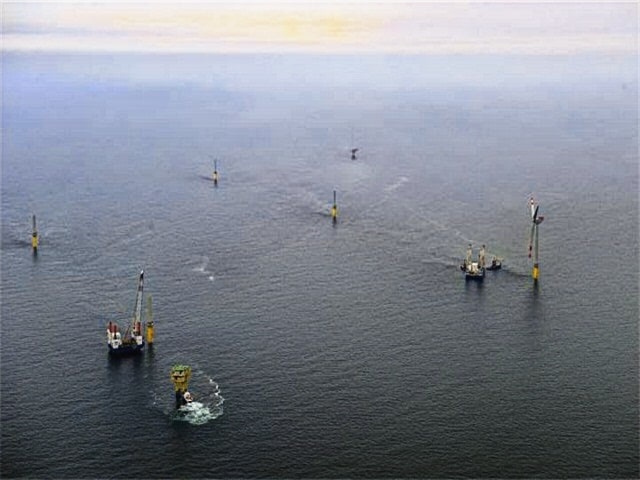}%
\label{}}
\hfil
\subfloat{\includegraphics[width=0.16\linewidth]{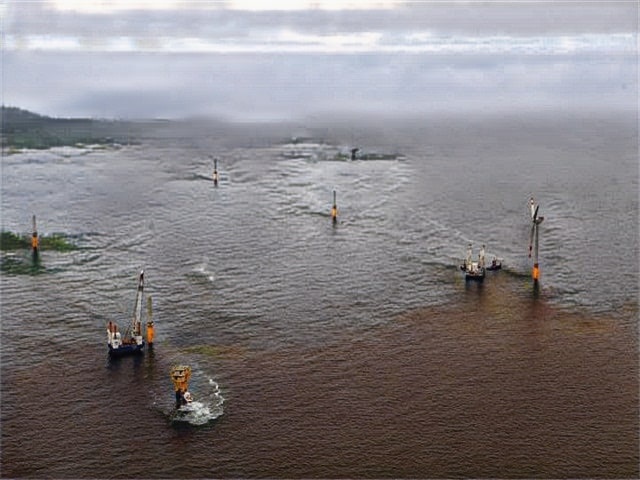}%
\label{}}
\hfil
\subfloat{\includegraphics[width=0.16\linewidth]{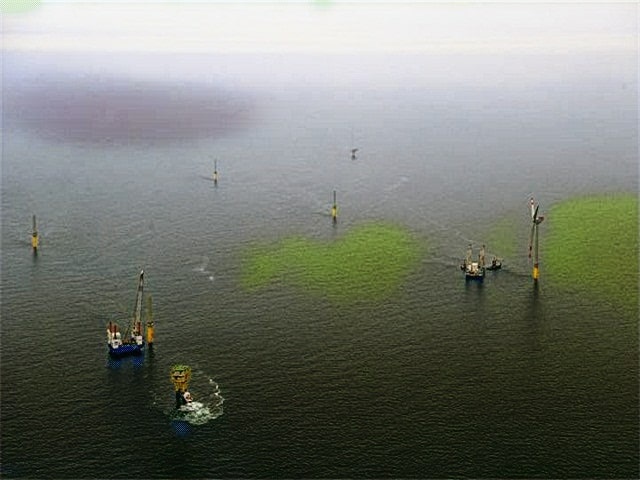}%
\label{}}
\hfil
\subfloat{\includegraphics[width=0.16\linewidth]{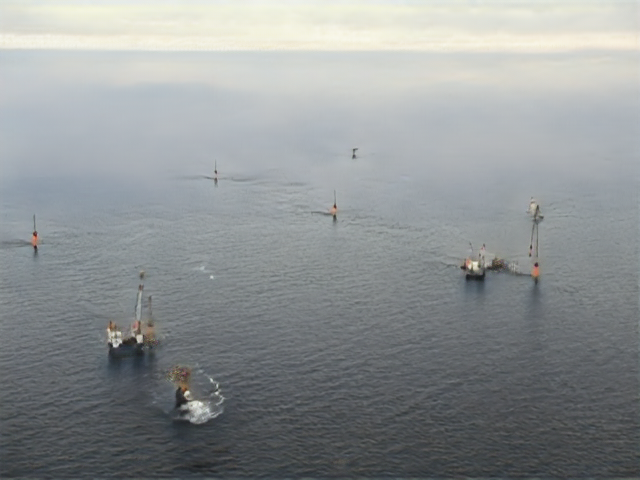}%
\label{}}
\subfloat{\includegraphics[width=0.16\linewidth]{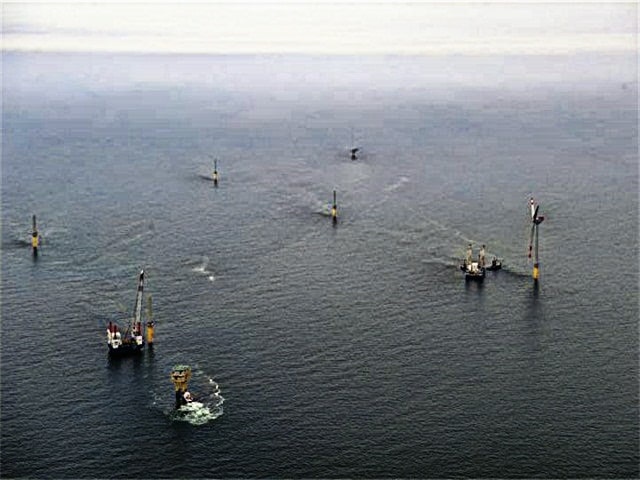}%
\label{}}
\\\vspace{-3mm}
\subfloat{\includegraphics[width=0.16\linewidth]{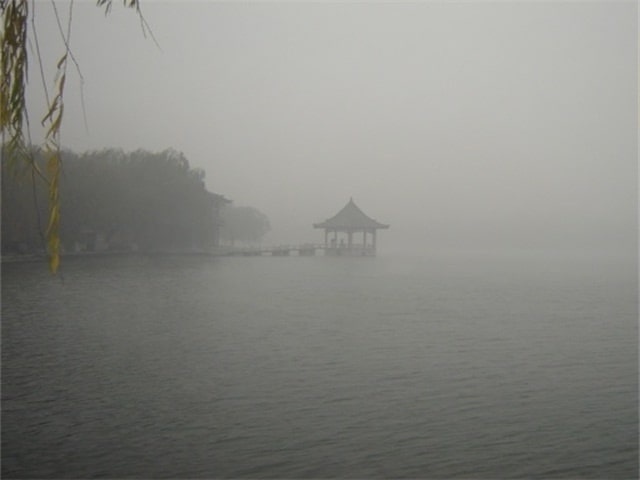}%
\label{}}
\hfil
\subfloat{\includegraphics[width=0.16\linewidth]{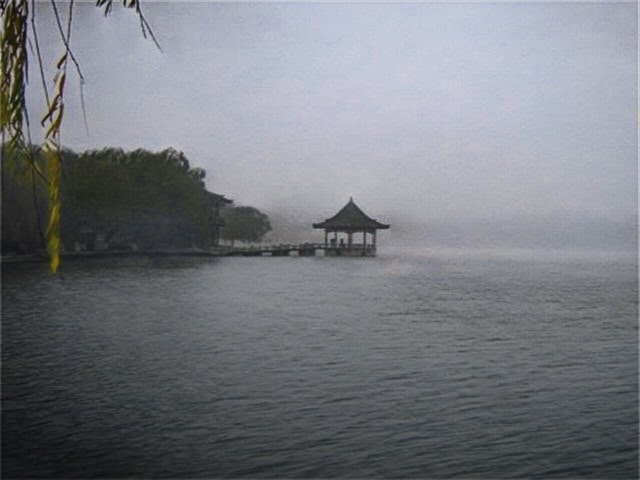}%
\label{}}
\hfil
\subfloat{\includegraphics[width=0.16\linewidth]{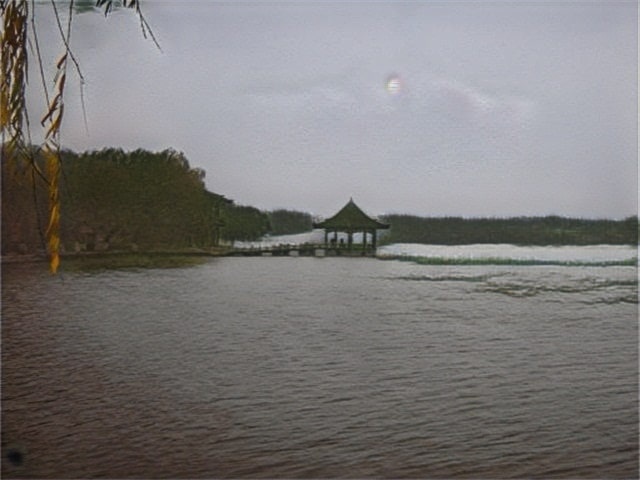}%
\label{}}
\hfil
\subfloat{\includegraphics[width=0.16\linewidth]{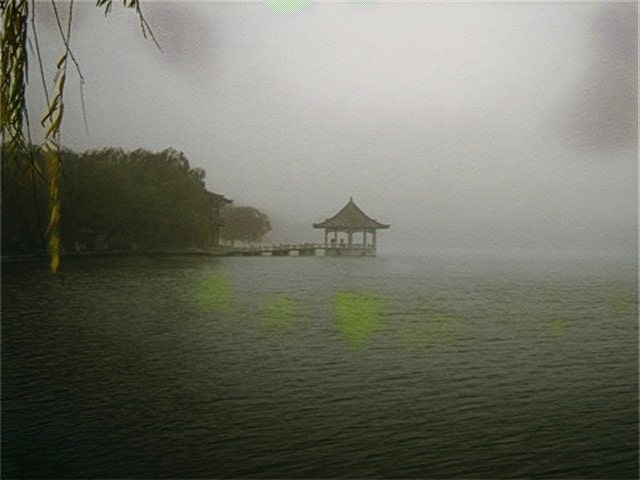}%
\label{}}
\hfil
\subfloat{\includegraphics[width=0.16\linewidth]{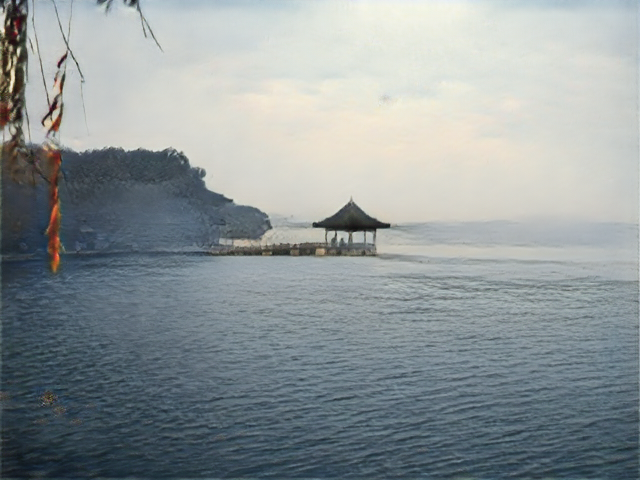}%
\label{}}
\subfloat{\includegraphics[width=0.16\linewidth]{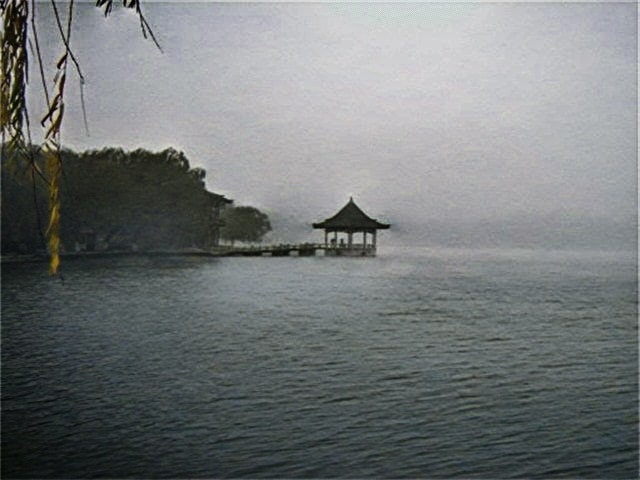}%
\label{}}\\
\hfil\makebox[0.16\linewidth][c]{(a)}\hfil\makebox[0.16\linewidth][c]{(b)}\hfil\makebox[0.16\linewidth][c]{(c)}\hfil\makebox[0.16\linewidth][c]{(d)}\hfil\makebox[0.16\linewidth][c]{(e)}\hfil\makebox[0.16\linewidth][c]{(f)}
\caption{Effectiveness of the prior map guided losses. (a) Foggy images (b) The results of PG-CycleGAN without using the prior map guided cycle-consistency loss. (c) The results of PG-CycleGAN with the PathGAN  \cite{isola2017image} discriminator. (d) The results of PG-CycleGAN with the regular discriminator and without the prior map. (e) The results of CycleGAN. (f) The results of PG-CycleGAN.}
\label{}
\end{figure}

For evaluating the effectiveness of the prior map guided losses, we conduct an ablation study involving the following three control groups: 1) not using the prior map guided cycle-consistency loss, 2) adopting PathGAN \cite{isola2017image} discriminator which is widely used in image generation tasks instead of regular discriminator, 3) using regular discriminator but not imposing the prior map. The visual results are shown in Fig. 13, which shows that with the absence of the prior map guided cycle-consistency loss, the resulting color tends to be blue. In Fig. 13 (c), some regions are blurred and the defogging results are very artificial. Furthermore, with the absence of the prior map, the regular discriminator cannot distinguish between real and generated images, which leads to undesired green regions in the defogging results as shown in Fig. 13 (d). We can see that in Fig. 13 (e), the results of CycleGAN show insufficient defogging. On the contrary, as shown in Fig. 13 (f), the PG-CycleGAN optimized with the prior map offers better visual effects, e.g., the objects are finely recovered.

Overall, as illustrated in Figs. 11-13, the LRC restores detailed information while also providing a desirable defogging effect. The UIM integrates the strengths of three upscaling branches and generates cleaner and visually more pleasing results. The prior map guided losses enhance the recovery of the objects on the water. Table \uppercase\expandafter{\romannumeral4} quantitatively compares the contribution of each part. When either part is removed, both FADE and HazDesNet have increased, i.e., the defogging performance deteriorates. This further demonstrates the effectiveness of our proposed method.

Table \uppercase\expandafter{\romannumeral5} gives the comparison of inference times on a 512$\times$512 image between the CycleGAN \cite{zhu2017unpaired}, the PG-CycleGAN with deconvolution instead of UIM, and the PG-CycleGAN. It can be observed that UIM does not significantly slow down the inference speed. This is benefiting from the 1x1 convolution, which reduces the dimension before upscaling feature maps.

\begin{table}[h]
\renewcommand\arraystretch{1.2}
\begin{center}
\caption{Effectiveness of the PG-CycleGAN}
\label{tab1}
\begin{tabular}{ l  c  c }
\toprule
 & FADE $\downarrow$  & HazDesNet $\downarrow$ \\
\hline
w/ Deconv  & 1.0522 & 0.3836 \\
w/ Bilinear  & 1.2644 & 0.3727 \\
w/ PixelShuffle  & 0.7839 & 0.3743 \\
w/o $L_{pgcyc}$  & 1.0411 & 0.3728 \\
w/ PatchGAN D  & 0.9966 & 0.3972 \\
w/ regular D  & 0.9472 & 0.3725 \\
CycleGAN  & 1.5970 & 0.6032 \\
PG-CycleGAN  & 0.7544 & 0.3240 \\
\bottomrule
\end{tabular}
\end{center}
\end{table}

\begin{table}[h]
\renewcommand\arraystretch{1.2}
\begin{center}
\caption{Comparison of inference time}
\label{tab1}
\begin{tabular}{ l  c  c }
\toprule
 & Time(s) \\
\hline
CycleGAN  & 0.118 \\
w/ Deconv  & 0.164 \\
PG-CycleGAN & 0.176 \\
\bottomrule
\end{tabular}
\end{center}
\end{table}

\subsection{Limitations}
As with previous methods, there are limitations in the proposed method. As shown in Fig. 14, when the local area in the original input image is near to white, which usually happens on white hulls, the pixel in the corresponding prior map will tend to 0. The reason is that if all the pixels of the three channels in the input possess a large value, the local region will show brightly in the dark channel, and then appear dark after the dark channel is inverted. This may lead to suboptimal solutions in network optimization, as it cannot give the objects a high penalty. The following solutions could be used to overcome this problem, i.e., refining the weights of the prior map guided losses, or adding an extra PatchGAN \cite{isola2017image} discriminator to optimize the model.
\begin{figure}[h]
\centering
\subfloat{\includegraphics[width=0.3\linewidth]{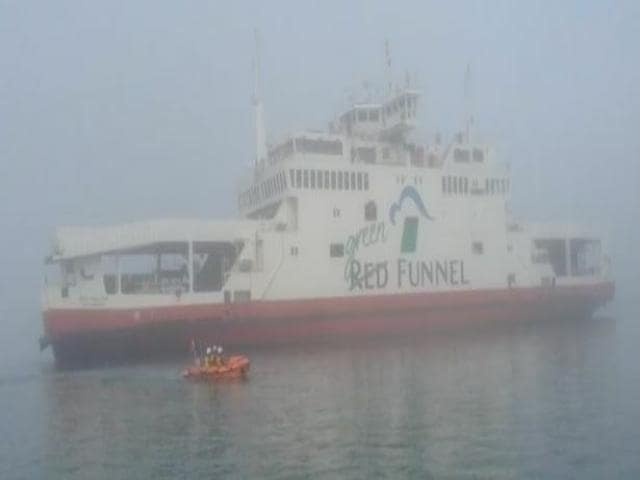}%
\label{}}
\hfil
\subfloat{\includegraphics[width=0.3\linewidth]{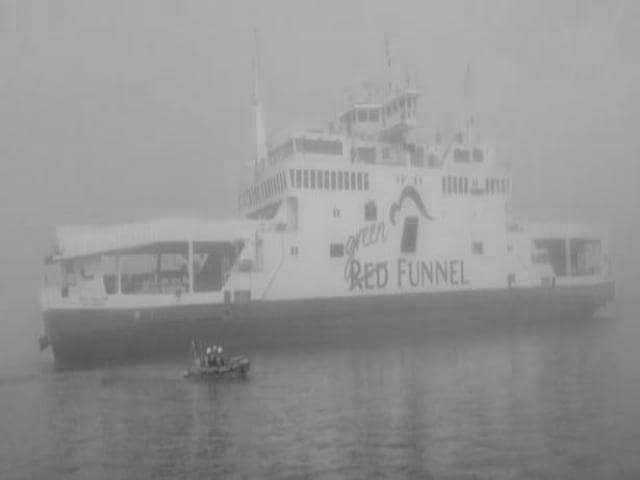}%
\label{}}
\hfil
\subfloat{\includegraphics[width=0.3\linewidth]{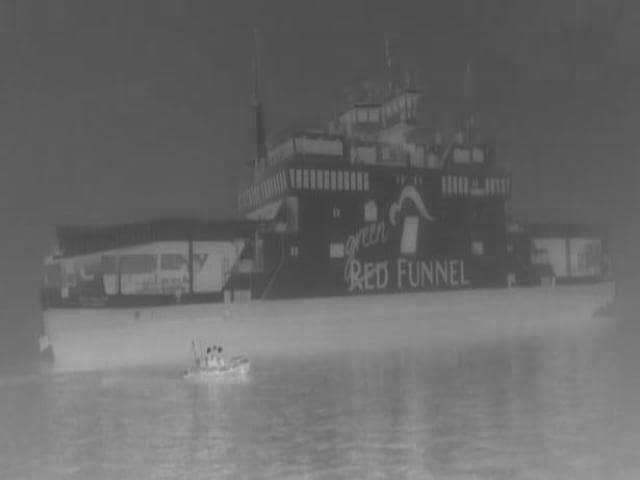}%
\label{}}
\\\vspace{-3mm}
\subfloat{\includegraphics[width=0.3\linewidth]{fig13_a_1.jpg}%
\label{}}
\hfil
\subfloat{\includegraphics[width=0.3\linewidth]{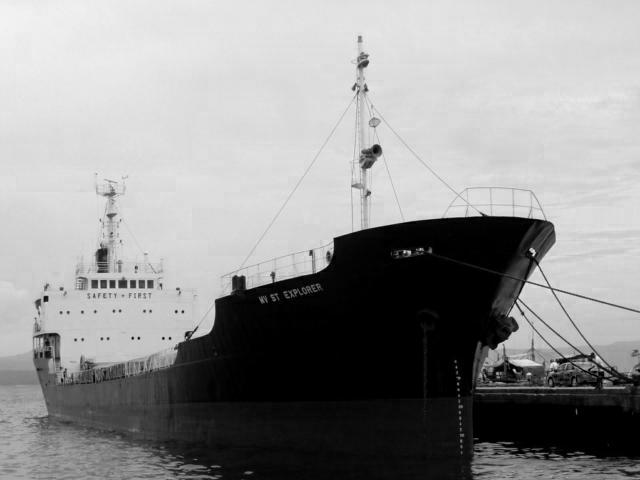}%
\label{}}
\hfil
\subfloat{\includegraphics[width=0.3\linewidth]{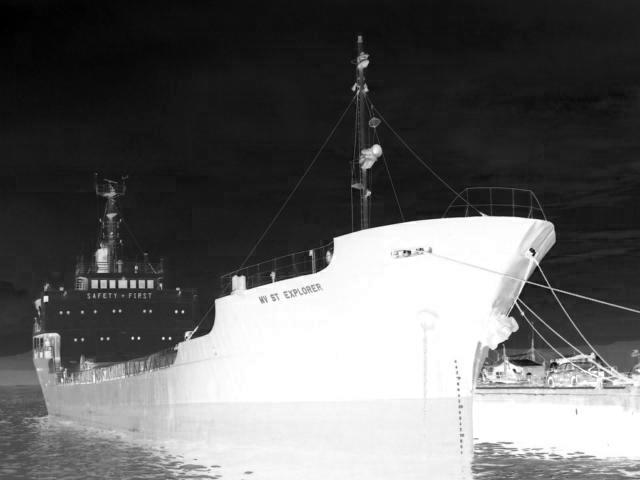}%
\label{}}\\
\hfil\makebox[0.3\linewidth][c]{(a) Input image}\hfil\makebox[0.3\linewidth][c]{(b) Dark channel}\hfil\makebox[0.3\linewidth][c]{(c) Prior map}
\caption{The prior map does not work well in white areas.}
\label{}
\end{figure}

\section{Concluding Remarks}
In this paper, we have presented a prior map guided cycle-consistent adversarial network for overwater image defogging. We have proposed a method for estimating the prior map and incorporated it in two losses, imposing a greater penalty on objects on the water. Due to the fact that the unpaired training set may lead to color distortion, we have proposed two complementary methods to improve the defogging results: the long-range residual connection and coarse-to-fine framework. Moreover, an intuitive and effective upscaling inception module is exploited, by combining the strengths of three upscaling branches: deconvolution, bilinear interpolation, and PixelShuffle \cite{shi2016real}, to generate images of improved quality. Extensive experiments on the qualitative and quantitative comparisons demonstrate the advantages of PG-CycleGAN against several state-of-the-art methods. Furthermore, benefiting from the feature that the proposed prior map can highlight objects on water and distinguish them from the background, the prior map may help in other research directions, such as maritime object detection, tracking, and segmentation.




%



\AtNextBibliography{\small}\printbibliography{}

@inproceedings{O-HAZE_2018,
author = { Codruta O. Ancuti and Cosmin Ancuti and Radu Timofte and Christophe De Vleeschouwer},
title = {O-HAZE: a dehazing benchmark with real hazy and haze-free outdoor images},
booktitle =  {IEEE Conference on Computer Vision and Pattern Recognition, NTIRE Workshop },
series = {NTIRE CVPR'18},
year = {2018},
location = {Salt Lake City, Utah, USA},
}

@inproceedings{NH-Haze_2020,
author = {Codruta O. Ancuti and Cosmin Ancuti and Radu Timofte},
title = {{NH-HAZE:} An Image Dehazing Benchmark with Non-Homogeneous Hazy and Haze-Free Images},
booktitle =  {Proceedings of the IEEE Conference on Computer Vision and Pattern Recognition Workshops},
series = {IEEE CVPR 2020},
year = {2020},
location = {Washington, US},
}

@InProceedings{Liang_2022_IJCAI,
    author = {Liang, Yudong and Wang, Bin and Zuo, Wangmeng and Liu, Jiaying and Ren, Wenqi},
  title = {Self-supervised Learning and Adaptation for Single Image Dehazing},
  booktitle = {Proceedings of the 31st International Joint Conference on Artificial Intelligence (IJCAI-22)},
   pages = {1137-1143},
  year = {2022}
}

@InProceedings{Dong_2020_CVPR,
author = {Dong, Hang and Pan, Jinshan and Xiang, Lei and Hu, Zhe and Zhang, Xinyi and Wang, Fei and Yang, Ming-Hsuan},
title = {Multi-Scale Boosted Dehazing Network With Dense Feature Fusion},
booktitle = {Proceedings of the IEEE/CVF Conference on Computer Vision and Pattern Recognition (CVPR)},
month = {June},
year = {2020}
}

@article{he2010single,
  title={Single image haze removal using dark channel prior},
  author={He, Kaiming and Sun, Jian and Tang, Xiaoou},
  journal={IEEE transactions on pattern analysis and machine intelligence},
  volume={33},
  number={12},
  pages={2341--2353},
  year={2010},
  publisher={IEEE}
}

@inproceedings{li2017aod,
  title={Aod-net: All-in-one dehazing network},
 author={Li, Boyi and Peng, Xiulian and Wang, Zhangyang and Xu, Jizheng and Feng, Dan},
  booktitle={2017 IEEE International Conference on Computer Vision (ICCV)}, 
  title={AOD-Net: All-in-One Dehazing Network}, 
  year={2017},
  volume={},
  number={},
  pages={4780-4788}}

@inproceedings{qu2019enhanced,
  author={Qu, Yanyun and Chen, Yizi and Huang, Jingying and Xie, Yuan},
  booktitle={2019 IEEE/CVF Conference on Computer Vision and Pattern Recognition (CVPR)}, 
  title={Enhanced Pix2pix Dehazing Network}, 
  year={2019},
  volume={},
  number={},
  pages={8152-8160}}

@inproceedings{nayar1999vision,
  title={Vision in bad weather},
  author={Nayar, Shree K and Narasimhan, Srinivasa G},
  booktitle={Proceedings of the seventh IEEE international conference on computer vision},
  volume={2},
  pages={820--827},
  year={1999},
  organization={IEEE}
}

@article{mccartney1976optics,
  author={Cantor, A.},
  journal={IEEE Journal of Quantum Electronics}, 
  title={Optics of the atmosphere--Scattering by molecules and particles}, 
  year={1978},
  volume={14},
  number={9},
  pages={698-699}}

@article{fattal2014dehazing,
  title={Dehazing using color-lines},
  author={Fattal, Raanan},
  journal={ACM transactions on graphics (TOG)},
  volume={34},
  number={1},
  pages={1--14},
  year={2014},
  publisher={ACM New York, NY, USA}
}

@article{zhu2015fast,
  title={A fast single image haze removal algorithm using color attenuation prior},
  author={Zhu, Qingsong and Mai, Jiaming and Shao, Ling},
  journal={IEEE transactions on image processing},
  volume={24},
  number={11},
  pages={3522--3533},
  year={2015},
  publisher={IEEE}
}

@inproceedings{berman2016non,
  title={Non-local image dehazing},
  author={Berman, Dana and Avidan, Shai and others},
  booktitle={Proceedings of the IEEE conference on computer vision and pattern recognition},
  pages={1674--1682},
  year={2016}
}

@article{goodfellow2014generative,
  title={Generative adversarial nets},
  author={Goodfellow, Ian and Pouget-Abadie, Jean and Mirza, Mehdi and Xu, Bing and Warde-Farley, David and Ozair, Sherjil and Courville, Aaron and Bengio, Yoshua},
  journal={Advances in neural information processing systems},
  volume={27},
  year={2014}
}

@article{li2019semi,
  title={Semi-supervised image dehazing},
  author={Li, Lerenhan and Dong, Yunlong and Ren, Wenqi and Pan, Jinshan and Gao, Changxin and Sang, Nong and Yang, Ming-Hsuan},
  journal={IEEE Transactions on Image Processing},
  volume={29},
  pages={2766--2779},
  year={2019},
  publisher={IEEE}
}

@inproceedings{shao2020domain,
  title={Domain adaptation for image dehazing},
  author={Shao, Yuanjie and Li, Lerenhan and Ren, Wenqi and Gao, Changxin and Sang, Nong},
  booktitle={Proceedings of the IEEE/CVF conference on computer vision and pattern recognition},
  pages={2808--2817},
  year={2020}
}

@article{golts2019unsupervised,
  title={Unsupervised single image dehazing using dark channel prior loss},
  author={Golts, Alona and Freedman, Daniel and Elad, Michael},
  journal={IEEE Transactions on Image Processing},
  volume={29},
  pages={2692--2701},
  year={2019},
  publisher={IEEE}
}

@inproceedings{zheng2020overwater,
  title={Overwater Image Dehazing via Cycle-Consistent Generative Adversarial Network},
  author={Zheng, Shunyuan and Sun, Jiamin and Liu, Qinglin and Qi, Yuankai and Zhang, Shengping},
  booktitle={Proceedings of the Asian Conference on Computer Vision},
  year={2020}
}

@inproceedings{zhu2017unpaired,
  title={Unpaired image-to-image translation using cycle-consistent adversarial networks},
  author={Zhu, Jun-Yan and Park, Taesung and Isola, Phillip and Efros, Alexei A},
  booktitle={Proceedings of the IEEE international conference on computer vision},
  pages={2223--2232},
  year={2017}
}

@inproceedings{ren2018gated,
  title={Gated fusion network for single image dehazing},
  author={Ren, Wenqi and Ma, Lin and Zhang, Jiawei and Pan, Jinshan and Cao, Xiaochun and Liu, Wei and Yang, Ming-Hsuan},
  booktitle={Proceedings of the IEEE Conference on Computer Vision and Pattern Recognition},
  pages={3253--3261},
  year={2018}
}

@inproceedings{szegedy2015going,
  title={Going deeper with convolutions},
  author={Szegedy, Christian and Liu, Wei and Jia, Yangqing and Sermanet, Pierre and Reed, Scott and Anguelov, Dragomir and Erhan, Dumitru and Vanhoucke, Vincent and Rabinovich, Andrew},
  booktitle={Proceedings of the IEEE conference on computer vision and pattern recognition},
  pages={1--9},
  year={2015}
}

@inproceedings{shi2016real,
  title={Real-time single image and video super-resolution using an efficient sub-pixel convolutional neural network},
  author={Shi, Wenzhe and Caballero, Jose and Husz{\'a}r, Ferenc and Totz, Johannes and Aitken, Andrew P and Bishop, Rob and Rueckert, Daniel and Wang, Zehan},
  booktitle={Proceedings of the IEEE conference on computer vision and pattern recognition},
  pages={1874--1883},
  year={2016}
}

@inproceedings{hu2018squeeze,
  title={Squeeze-and-excitation networks},
  author={Hu, Jie and Shen, Li and Sun, Gang},
  booktitle={Proceedings of the IEEE conference on computer vision and pattern recognition},
  pages={7132--7141},
  year={2018}
}

@article{fattal2008single,
  title={Single image dehazing},
  author={Fattal, Raanan},
  journal={ACM transactions on graphics (TOG)},
  volume={27},
  number={3},
  pages={1--9},
  year={2008},
  publisher={ACM New York, NY, USA}
}

@inproceedings{tan2008visibility,
  title={Visibility in bad weather from a single image},
  author={Tan, Robby T},
  booktitle={2008 IEEE conference on computer vision and pattern recognition},
  pages={1--8},
  year={2008},
  organization={IEEE}
}

@inproceedings{tarel2009fast,
  title={Fast visibility restoration from a single color or gray level image},
  author={Tarel, Jean-Philippe and Hautiere, Nicolas},
  booktitle={2009 IEEE 12th international conference on computer vision},
  pages={2201--2208},
  year={2009},
  organization={IEEE}
}

@inproceedings{meng2013efficient,
  title={Efficient image dehazing with boundary constraint and contextual regularization},
  author={Meng, Gaofeng and Wang, Ying and Duan, Jiangyong and Xiang, Shiming and Pan, Chunhong},
  booktitle={Proceedings of the IEEE international conference on computer vision},
  pages={617--624},
  year={2013}
}

@article{cai2016dehazenet,
  title={Dehazenet: An end-to-end system for single image haze removal},
  author={Cai, Bolun and Xu, Xiangmin and Jia, Kui and Qing, Chunmei and Tao, Dacheng},
  journal={IEEE Transactions on Image Processing},
  volume={25},
  number={11},
  pages={5187--5198},
  year={2016},
  publisher={IEEE}
}

@inproceedings{ren2016single,
  title={Single image dehazing via multi-scale convolutional neural networks},
  author={Ren, Wenqi and Liu, Si and Zhang, Hua and Pan, Jinshan and Cao, Xiaochun and Yang, Ming-Hsuan},
  booktitle={European conference on computer vision},
  pages={154--169},
  year={2016},
  organization={Springer}
}

@inproceedings{liu2019griddehazenet,
  title={Griddehazenet: Attention-based multi-scale network for image dehazing},
  author={Liu, Xiaohong and Ma, Yongrui and Shi, Zhihao and Chen, Jun},
  booktitle={Proceedings of the IEEE/CVF International Conference on Computer Vision},
  pages={7314--7323},
  year={2019}
}

@inproceedings{dong2020multi,
  title={Multi-scale boosted dehazing network with dense feature fusion},
  author={Dong, Hang and Pan, Jinshan and Xiang, Lei and Hu, Zhe and Zhang, Xinyi and Wang, Fei and Yang, Ming-Hsuan},
  booktitle={Proceedings of the IEEE/CVF conference on computer vision and pattern recognition},
  pages={2157--2167},
  year={2020}
}

@inproceedings{qin2020ffa,
  title={FFA-Net: Feature fusion attention network for single image dehazing},
  author={Qin, Xu and Wang, Zhilin and Bai, Yuanchao and Xie, Xiaodong and Jia, Huizhu},
  booktitle={Proceedings of the AAAI Conference on Artificial Intelligence},
  volume={34},
  number={07},
  pages={11908--11915},
  year={2020}
}

@inproceedings{engin2018cycle,
  title={Cycle-dehaze: Enhanced cyclegan for single image dehazing},
  author={Engin, Deniz and Gen{\c{c}}, Anil and Kemal Ekenel, Hazim},
  booktitle={Proceedings of the IEEE conference on computer vision and pattern recognition workshops},
  pages={825--833},
  year={2018}
}

@article{zhao2021refinednet,
  title={RefineDNet: a weakly supervised refinement framework for single image dehazing},
  author={Zhao, Shiyu and Zhang, Lin and Shen, Ying and Zhou, Yicong},
  journal={IEEE Transactions on Image Processing},
  volume={30},
  pages={3391--3404},
  year={2021},
  publisher={IEEE}
}

@inproceedings{chen2021psd,
  title={PSD: Principled synthetic-to-real dehazing guided by physical priors},
  author={Chen, Zeyuan and Wang, Yangchao and Yang, Yang and Liu, Dong},
  booktitle={Proceedings of the IEEE/CVF Conference on Computer Vision and Pattern Recognition},
  pages={7180--7189},
  year={2021}
}

@inproceedings{he2016deep,
  title={Deep residual learning for image recognition},
  author={He, Kaiming and Zhang, Xiangyu and Ren, Shaoqing and Sun, Jian},
  booktitle={Proceedings of the IEEE conference on computer vision and pattern recognition},
  pages={770--778},
  year={2016}
}

@inproceedings{isola2017image,
  title={Image-to-image translation with conditional adversarial networks},
  author={Isola, Phillip and Zhu, Jun-Yan and Zhou, Tinghui and Efros, Alexei A},
  booktitle={Proceedings of the IEEE conference on computer vision and pattern recognition},
  pages={1125--1134},
  year={2017}
}

@article{choi2015referenceless,
  title={Referenceless prediction of perceptual fog density and perceptual image defogging},
  author={Choi, Lark Kwon and You, Jaehee and Bovik, Alan Conrad},
  journal={IEEE Transactions on Image Processing},
  volume={24},
  number={11},
  pages={3888--3901},
  year={2015},
  publisher={IEEE}
}

@ARTICLE{zhang2020hazdesnet,
  author={Zhang, Jiahe and Min, Xiongkuo and Zhu, Yucheng and Zhai, Guangtao and Zhou, Jiantao and Yang, Xiaokang and Zhang, Wenjun},
  journal={IEEE Transactions on Intelligent Transportation Systems}, 
  title={HazDesNet: An End-to-End Network for Haze Density Prediction}, 
  year={2022},
  volume={23},
  number={4},
  pages={3087-3102}}

@article{hautiere2008blind,
  title={Blind contrast enhancement assessment by gradient ratioing at visible edges},
  author={Hautiere, Nicolas and Tarel, Jean-Philippe and Aubert, Didier and Dumont, Eric},
  journal={Image Analysis \& Stereology},
  volume={27},
  number={2},
  pages={87--95},
  year={2008}
}

@article{simonyan2014very,
  title={Very deep convolutional networks for large-scale image recognition},
  author={Simonyan, Karen and Zisserman, Andrew},
  journal={arXiv preprint arXiv:1409.1556},
  year={2014}
}

@article{kingma2014adam,
  title={Adam: A method for stochastic optimization},
  author={Kingma, Diederik P and Ba, Jimmy},
  journal={arXiv preprint arXiv:1412.6980},
  year={2014}
}

@article{guan2022visibility,
  author={Guan, Tuxin and Li, Chaofeng and Gu, Ke and Liu, Hantao and Zheng, Yuhui and Wu, Xiao-Jun},
  journal={IEEE Transactions on Multimedia}, 
  title={Visibility and Distortion Measurement for No-Reference Dehazed Image Quality Assessment via Complex Contourlet Transform}, 
  year={2022},
  volume={},
  number={},
  pages={1-17},
  doi={10.1109/TMM.2022.3168438}}

@ARTICLE{7814312,  author={Wang, Wencheng and Yuan, Xiaohui and Wu, Xiaojin and Liu, Yunlong},  journal={IEEE Transactions on Multimedia},   title={Fast Image Dehazing Method Based on Linear Transformation},   year={2017},  volume={19},  number={6},  pages={1142-1155},  doi={10.1109/TMM.2017.2652069}}

@ARTICLE{9537303,  author={Liu, Xiaoning and Li, Hui and Zhu, Ce},  journal={IEEE Transactions on Multimedia},   title={Joint Contrast Enhancement and Exposure Fusion for Real-World Image Dehazing},   year={2021},  volume={},  number={},  pages={1-1},  doi={10.1109/TMM.2021.3110483}}

@ARTICLE{9473023,  author={Yi, Qiaosi and Li, Juncheng and Fang, Faming and Jiang, Aiwen and Zhang, Guixu},  journal={IEEE Transactions on Multimedia},   title={Efficient and Accurate Multi-scale Topological Network for Single Image Dehazing},   year={2021},  volume={},  number={},  pages={1-1},  doi={10.1109/TMM.2021.3093724}}

@ARTICLE{9745359,  author={Li, Jiafeng and Li, Yaopeng and Zhuo, Li and Kuang, Lingyan and Yu, Tianjian},  journal={IEEE Transactions on Multimedia},   title={USID-Net: Unsupervised Single Image Dehazing Network via Disentangled Representations},   year={2022},  volume={},  number={},  pages={1-1},  doi={10.1109/TMM.2022.3163554}}

@inproceedings{yang2018towards,
  title={Towards perceptual image dehazing by physics-based disentanglement and adversarial training},
  author={Yang, Xitong and Xu, Zheng and Luo, Jiebo},
  booktitle={Proceedings of the AAAI Conference on Artificial Intelligence},
  volume={32},
  number={1},
  year={2018}
}

\end{document}